\documentclass[10pt,journal,compsoc]{IEEEtran}
\usepackage{amsmath,epsfig,amssymb,amsfonts,amsthm,epsfig,verbatim}
\usepackage{color,graphicx,lscape,longtable, subfig}
\usepackage{xr}
\usepackage[usenames,dvipsnames]{xcolor}
\usepackage{fancyhdr}
\usepackage{amsmath}
\usepackage[numbers]{natbib}
\usepackage{enumitem}
\renewcommand{\baselinestretch}{1.5}

\def\boxit#1{\vbox{\hrule\hbox{\vrule\kern6pt
          \vbox{\kern6pt#1\kern6pt}\kern6pt\vrule}\hrule}}

\newtheorem{Th}{\underline{\bf Theorem}}

\def\bse{\begin{eqnarray*}}
  \def\ese{\end{eqnarray*}}
\def\be{\begin{eqnarray}}
  \def\ee{\end{eqnarray}}
\def\bsq{\begin{equation*}}
  \def\esq{\end{equation*}}
\def\bq{\begin{equation}}
  \def\eq{\end{equation}}

\def\trans{^{\rm T}}

\def\0{{\bf 0}}

\def\g{{\bf g}}

\def\w{{\bf w}}
\def\X{{\bf X}}

\def\bSig{{\bf \Sigma}}
\def\bOmega{{\bf \Omega}}

\def\bq{\begin{equation}}
  \def\eq{\end{equation}}

\def\trans{^{\rm T}}

\def\log{{\rm log}}

\def\squarebox#1{\hbox to #1{\hfill\vbox to #1{\vfill}}}
\def\btheta{{\boldsymbol \theta}}

\def\bse{\begin{eqnarray*}}
  \def\ese{\end{eqnarray*}}
\def\be{\begin{eqnarray}}
  \def\ee{\end{eqnarray}}
\def\bsq{\begin{equation*}}
  \def\esq{\end{equation*}}
\def\bq{\begin{equation}}
  \def\eq{\end{equation}}

\def\trans{^{\rm T}}

\def\boxit#1{\vbox{\hrule\hbox{\vrule\kern6pt\vbox{\kern6pt#1\kern6pt}\kern6pt\vrule}\hrule}}

\renewcommand{\baselinestretch}{1.05}

\newcommand{\twofigs}[2]{
\hbox to\hsize{\hss
\vbox{\psfig{figure=#1,width=2.7in,height=2.0in}}\qquad
\vbox{\psfig{figure=#2,width=2.7in,height=2.0in}}
\hss}}

\def\boxit#1{\vbox{\hrule\hbox{\vrule\kern6pt
      \vbox{\kern6pt#1\kern6pt}\kern6pt\vrule}\hrule}}

\def\theequation{\arabic{equation}}

\begin{document}
\thispagestyle{empty}
\allowdisplaybreaks

\title{Bayesian Outdoor Defect Detection
}

\author{Fei Jiang, 
Guosheng Yin
\IEEEcompsocitemizethanks{\IEEEcompsocthanksitem Fei Jiang was with the Department
of Department of Statistics and Actuarial Science,  University of Hong
Kong. \protect\\
Email: feijiang@hku.hk
\IEEEcompsocthanksitem Guosheng Yin was with the Department
of Department of Statistics and Actuarial Science,  University of Hong
Kong. \protect\\
Email: gyin@hku.hk}
\thanks{Manuscript received 00; revised 00, 2018.}}
\markboth{Journal of \LaTeX\ Class Files,~Vol.~14, No.~8, August~2015}%
{Shell \MakeLowercase{\textit{et al.}}: Bare Advanced Demo of IEEEtran.cls for IEEE Computer Society Journals}
%\doublespacing
\IEEEtitleabstractindextext{
\begin{abstract}
We
introduce a Bayesian defect detector to facilitate the defect
detection on the motion blurred images on rough texture surfaces. To enhance the  accuracy of
Bayesian detection on removing non-defect pixels, we develop a class of reflected
non-local prior distributions, which is constructed by using the mode of
a distribution to subtract its density. The reflected
non-local priors forces the Bayesian detector to approach 0 at the
non-defect locations. We conduct experiments studies to demonstrate
the superior performance of the Bayesian detector in eliminating
the non-defect points. We implement the Bayesian detector in the motion
blurred drone images, in which the detector successfully identifies
 the hail damages on the rough
surface and substantially enhances the accuracy of the
entire defect detection pipeline.
\end{abstract}

\begin{IEEEkeywords}
Bayes factor,  Canny edge detection,  Generalized Gaussian
distribution,  Hail
  damage detection,  Image
  analysis,  Non-local
prior. 
\end{IEEEkeywords}
}

\maketitle
\IEEEdisplaynontitleabstractindextext
\IEEEpeerreviewmaketitle
\ifCLASSOPTIONcompsoc
\IEEEraisesectionheading{\section{Introduction}\label{sec:introduction}}
\else
\section{Introduction}
\fi
Computer aided automatic defect detection have been successfully used
in the industry to ensure the quantity of the product and
allow the
timely maintenance. Along with the development of unmanned 
vehicle, such as drone, more images are taken under imperfect
setting. The resulting images could contain rough surface or could be blurred by the
motion in the image capture process, which leads to serious
over identification of the abnormals. The problem
directly impedes the development of the outdoor defect detection, such
as hail damages detection, 
especially in the house roof industry where the defects occur on the
rough roof shingles and the over identification could significantly level up the
costs of maintenance.

To label and extract the defect automatically, one useful tool is the edge
detection, which identifies the points at the locations of  significant
local intensity changes \citep{jain1995}.
 The commonly used edge detectors
include the gradient based Sobel
\citep{matthews2002}, Roberts \citep{roberts1963},
Prewitt \citep{prewitt1970} detectors, the second derivative Laplacian of a Gaussian
detector \citep{torre1986}, and the Canny detector
\citep{canny1986}. Because of the multi-stage improvements, the Canny detector often
performs better than
the others \citep{bhardwaj2012}.
As an application, we use the  Canny procedure
\citep{canny1986} to search the edge surrounding the defects on the roof
shingles. Note that  the images taken in motion are inevitably  blurred due to the unstable
movements in the image capture process. These motion blurs
exacerbate
the appearances of the surfaces, and in turn increase the difficulties
of suppressing the non-edge pixels. To see that, we implement the Canny detector on
the three blurred non-defect shingles. The right panel in Figure \ref{fig:nodm} shows
that the Canny detector is substantially  misled by the motion
blurs, while the proposed Bayesian detector (will be discussed later) is robust in removing the
non-edge pixels for the blurred images.
\begin{figure}[!h]
\centering
\begin{tabular}{ccc}
Images & Bayesian detector& Canny detector\\
\includegraphics[scale = 0.1]{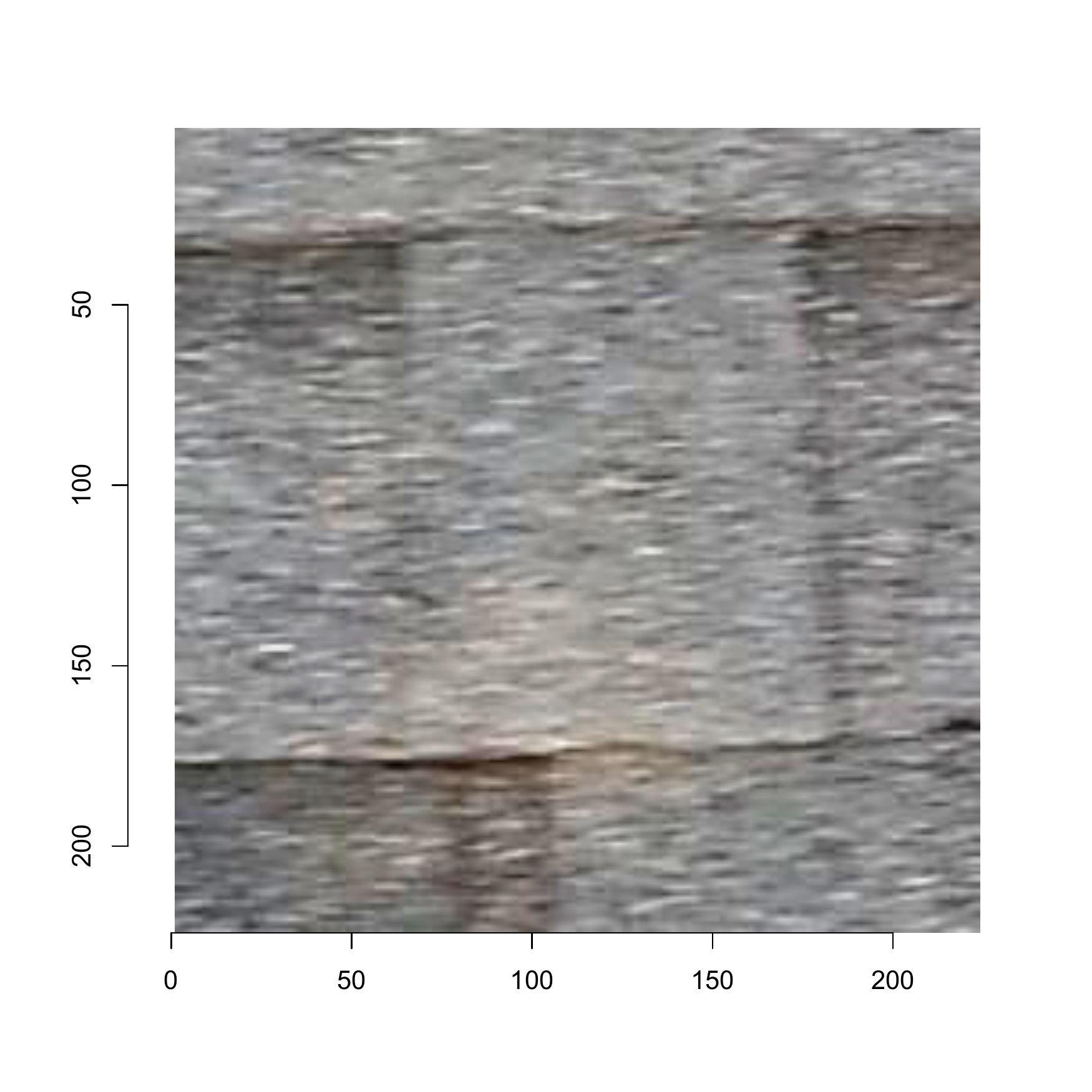}&
\includegraphics[scale = 0.1]{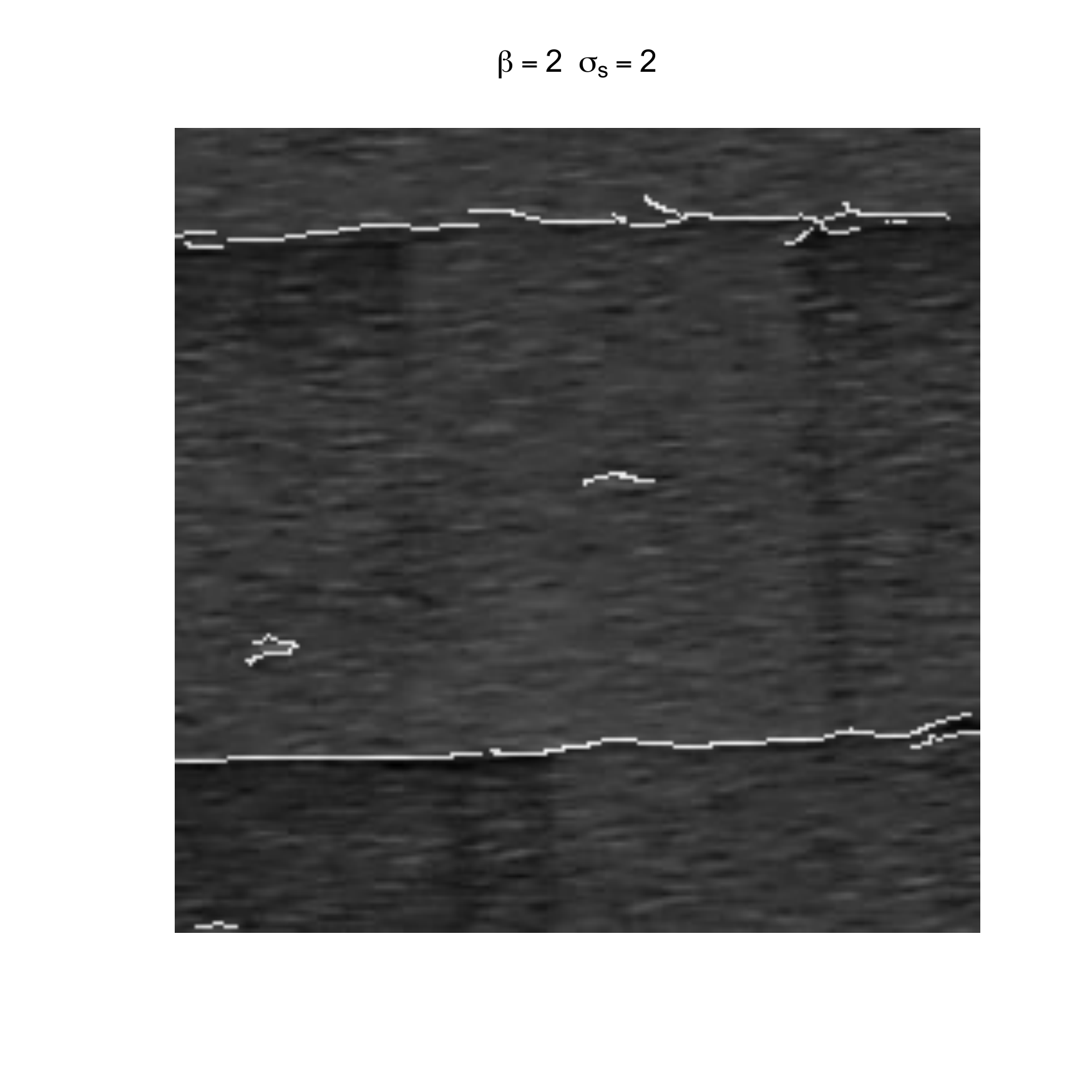}&
\includegraphics[scale = 0.1]{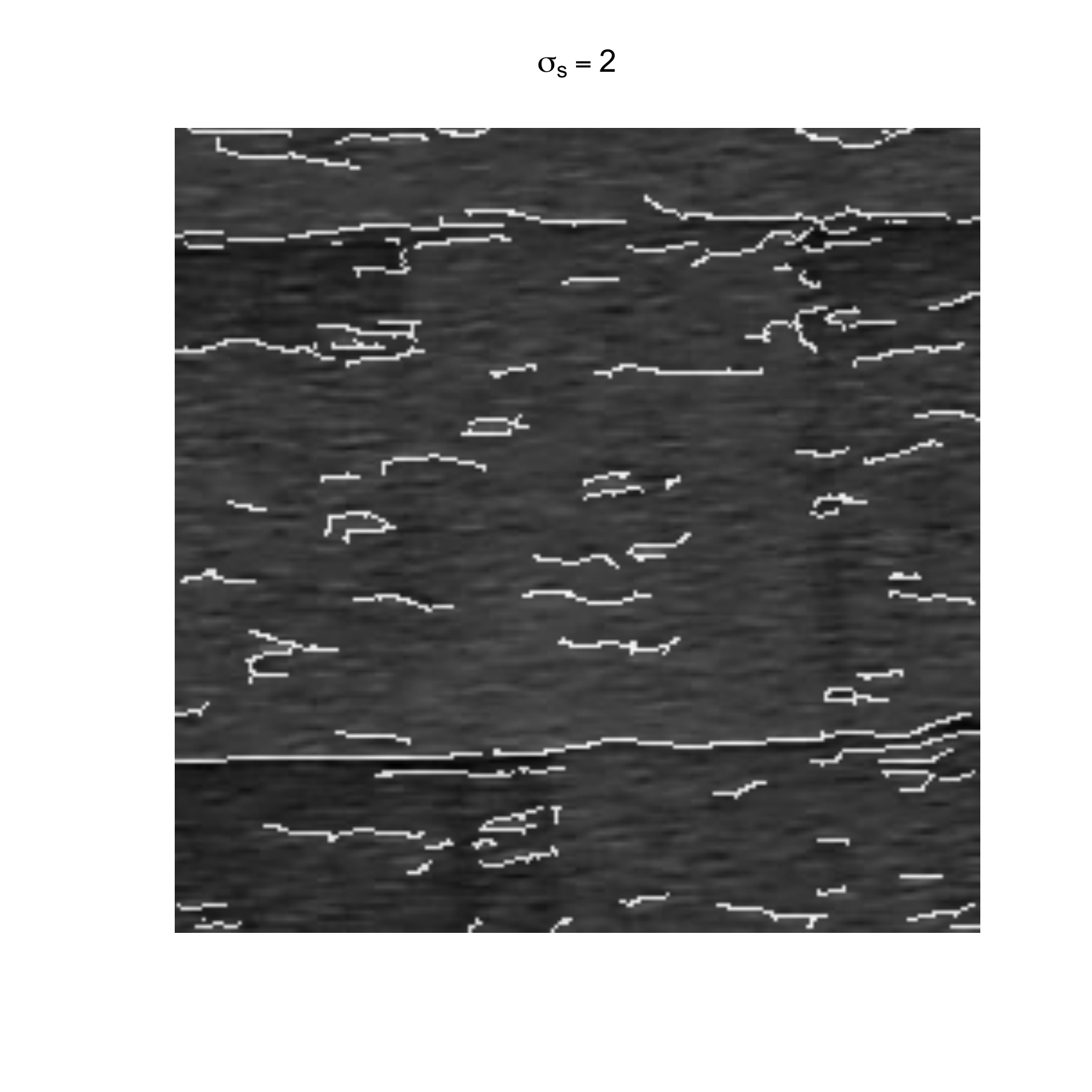}\vspace{-3mm}\\
\includegraphics[scale = 0.1]{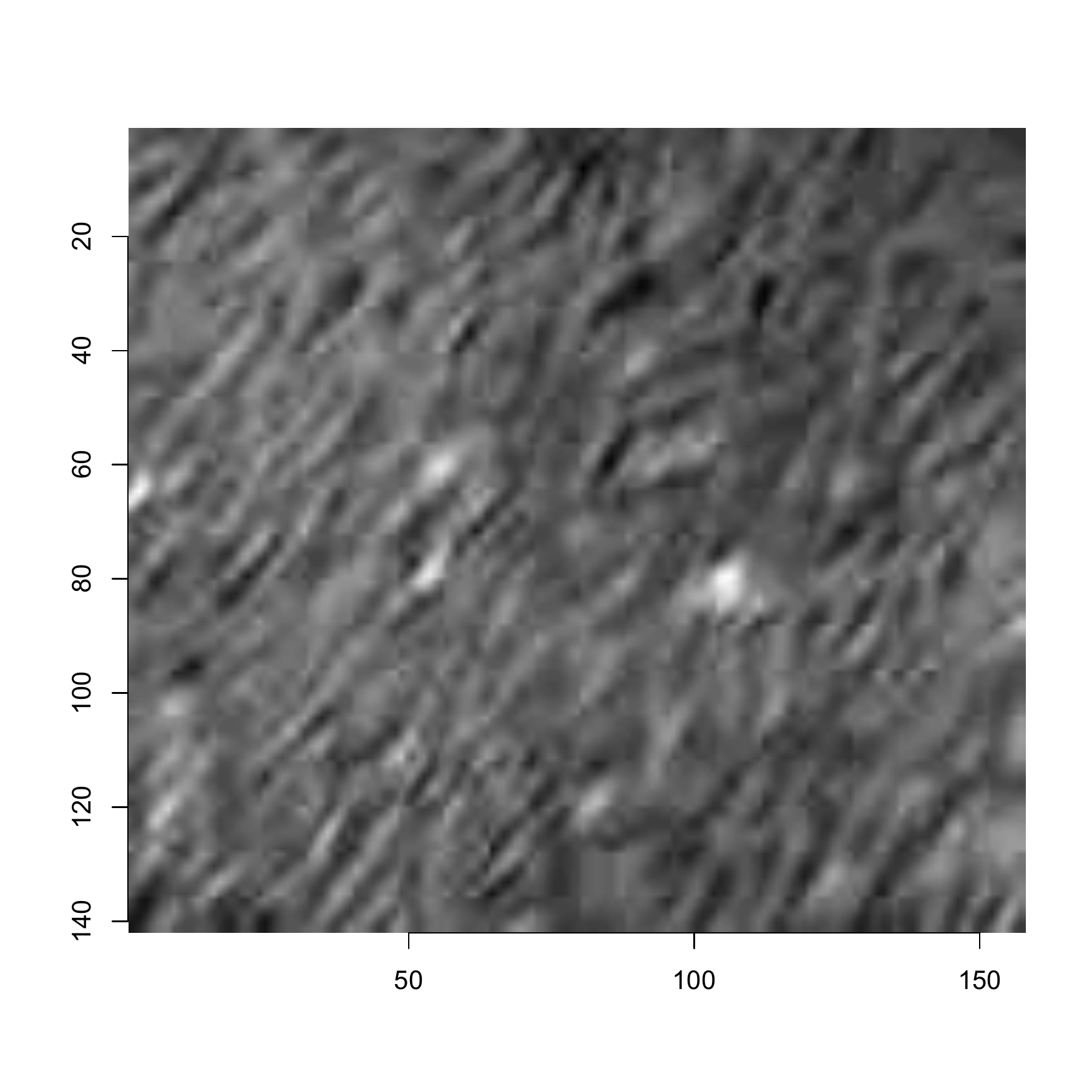}&
\includegraphics[scale = 0.1]{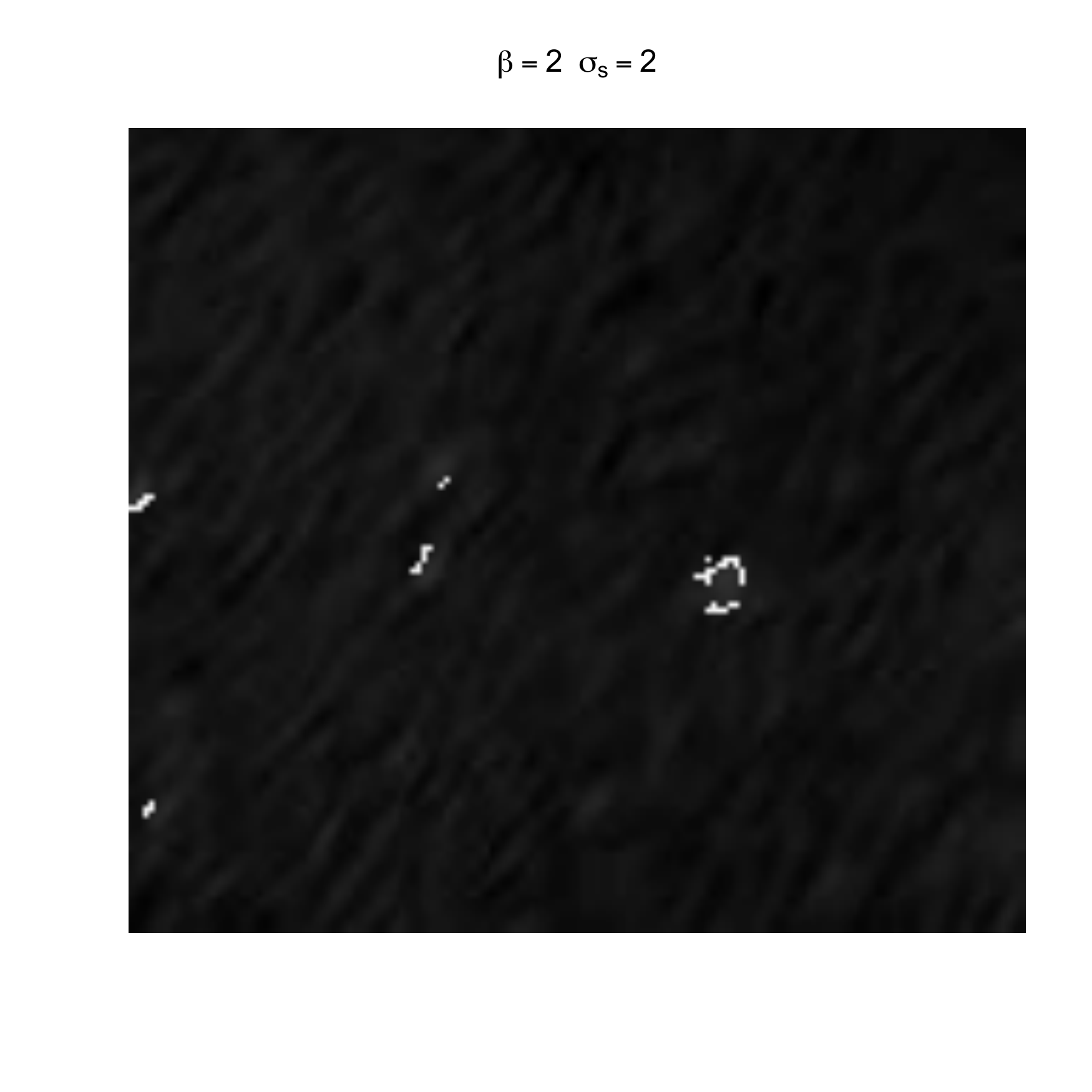}&
\includegraphics[scale = 0.1]{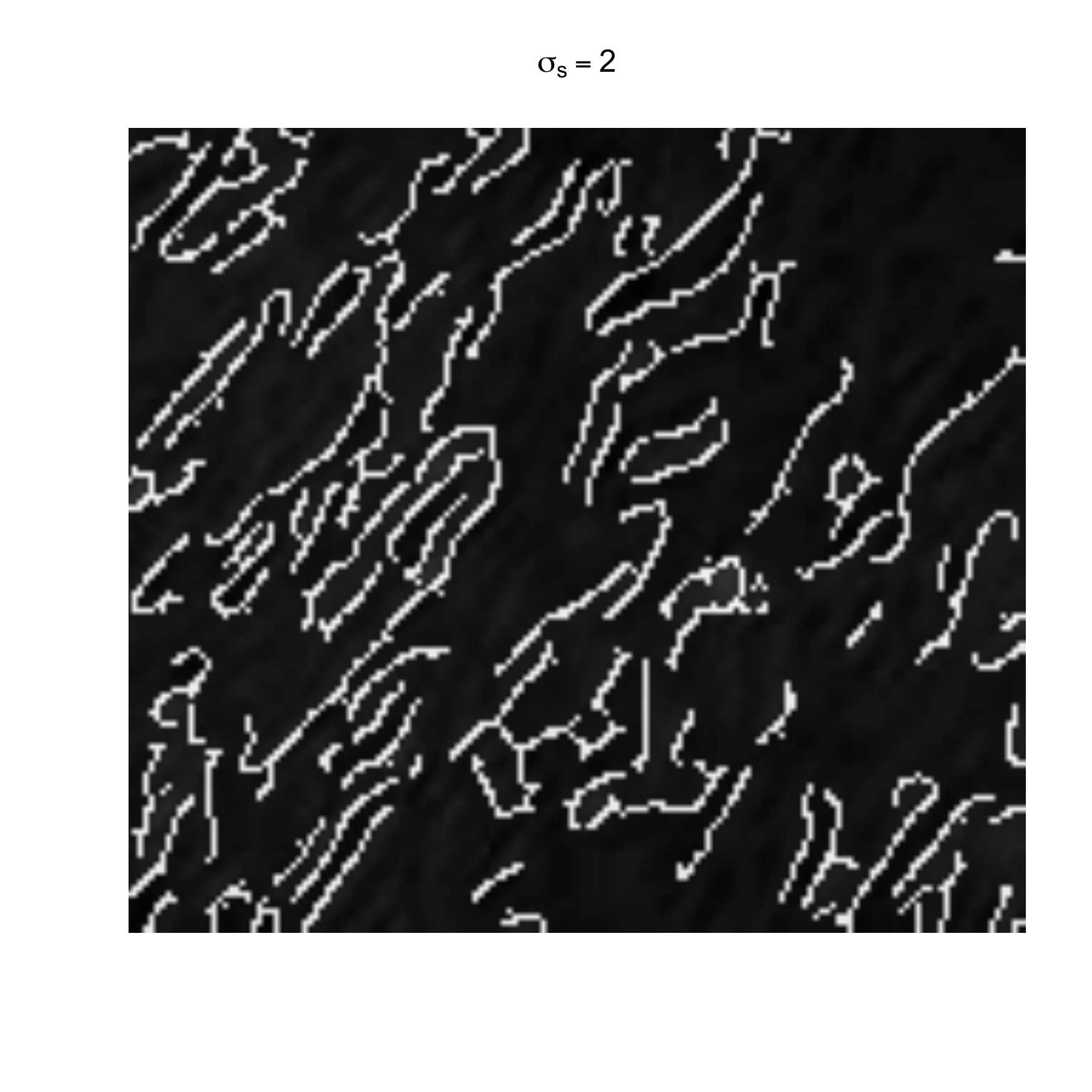}\vspace{-3mm}\\
\includegraphics[scale = 0.1]{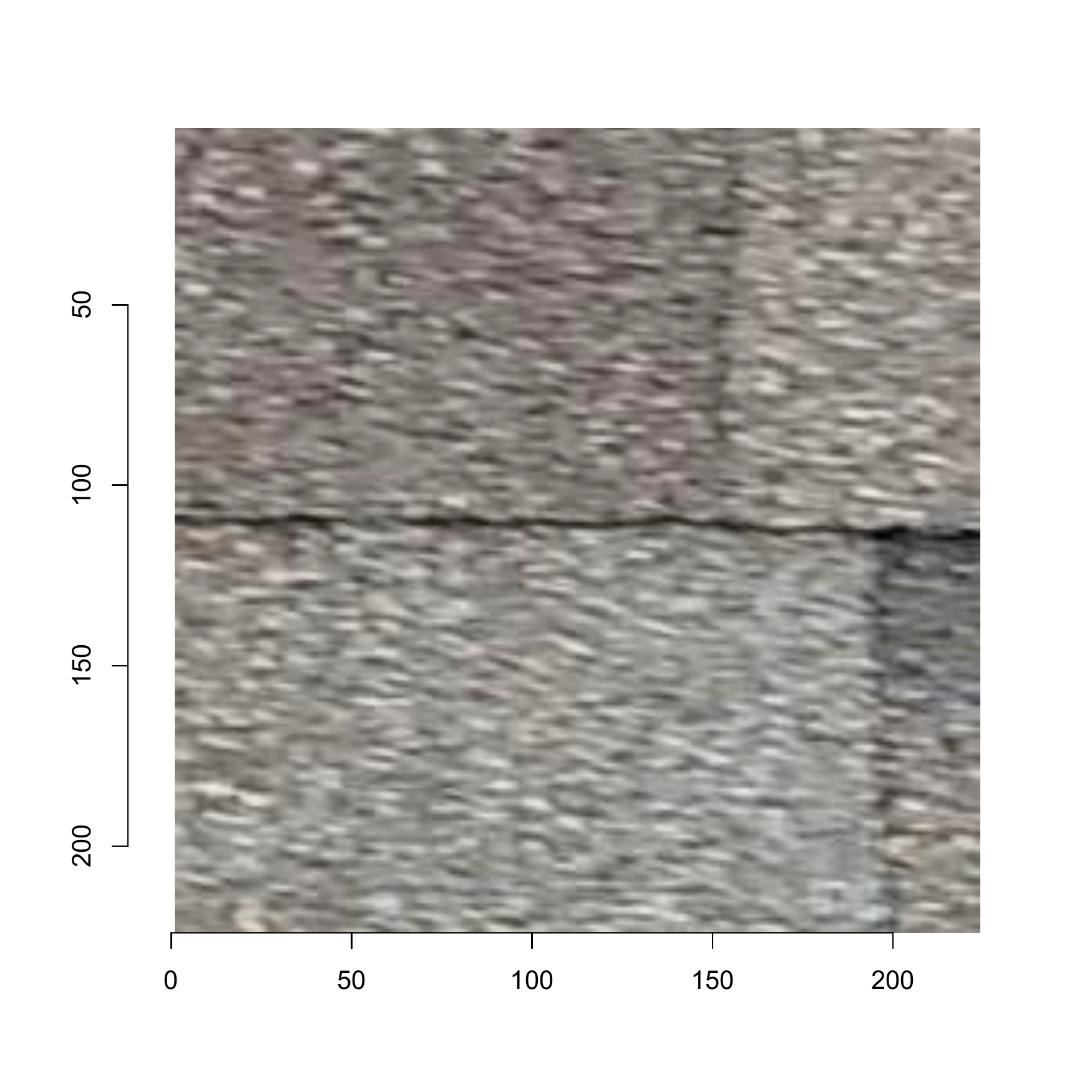}&
\includegraphics[scale = 0.1]{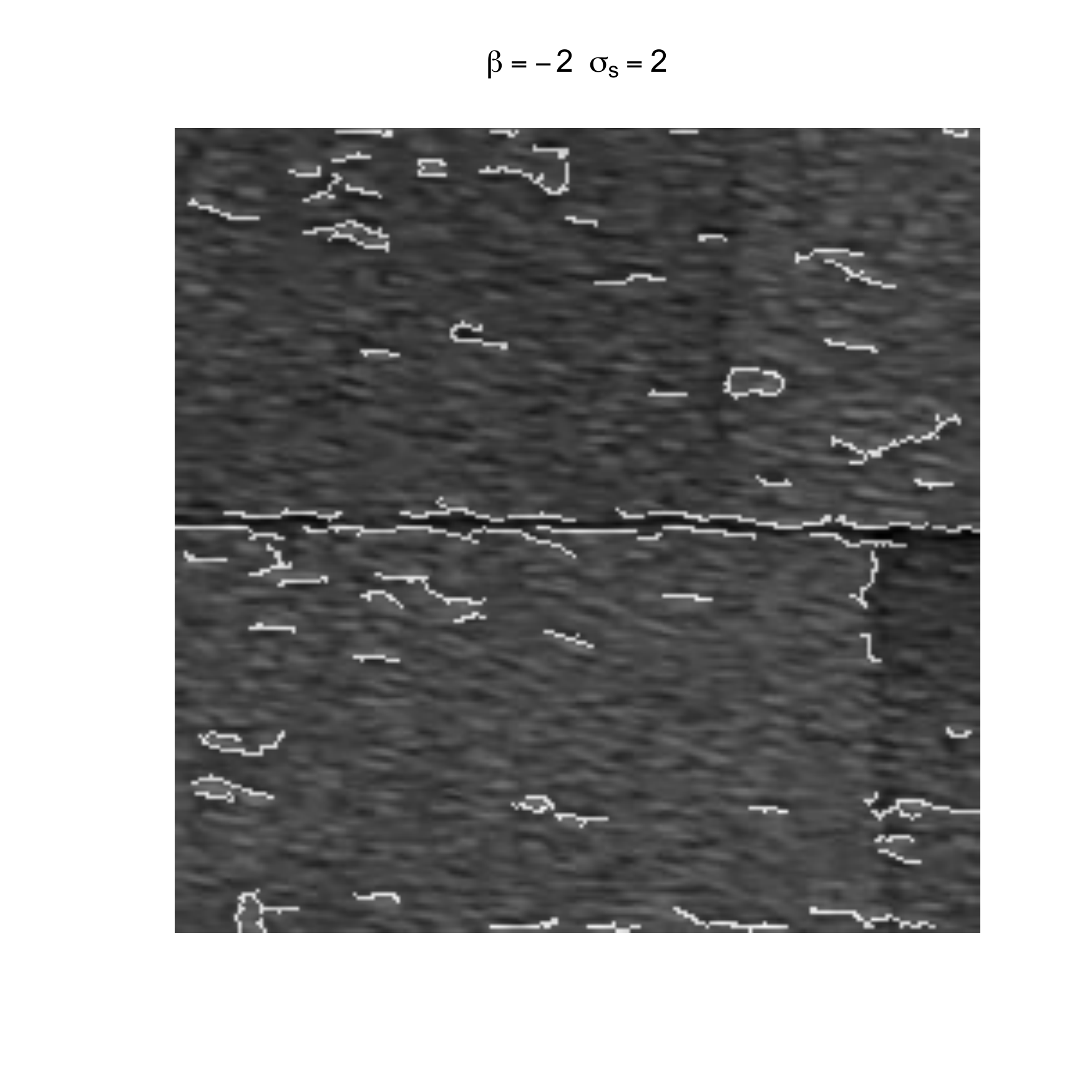}&
\includegraphics[scale = 0.1]{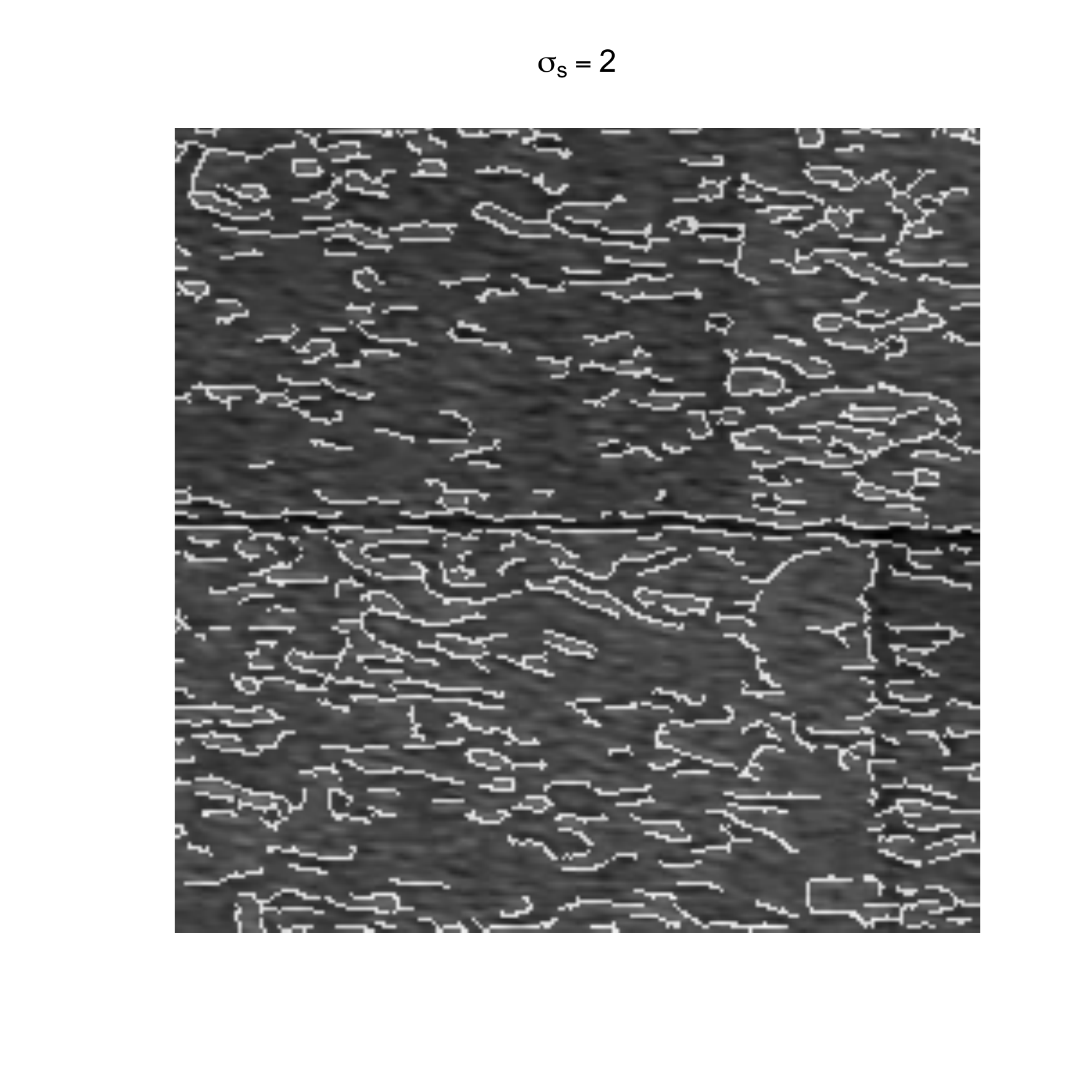}\vspace{-3mm}
\end{tabular}
\caption{\footnotesize The defect detection for the motion blurred roof images. The
  original images are in the first column, and the edges are overlaid
  with the original images in the second and  the
  third columns.  From the top to the bottom, we choose $\beta = 2, 2, -2$ respectively
and $\sigma_s = 2$ in all settings, where $\sigma_s$ is the Gaussian smoothing standard
  deviation.}\label{fig:nodm}
\end{figure}
Controlling the false discovery rate is particularly
important for the drone image analysis. Because the drones take pictures in motion,
some images are inevitably  blurred due to the unstable
movements in the roof inspection process. These motion blurs
exacerbate
the appearances of the surfaces, and in turn increase the difficulties
of suppressing the non-edge pixels. To see that, we implement the Canny detector on
the three blurred non-damage shingles. The right panel in Figure \ref{fig:nodm} shows
that the Canny detector is substantially  misled by the motion
blurs, while the proposed Bayesian detector (will be discussed later) is robust in removing the
non-edge pixels for the blurred images.
It is worth mentioning that for the insurance companies, the
false discovery is a more serious problem than the false non-discovery
of the damages, because no insurance company is willing to cover the
costs for ``fixing'' no damage areas.

\begin{figure}[!h]
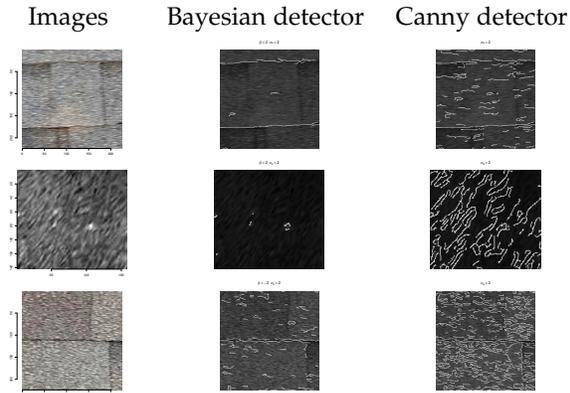

\centering
\begin{tabular}{ccc}
Images & Bayesian detector& Canny detector\\
\includegraphics[scale = 0.1]{figures/nreal1.pdf}&
\includegraphics[scale = 0.1]{figures/nreal1b.pdf}&
\includegraphics[scale = 0.1]{figures/nreal1c.pdf}\vspace{-3mm}\\
\includegraphics[scale = 0.1]{figures/nreal4.pdf}&
\includegraphics[scale = 0.1]{figures/nreal4b.pdf}&
\includegraphics[scale = 0.1]{figures/nreal4c.pdf}\vspace{-3mm}\\
\includegraphics[scale = 0.1]{figures/nreal2.pdf}&
\includegraphics[scale = 0.1]{figures/nreal2b.pdf}&
\includegraphics[scale = 0.1]{figures/nreal2c.pdf}\vspace{-3mm}\\
\end{tabular}
\caption{The edge detection for the motion blurred roof images. The
  original images are in the first column, and the edges are overlaid
  with the original images in the second and  the
  third columns.  From the top to the bottom, we choose $\beta = 2, 2, -2$ respectively
and $\sigma_s = 2$ in all settings, where $\sigma_s$ is the Gaussian smoothing standard
  deviation.}\label{fig:nodm}
\end{figure}

% To explore the reason of the failure cases, we briefly describe
% the Canny procedure. Assume the intensity of the image is $f(x, y) =
% \nu(x, y) + \epsilon(x, y)$ and $\epsilon(x, y)$ is a mean 0 random
% error with variance $\sigma^2$.  In the Canny procedure, the first step is a local
% smoothing by taking the weighted averaging over $m$ neighborhood of
% $(x, y)$. For simplicity, we assume the uniform weights so that the
% smoothing result $h(x, y)$ is the local mean surrounding $(x, y)$. The procedure then finds
% \bse
% H(x, y) = \sqrt{\partial h(x, y)/\partial x^2 + \partial h(x, y)/\partial y^2}.
% \ese
% Ideally, the points on/off the edge would have high/low value of $H(x, y)$,
% so that the algorithm can identify the edges by selecting $(x, y)$'s
% with high $H(x, y)$ values.    In the discrete settings, $\partial h(x, y)/\partial
% x = h(x - 1, y - 1) -  h(x  + 1, y- 1) + 2 \{ h(x - 1, y) -  h(x  + 1,
% y)\} +  h(x - 1, y + 1) -  h(x  + 1, y)$ and  $\partial h(x, y)/\partial
% x = h(x - 1, y - 1) -  h(x  - 1, y + 1) + 2 \{ h(x, y - 1) -  h(x ,
% y +1)\} +  h(x + 1, y - 1) -  h(x  + 1, y+ 1)$.   Let $\mu_x(x, y) = E\{\partial h(x, y)/\partial
% x\}$ and $\mu_y(x, y) = E\{\partial h(x, y)/\partial
% y\}$, $\mu_x(x, y)$, $\mu_y(x, y)$ are the mean differences in the 8 pixel
% neighborhood of $x, y$ in the horizontal and vertical directions,
% respectively.
To explore the reason of the failure cases, we rephrase the edge
detection to a hypothesis testing problem,  where at the
$(x, y)$th pixel,   the null hypothesis is $H_0: \theta_x = \theta_y= 0$ and the alternative hypothesis is $H_1:
\theta_x\neq 0$ or $\theta_y \neq 0$.  Here  $\theta_x$ and $\theta_y$ are the means of the partial difference on the
horizontal $x$ and vertical $y$
directions, respectively.  The Canny method uses the sample
mean difference as the test statistics for evaluating the hypothesis.
However,  this standard frequentist test does not lead to the conclusion
of accepting the null hypothesis, and hence it is expected that the Canny
method is ineffective in eliminating  the non-edge pixels. In contrast,
the Bayesian hypothesis test results
 in the calculation of the posterior probability that the null
 hypothesis is true. Therefore, we resort to the Bayesian  methods  to
 determine whether the specific pixel is a
 non-edge point.

Under the Bayesian paradigm, the Bayes factor is commonly used to
test whether the null hypothesis is true. Commonly, the performance of
the Bayes factors heavily relies  on  the
proper specifications of the prior distributions, particularly Bayes
factors require the priors to be proper (integration of the prior is one). To this end,
various prior distributions have been proposed to facilitate  computation of
the Bayes factors,  including local priors
 \citep{de2001, cano2004, walker2004, moreno2005, casella2012}; the fractional Bayes priors \citep{o1995, o1997, conigliani2000, de2001};
  and the intrinsic priors \citep{berger1996,
    berger1998, berger1999, perez2002}. These
priors assign non-negligible probabilities to regions  consistent
with the null hypothesis,  resulting in the asymmetric accumulation of  evidence in
favor of the true alternative and the  true null
hypothesis.  To address this issue,  \cite{verdinelli1996} and
\cite{rousseau2007} proposed non-local priors, which assign zero densities to the regions corresponding to
the null hypothesis.  \cite{johnson2010} further introduced  the  non-local alterative
priors, which are flexible in specifying the rate at which the prior
approaches to 0. These non-local priors balance the rates of the convergence for
the Bayes factors in  favor of the true null and alternative
hypotheses. \cite{johnson2012} later applied these
non-local priors to the model selection in
high-dimensional settings. In addition,  \cite{rossell2015} recently considered  parameter
estimation with the non-local priors in the high-dimensional settings.

The non-local moment and inverse
moment priors \citep{johnson2010} have been successfully
implemented in the hypothesis testing, model selection, and variable
selection contexts
\citep{johnson2012, shin2015}. These two seemingly unrelated priors share
the same property that the densities approach to zero at the origin. Realizing
this connection,  we propose a general framework in
constructing the no-local priors, namely the reflected non-local priors. The unified framework is more
interpretable and accommodates
both the moment and inverse moment priors. Further, the convergence of
the Bayes factor is fully depicted by a single parameter, which yields a
systematic routine to adjust the Bayes factors.

The reflected non-local prior has wide applications in the hypothesis
testing and variable selection. As discussed earlier, when the null hypothesis is
true, the convergence rate for the Bayes factor to  $0$ largely
depends on the choice of the prior. When the local, non-local moment and inverse
moment prior are selected, the Bayes factors converge to 0 at the
rates of $O_p(n^{-1/2})$, $O_p(n^{-1/2 - \lambda})$, and $O_p[\exp\{-n^{-k/(1
  -k)}\}]$, respectively, with the sample size $n$ and the prior density parameters $\lambda,
k >0$ \citep{johnson2010}.  It is clear
that the non-local priors have faster convergence rates. Furthermore, with
proper choices of parameters, the reflected non-local priors
achieve the same convergence rates as the moment and the inverse moment
priors. In addition to the merits in the tuning and interpretation, the reflected
non-local prior is a superior choice in the hail edge detection
problem.

The rest of the article is organized as the following.  We introduce
the reflected non-local prior in the general setting, derive their asymptotic properties, and investigate their
finite sample properties in Section \ref{sec:ref}.  In Section
\ref{sec:edgesimu}, we investigate their application in the edge
detection through simulation and hail damage detection data
analysis. We conclude with some discussion in Section
\ref{sec:con}.  The theoretical derivations are presented in the
Appendix.

\section{Reflected non-local priors }\label{sec:ref}
\subsection{A truncated reflected non-local prior}
Let $\X = (\X_1\trans, \ldots,\X_n\trans)\trans $ denote a $d$-dimensional random
variable with the likelihood
$$p_n(\X|\btheta) = \prod_{i = 1}^n f_X(\X_i
|\btheta),$$
where $f_X$ is the density function of $\X_i$ and $\btheta$ is
the parameter of interest, $\btheta\in \Theta
\subset \mathbb{R}^d$.
Under the Bayesian paradigm, we define the null and alternative
hypotheses as
$$H_0: \btheta \sim \pi_0(\btheta) \quad {\rm versus} \quad H_1:
\btheta\sim\pi_1(\btheta),$$
where $\pi_0(\btheta)$ and $\pi_1(\btheta)$ are
the priors of the parameter $\btheta $ under the null and alternative
hypotheses, respectively.  The marginal density with the prior
$\pi_j(\btheta)$ is given by
$$m_j(\X) = \int_{\Theta} p_n(\X |\btheta)
\pi_j(\btheta) d\btheta,$$
for $j=0,1$.  The Bayes factor based on a sample of size $n$
is defined as
\be\label{eq:BF}
{\rm BF} = m_1(\X)/m_0(\X).
\ee

If for every $\epsilon>0$,  there is $\xi
>0$ such that $\pi(\btheta) <\epsilon$ for all $\btheta \in \Theta$
with $\inf_{\btheta_0\in \Theta}|\btheta - \btheta_0| <\xi$
\citep{johnson2010}, then $\pi(\btheta)$ is named a non-local
prior. For ease of exposition,  we discuss the setting when
$d=1$; that is,  $\btheta$ is a scalar, denoted by $\theta$. We focus on testing a point null
hypothesis with the null prior $\pi_0(\theta) =
\delta_{\theta_0}(\theta)$,  where $\delta_{\theta_0}$ is a Dirac
measure. Further, we specify the prior
$\pi_1(\theta)$ under the alternative hypothesis to be a non-local prior.

To motivate our non-local prior, first note that if we flip a
bounded density upside down, e.g., a zero-mean normal distribution,
after proper normalization, we can obtain
a density with $0$ mass at the maximum point in the original
density. For example, if we consider a standard normal density $\phi(\cdot)$,
then $\{\phi(0 ) - \phi(\theta)\}I_{\theta\in (a, b)} /\tau$
is a non-local prior for $\theta \in (a, b)$ as it places zero mass at the origin,
where $\tau = \int _\Theta \{\phi(0 ) -
\phi(\theta)\}I_{\theta\in (a, b)} d \theta$ is the normalizing
constant. Considering a generalized normal
density,   we define $$\phi(\theta|\theta_0, \sigma,
\beta ) = \exp[-\{(\theta
-\theta_0)/\sigma\}^{2\beta}], $$ which is the kernel of a generalized
normal density.  When $\beta=1$, it reduces to the kernel of the normal density
with mean $\theta_0$ and variance $\sigma^2$. As a result, we define
the reflected non-local prior
as
\be\label{eq:bf1}
\pi(\theta|\theta_0, \sigma, \beta, a, b) &=& \{\phi(\theta_0|\theta_0, \sigma,
\beta ) -   \phi(\theta|\theta_0, \sigma,
\beta ) \}\nonumber\\
&&\times I(\theta\in (a, b))/\tau,
\ee
where $\tau = \int_a^b \{\phi(\theta_0|\theta_0, \sigma,
\beta ) -   \phi(\theta|\theta_0, \sigma,
\beta )\} d\theta$.

It is worth mentioning that by the Taylor expansion, we can write
\bse
\pi(\theta|\theta_0, \sigma, \beta, a, b) &\propto& \sum_{q =
  1}^{\infty}- \frac{\{-(|\theta
-\theta_0|/\sigma)^{2\beta}\}^{q}}{q !}.
\ese
When $q = 1$, the summand $-(|\theta
-\theta_0|/\sigma)^{2\beta}$ is the main part of \cite{johnson2010}'s
moment prior,  which forces the prior to
be 0 at the hypothetical true value. This suggests that the reflected
non-local prior would have similar performances as
\cite{johnson2010}'s  moment priors when $\beta >0$ in the neighborhood of
$\theta_0$.

One major difference between the reflected non-local prior and
\cite{johnson2010}'s priors lies in the treatment of the tails of the
density.   \cite{johnson2010}  allow the prior density to go to 0 gradually, while we
force the density to be 0 at the truncation endpoints. Note that   the
truncation is induced by $I\{\theta\in (a, b)\}$, which does not
affect the behavior of the density around $\theta_0$. Therefore,  if the parameters
are unbounded, we can use a smooth function
$g(\theta)$ in replacing the indicator function. This yields a
generalized reflected prior as discussed in the next section.

\subsection{A generalized reflected prior}
We generalize the definition of the reflected non-local prior so
that it mimics the performance of the moment prior and inverse moment prior
in \cite{johnson2010}. To cover broader cases, we allow
$\beta$  to be negative in the definition of the kernel; that is,
$$\phi(\theta|\theta_0, \sigma,
\beta ) = \exp[-\{(\theta
-\theta_0)/\sigma\}^{2\beta}], \quad \beta \in \mathbb {R}.$$
Hence, our generalized reflected non-local prior is
\be\label{eq:bf2}
&&\pi_g(\theta|\theta_0, \sigma, \beta, g)\nonumber \\
&=& \{\phi(\theta_0|\theta_0, \sigma,
\beta ) - \phi(\theta|\theta_0, \sigma,
\beta )\}g(\theta)/ \tau_g,
\ee
where $\tau_g$ is the normalizing constant.
Note that  $\beta
<0$ leads to $\tau_g<0$. Hence, although  $\{\phi(\theta_0|\theta_0, \sigma,
\beta ) - \phi(\theta|\theta_0, \sigma,
\beta )\} < 0 $ when $\theta \neq \theta_0$, the normalizing constant
$\tau_g$ helps to retain the positive sign of $\pi_g(\theta|\theta_0, \sigma,
\beta, g)$. Further, to allow various tail behaviors,
instead of truncating the distribution, we use a generic function $g(\theta)$ to force the density to be $0$ when
$\theta \to \infty$.  On the
other hand, to preserve the property of the non-local prior, we require
$\{\phi(\theta_0|\theta_0, \sigma,
\beta ) - \phi(\theta|\theta_0, \sigma,
\beta )\}g(\theta) \to 0$ in the
neighborhood of $\theta_0$.

When $\btheta$ is a $d$-dimensional vector,  we define the multi-dimensional reflected non-local
prior as
\be\label{eq:mgnonlocal}
\pi_g (\btheta|\btheta_0, \bSig, \beta, g) &=& \{\phi (\btheta_0|\btheta_0, \bSig,
\beta ) \nonumber \\
&&-   \phi (\btheta|\btheta_0, \bSig,
\beta ) \}g(\btheta)/\tau_g,
\ee
where $\tau_g$ is a normalizing constant, and
$\phi(\btheta|\btheta_0, \bSig,
\beta ) = \exp[-\{(\btheta - \btheta_0)\trans
\bSig^{-1} (\btheta - \btheta_0)\}^\beta]$ is the generalized
    multivariate Gaussian density.

Following the same arguments as in (8) of \cite{johnson2010}, the Bayes
factor in (\ref{eq:BF}) converges to infinity at the rate of
$O_p\{\exp(n)\}$ under the true alternative hypothesis. On the other
hand, under the true
null hypothesis, we establish its property as follows.
\begin{Th}\label{th:1}
Consider testing the null hypothesis $H_0: \btheta = \btheta_0$, versus
the alternative
hypothesis $H_1: \btheta \sim \pi_g (\btheta|\btheta_0, \bSig, \beta,
g)$, where $\pi_g(\btheta|\btheta_0, \bSig, \beta,
g)$ is the $d$-dimensional  generalized reflected local prior in (\ref{eq:mgnonlocal}).
Under Condition \ref{con:zero}--\ref{con:taubound}, when the null
hypothesis is true, if $\beta >0$,
then ${\rm BF} = O_p(n^{-d/2-\beta})$; if $\beta
<0$, then $\log ({\rm BF}) = C n^{-\beta
  /(1 - \beta)}$ for some $C <0$.
\end{Th}
From the theorem, we conclude that when $\beta >0$ and $\beta <0$, the generalized
reflected priors have the similar asymptotic behaviors as the moment
and inverse moment priors, respectively \citep{johnson2010}.  Hence,
the generalized reflected construction provides a unified way to
define the non-local priors, which accommodates both the moment and inverse moment priors. Further, the
simple structure facilitates the systematic investigation on their finite sample
performances.

\subsection{Finite sample properties}
\begin{figure}[!h]
\includegraphics[scale = 0.2]{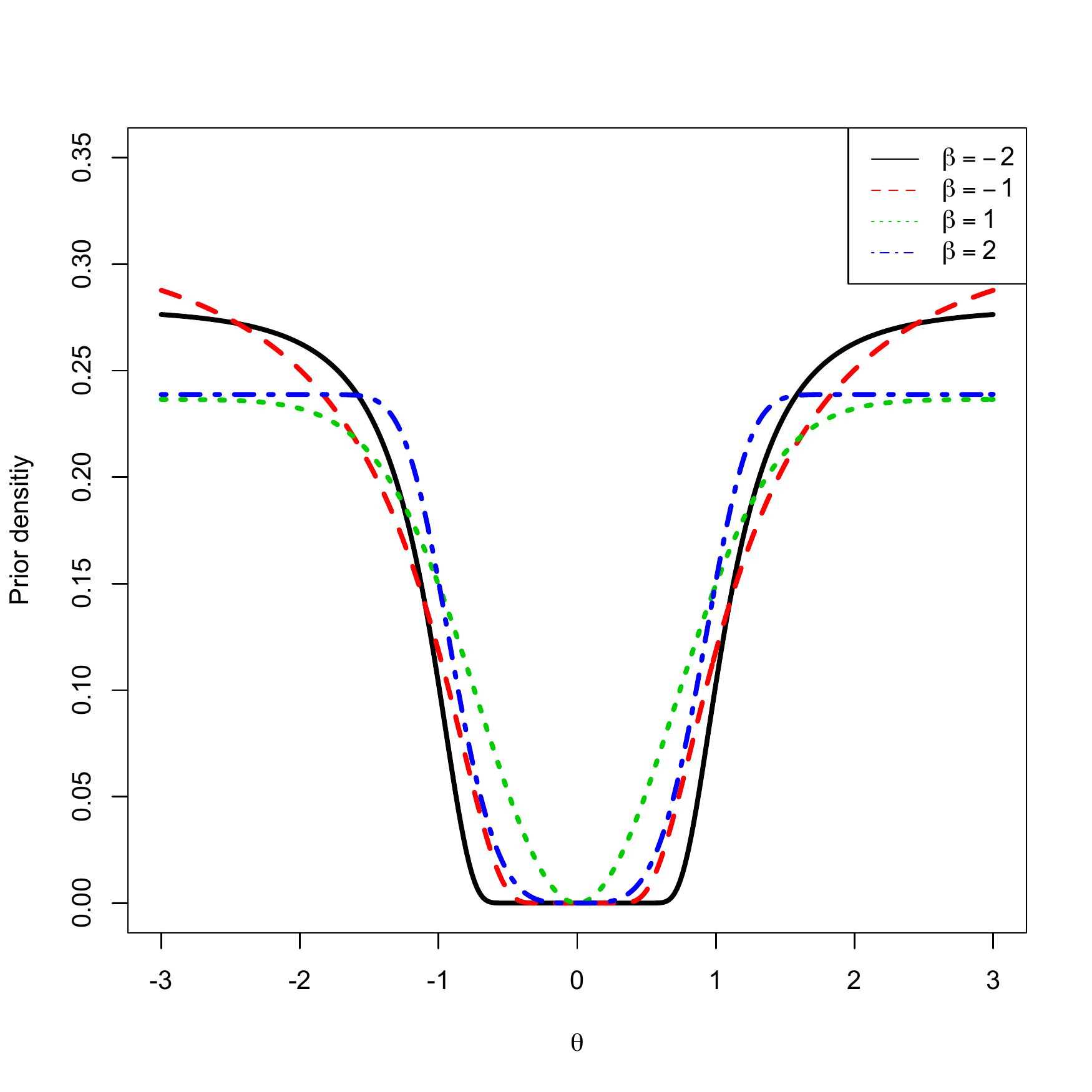}
\includegraphics[scale = 0.2]{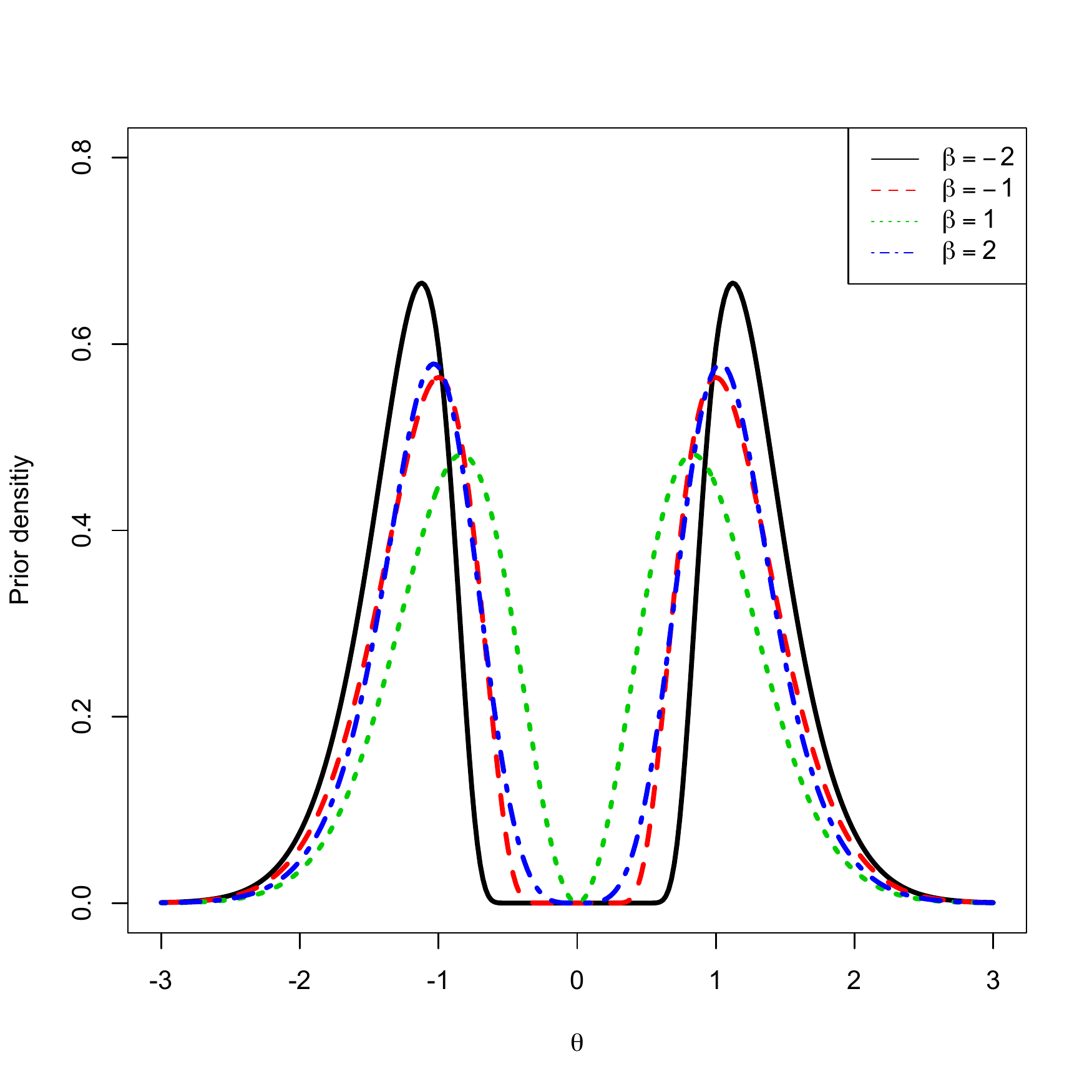}
\caption{The density curves for the  reflected non-local priors (left panel)
  truncated within $(-3,  3)$, and the
  generalized reflected priors (right panel) with $g(\theta) =
  \exp\{-(\theta/\sigma_g)^{2\alpha}\}$, where $\sigma_g  =1$ and $\alpha =
  1$. }\label{fig:shapes}
\end{figure}

In Figure \ref{fig:shapes}, we show the shapes of the  univariate
truncated and generalized reflected
priors. As $\theta$ approaches  the origin, the densities with $\beta <0$
starts the declination
earlier than the ones with $\beta >0$. This explains the phenomenon shown
in Theorem \ref{th:1} that when $\beta <0$ the reflected
prior has a better convergence rate under the null hypothesis.  Further,
the decreasing rate increases with $|\beta|$, which verifies  the
convergence order in  Theorem \ref{th:1}. In addition, by
introducing the smooth function $g(\theta)$, the tails of the density
diminish to $0$ when $\theta$ deviates from the origin.

\begin{figure}[!h]
\centering
\includegraphics[scale = 0.2]{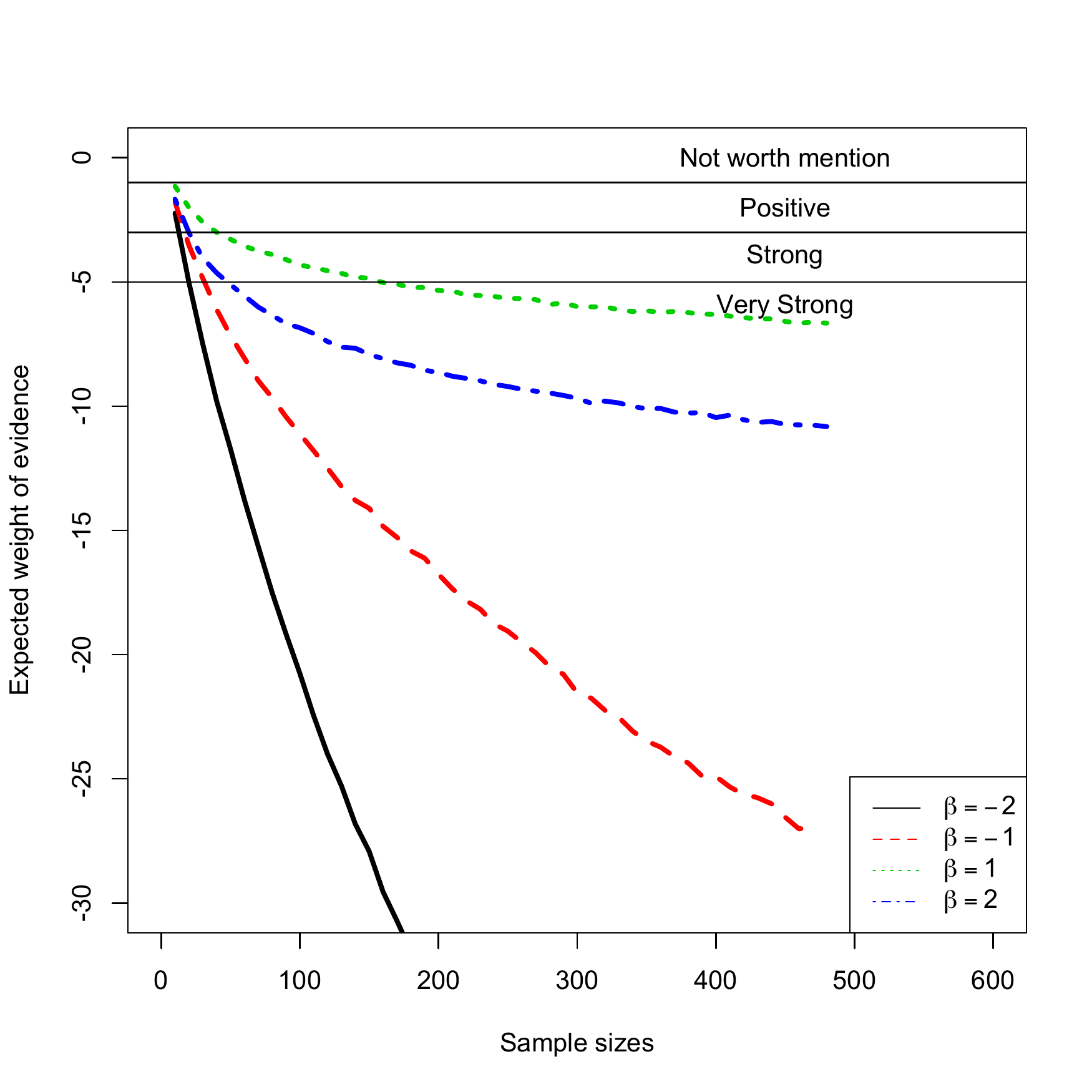}
\includegraphics[scale = 0.2]{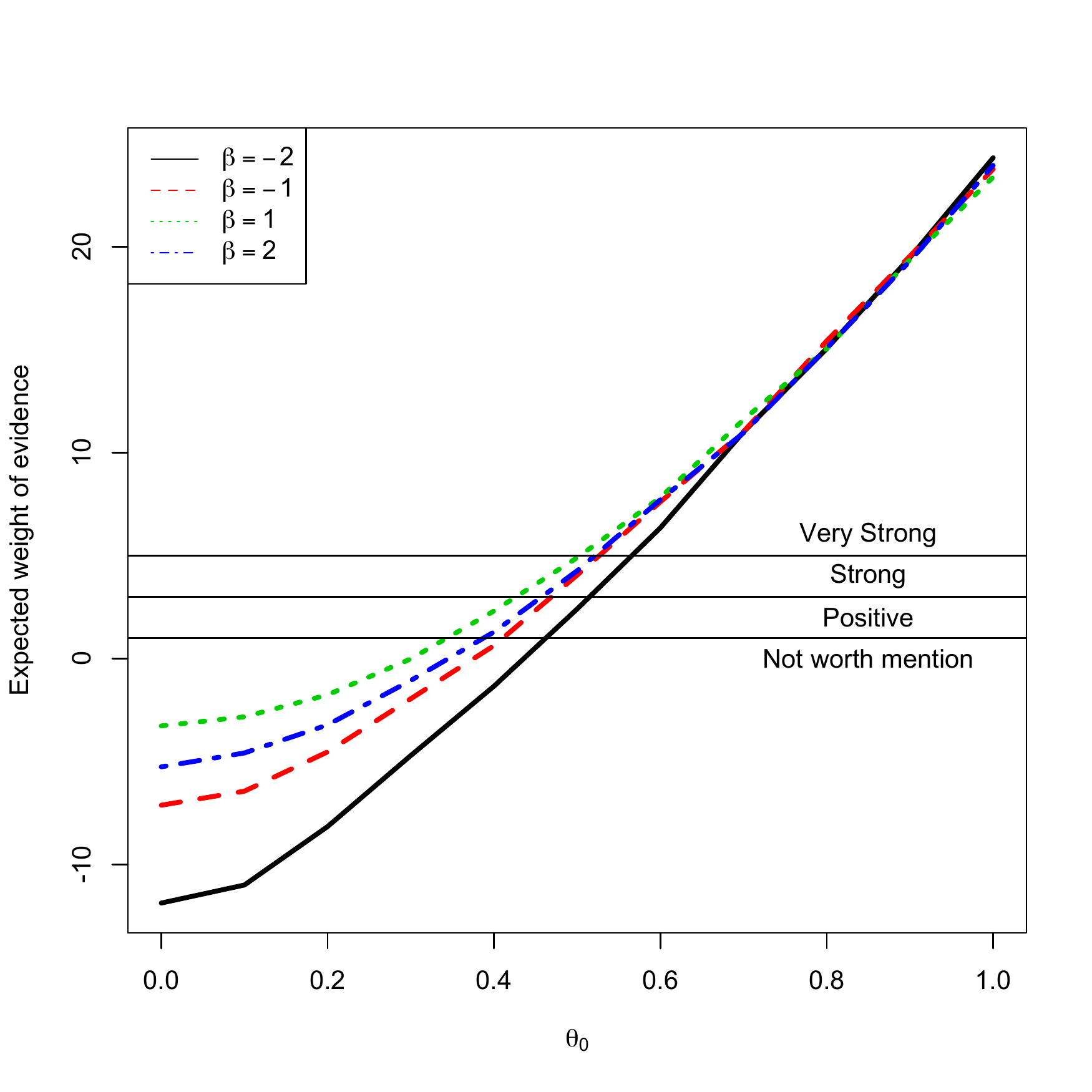}\\
\includegraphics[scale = 0.2]{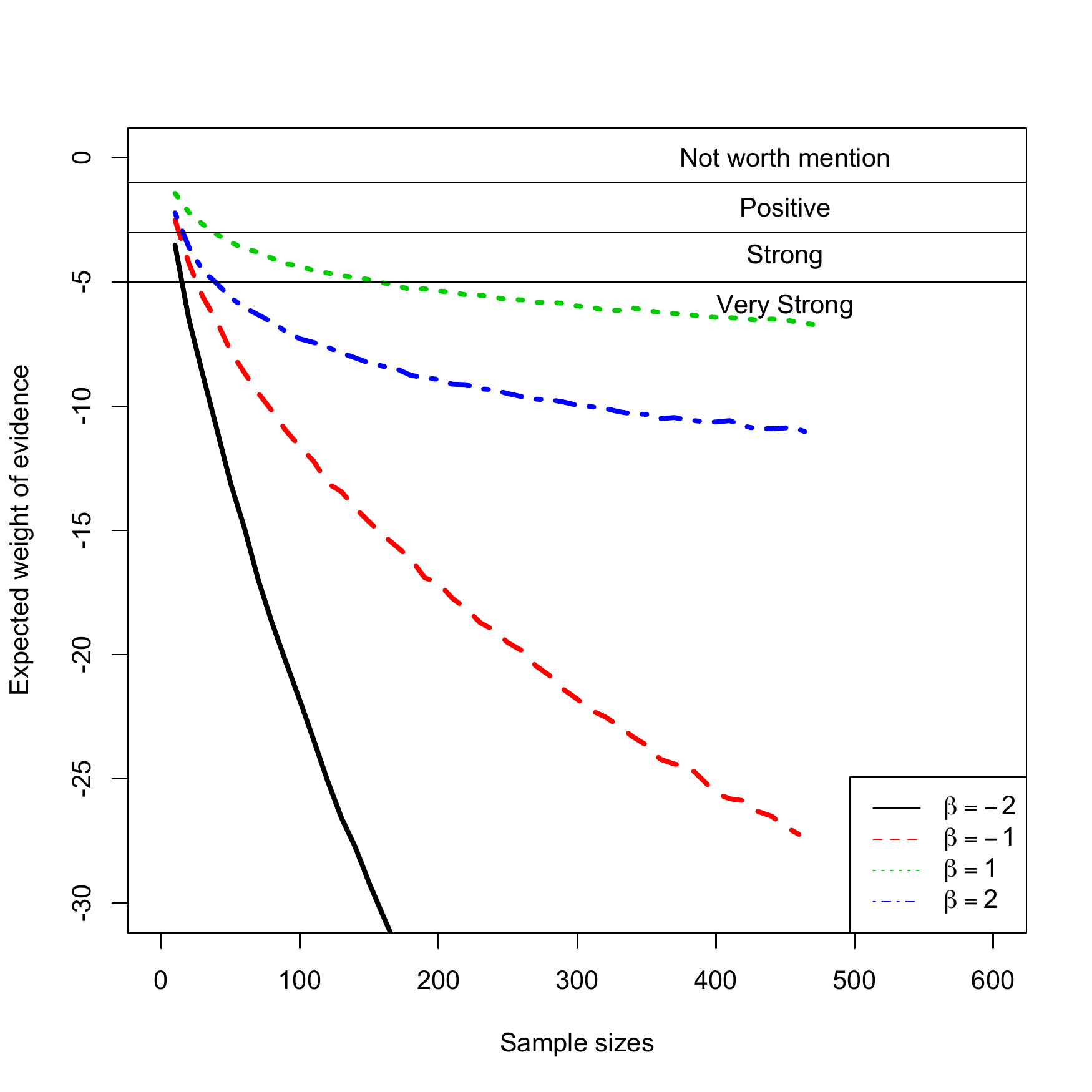}
\includegraphics[scale = 0.2]{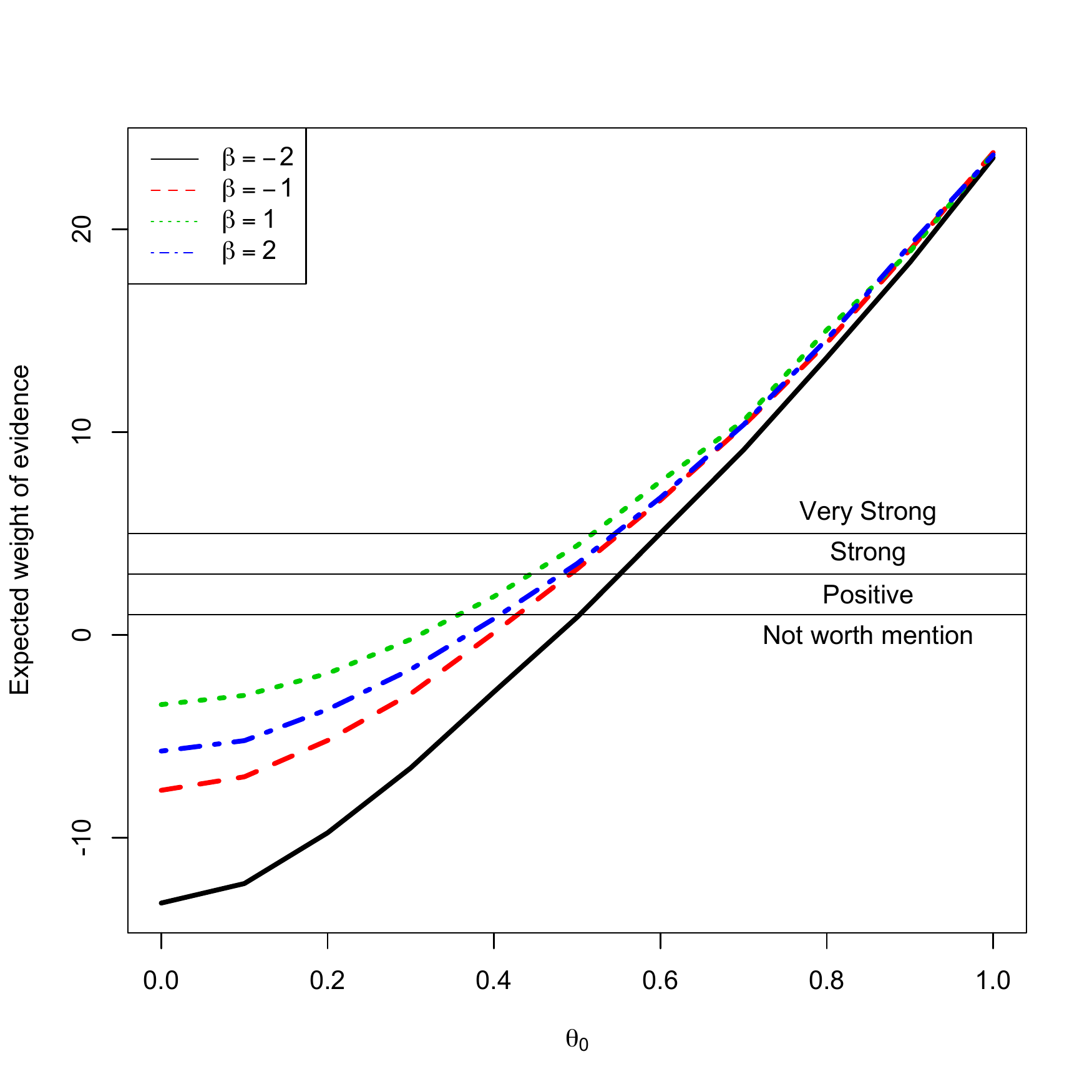}
\caption{The expected weight of evidence over 1000 simulations under
  the true null (left panel) and true
  alternative (right panel) settings for the reflected priors
  (upper panel) truncated within $(-1, 1)$, and the
  generalized reflected priors (lower panel) with $g(\theta) =
  \exp\{-(\theta/\sigma_g)^{2\alpha}\}$, where $\sigma_g  =1$ and $\alpha =
  1$. }\label{fig:weight}
\end{figure}

We further explore the weight of  evidence,  $\log (\rm BF)$,  through simulating
samples from the normal distribution with mean $\theta_0$ and variance
1. We use the priors in (\ref{eq:bf1}) and (\ref{eq:bf2}) with $\sigma = 1$ to
construct Bayes factors.   On the left panel of Figure
\ref{fig:weight}, we show that under
the true null hypothesis that $\theta_0 = 0$, $\log (\rm BF)$ decreases as the sample size increases. When $\beta >0$, the
decreasing rate is relatively slower than the settings with $\beta
<0$, but overall the decreasing is faster for
larger values of $|\beta|$. In the right pannel of  Figure \ref{fig:weight}, we
study  $\log (\rm BF)$ for different values of  $\theta_0$
under the true alternative. When $\theta_0$ is
close to 0,  $\log (\rm BF)$ is larger for the priors with $\beta >0$. In
addition,  $\log (\rm BF)$ decreases with the increase of $|\beta|$. However, for the priors with
$\beta<0$, the corresponding $\log (\rm BF)$ grows faster when $\theta_0$ is sufficiently
large. Overall, by choosing a non-local prior with faster declination
at the origin, we can substantially improve the convergence rate of the
Bayes factor under the
true null hypothesis.
On the other hand, we limit the loss of the convergence within an
acceptable range  under the true alternative
hypothesis.

\section{Bayesian defect detector}\label{sec:edgesimu}
Let $f(x, y)$ be
the smoothed image by the Gaussian filter, for testing whether the
$(x, y)$th pixel is on the defect edge, we collect $2n + 1$ sample
 differences on the $x$-axis; that is,
\bse
\g_x(x, y)  &=& (g_{xj}, j = -n, \ldots, n)\trans\\
&=&\{f(x+  1, y - n ) -  f(x- 1,  y - n ),\\
&&
\ldots, f(x+  1, y + n ) -  f(x- 1, y + n )\}\trans,
\ese and $2n+ 1$ samples on
the $y$-axis; that is,
\bse
\g_y(x, y) &=&  (g_{yj}, j = -n, \ldots, n)\trans \\
&=&
\{f(x-n, y+1 ) -  f(x- n, y-1 ),\\
&&\ldots,
f(x + n, y + 1 ) -  f(x + n, y -1 )\}\trans.
\ese  Further, let $\g_j(x, y)
= \{g_{xj}(x, y), g_{yj}(x, y)\}\trans $, and we assume that $\g_j(x, y)$
follows a bivariate normal distribution with mean
$\btheta = (\theta_x, \theta_y)\trans $ and variance $\bOmega(x,
y)$.
% In practice, $K = 1$ is often selected so that a patch of 9 pixels
% are evaluated each time.
% Note that, because of the first step Gaussian smoothing,
% $\bSig(x, y) = O_p\{(nmh)^{-1}\}$, where  $h$ is the standard
% deviation of the Gaussian kernel, which directly follows the
% property of the Nadaraya--Watson kernel regression estimator
% \citep{nadaraya1964, watson1964}.
% This implies the asymptotic  results in Theorem
% \ref{th:1} holds because when $nmh \to \infty$,  the Laplace
% approximations are still valid.
Under this setting,  the logarithm of the  likelihood for $g_{xj}(x,
y)$ and $g_{yj}(x, y)$, denoted as $L_n\{x, y; \btheta, \bOmega(x,
y)\} $. 
Note that the case of $(x, y)$ not on the
edge corresponds to the null hypothesis $H_0: \btheta = \bf 0$, and $(x, y)$ on the
edge corresponds to the alternative hypothesis $H_1: \btheta \sim
\pi_g (\btheta|\bf 0, \bSig, \beta, g)$, with dimension $d = 2$. Here
$\pi_g (\btheta|\bf 0, \bSig, \beta, g) $ is the prior distribution
at the edge pixels. Then we can write the Bayes
factor as
\bse
{\rm BF}(x, y) = \frac{\int \pi_g (\btheta|{\bf 0}, \bSig, \beta, g)
\exp[L_n\{x, y; \btheta, \bOmega(x, y)\} ] d\btheta}
{\tau_g  \exp[L_n\{x, y; {\bf 0}, \bSig (x, y)\}]}.
\ese

In the implementation, we let $n = 1$ so that a total of 8 surrounding
pixels are selected for each location $(x, y)$.
Further, we estimate $\bOmega(x, y) $ by 2 times the
sample variance of $f(x, y)$, $x
= 1, \ldots, n$, $y = 1, \ldots, m$. After obtaining the Bayes factor
for each pixel, we perform the non-maximum suppression and
thresholding
to thin the edges. In the non-maximum suppression procedure, we keep
the pixels that have the maximum ${\rm BF}(x, y)$ in its $3\times
3$ neighborhood as the potential edge points, and set the rest of
pixels to be the smallest ${\rm BF}(x, y)$  values among all
locations. Then we use the k-means procedure to split ${\rm BF}(x, y)$ to two clusters, and select the maximum value in the cluster
with a lower average ${\rm BF} (x, y)$ as the threshold $\rho$. The edge pixels
are the ones with ${\rm BF}(x, y) >\rho$. For comparison, we implement the standard $3\times 3$
non-maximum suppression, thresholding and edge tracking in the Canny
procedure \citep{canny1986}.  Note that the Canny detector use the
statistics ${\rm CD}(x, y)=  \sqrt{\{\w_x \trans
\g_x(x, y) \}^2  + \{\w_y\trans
\g_y(x, y) \}^2}$ to detect edges, where $\w_x, \w_y$ are weight
vectors.   
%  is  Let $T(x, y)$ be the Canny statistics
% resulting from the
% $3\times 3$
% non-maximum suppression procedure, we split the resulting Canny
% statistics $T(x, y)$ into three
% groups, while selecting the maximal  and minimal value in the groups
% with low $T(x, y)$ and high $T(x, y)$ as the lower and upper
% thresholds, $\rho_l$, $\rho_u$,
% respectively. The pixels with $T(x, y) \geq \rho_u$ are the strong edges
% points, while those with $\rho_l \leq T(x, y) \leq  \rho_u$ are the
% weak edges. To further confirm whether a weak edge point is truly the
% edge point, we implement additional edge tracking procedure

%  is  Let $T(x, y)$ be the Canny statistics
% resulting from the
% $3\times 3$
% non-maximum suppression procedure, we split the resulting Canny
% statistics $T(x, y)$ into three
% groups, while selecting the maximal  and minimal value in the groups
% with low $T(x, y)$ and high $T(x, y)$ as the lower and upper
% thresholds, $\rho_l$, $\rho_u$,
% respectively. The pixels with $T(x, y) \geq \rho_u$ are the strong edges
% points, while those with $\rho_l \leq T(x, y) \leq  \rho_u$ are the
% weak edges. To further confirm whether a weak edge point is truly the
% edge point, we implement additional edge tracking procedure

\subsection{Simulation studies}
\begin{figure}[!h]
\centering
\begin{tabular}{ccc}
\footnotesize  Contaminated Images & \footnotesize  Bayesian detector&
                                                                       \footnotesize  Canny detector\\
\includegraphics[scale = 0.1]{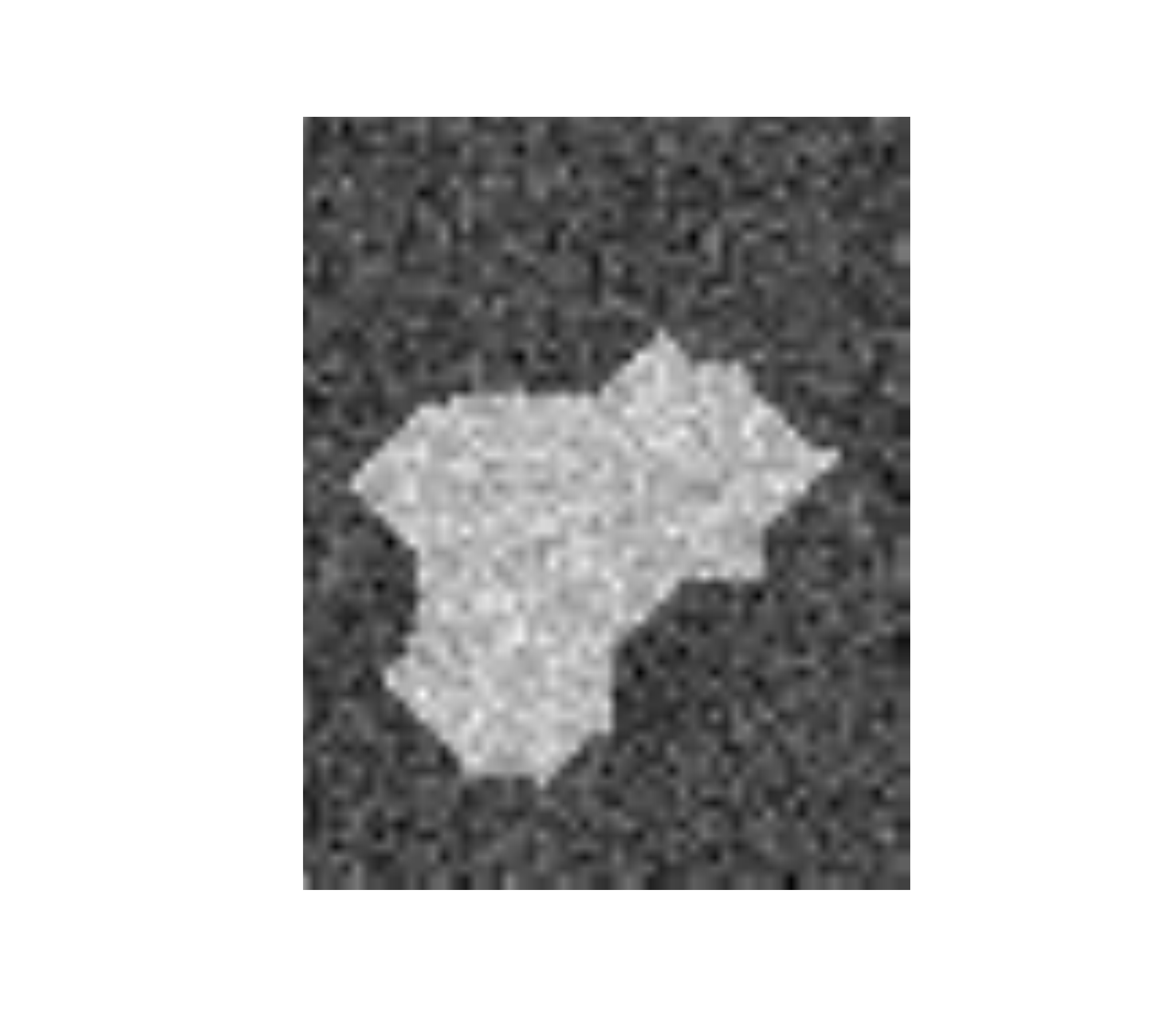}&
\includegraphics[scale = 0.1]{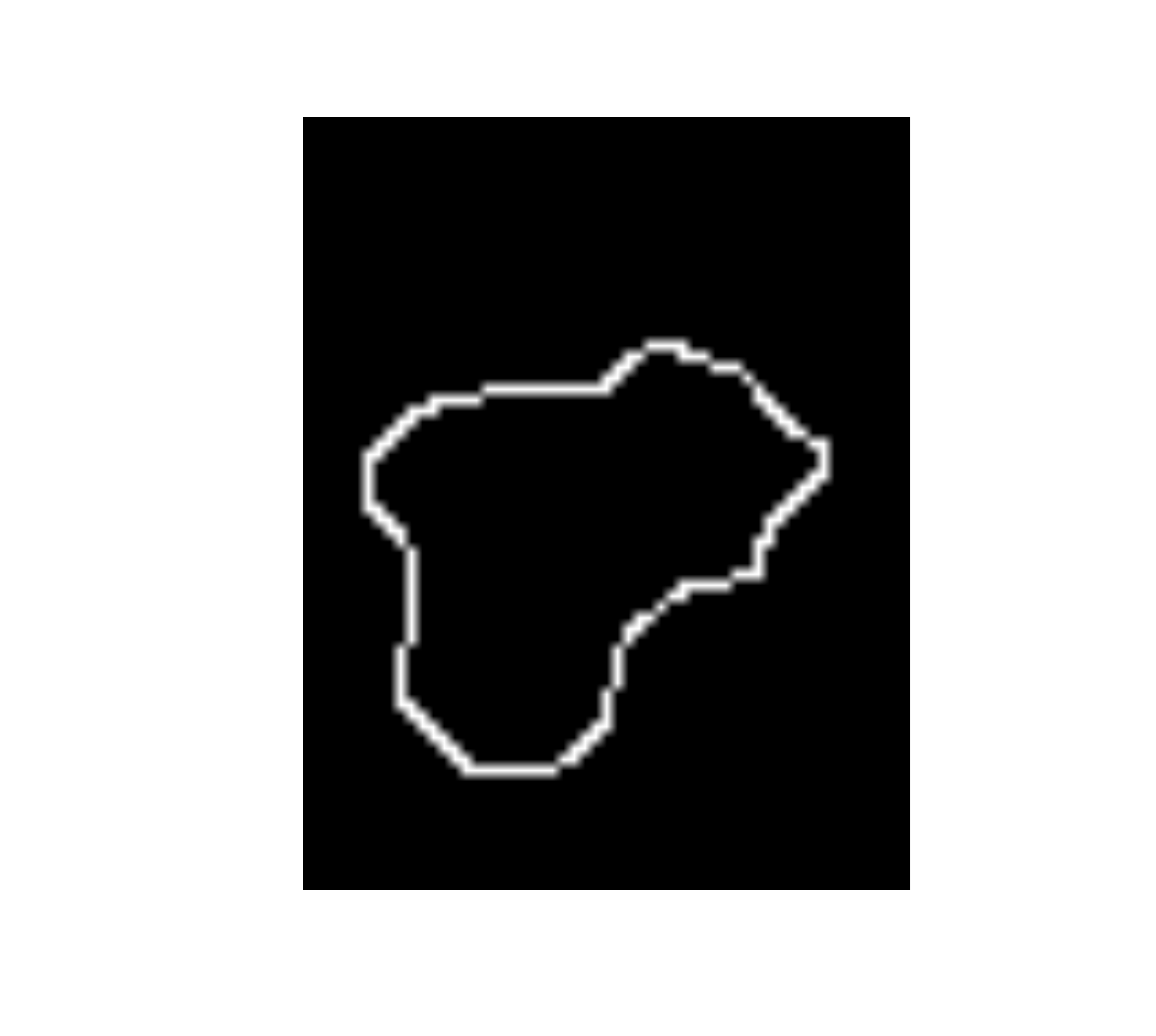}&
\includegraphics[scale = 0.1]{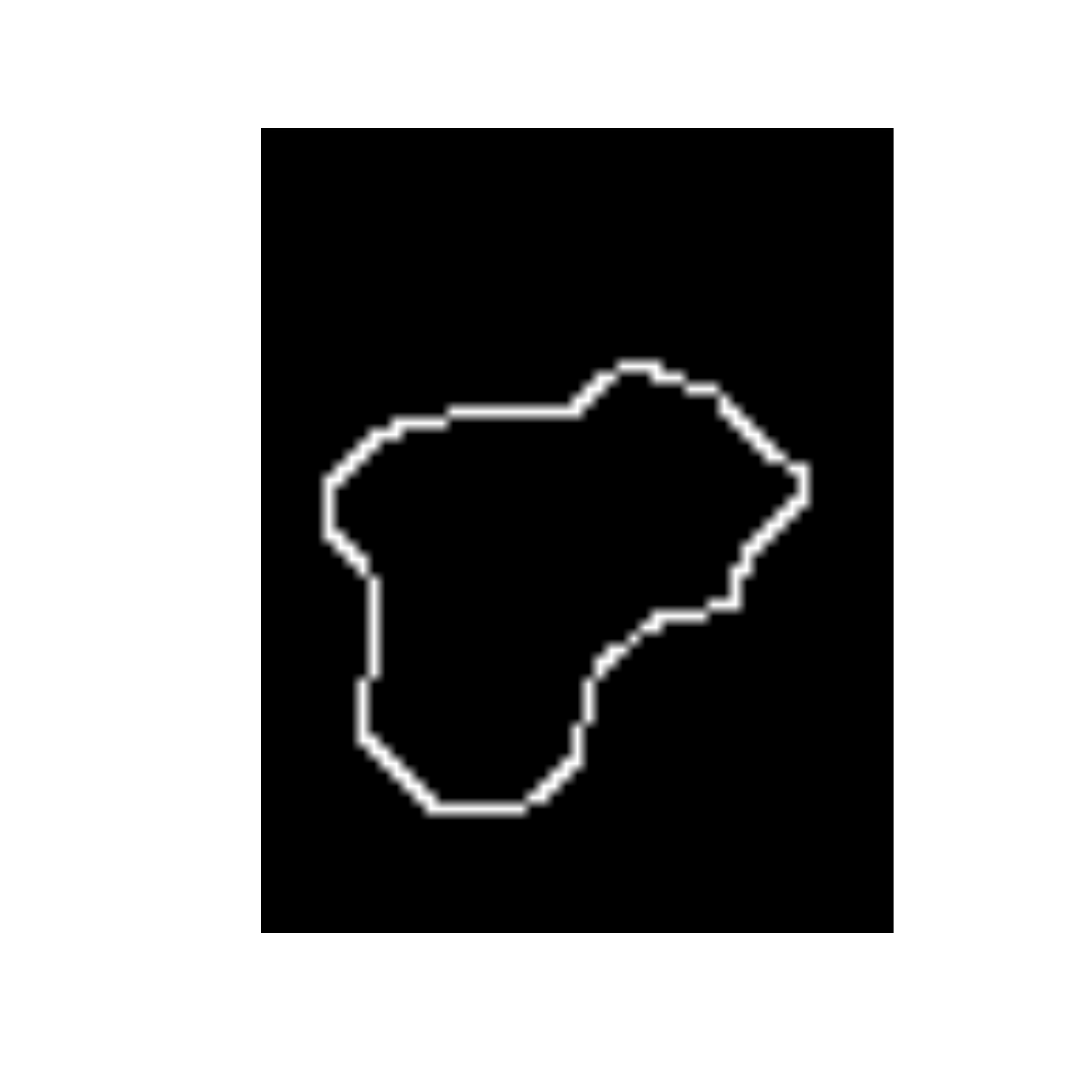}\vspace{-3mm}\\
\includegraphics[scale = 0.1]{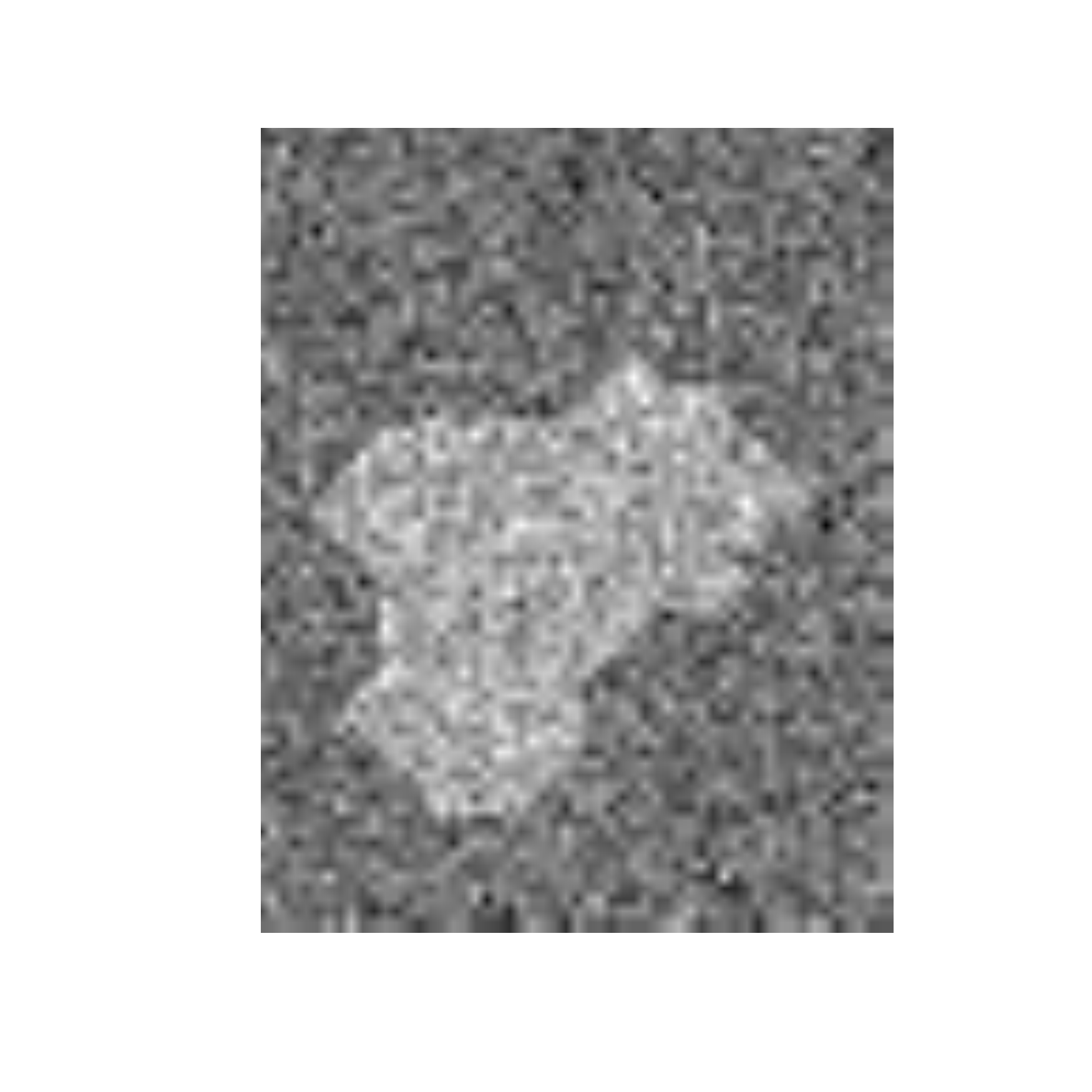}&
\includegraphics[scale = 0.1]{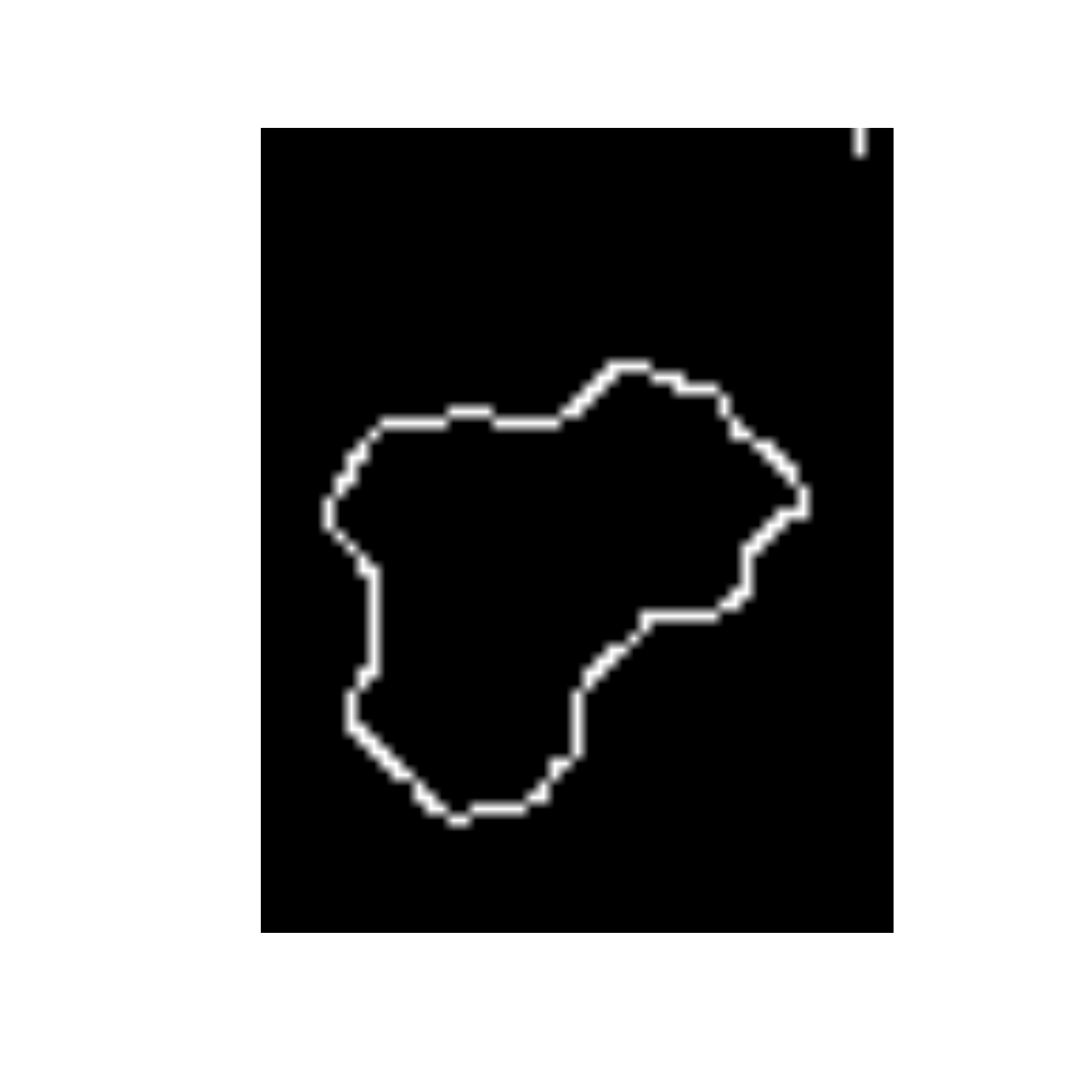}&
\includegraphics[scale = 0.1]{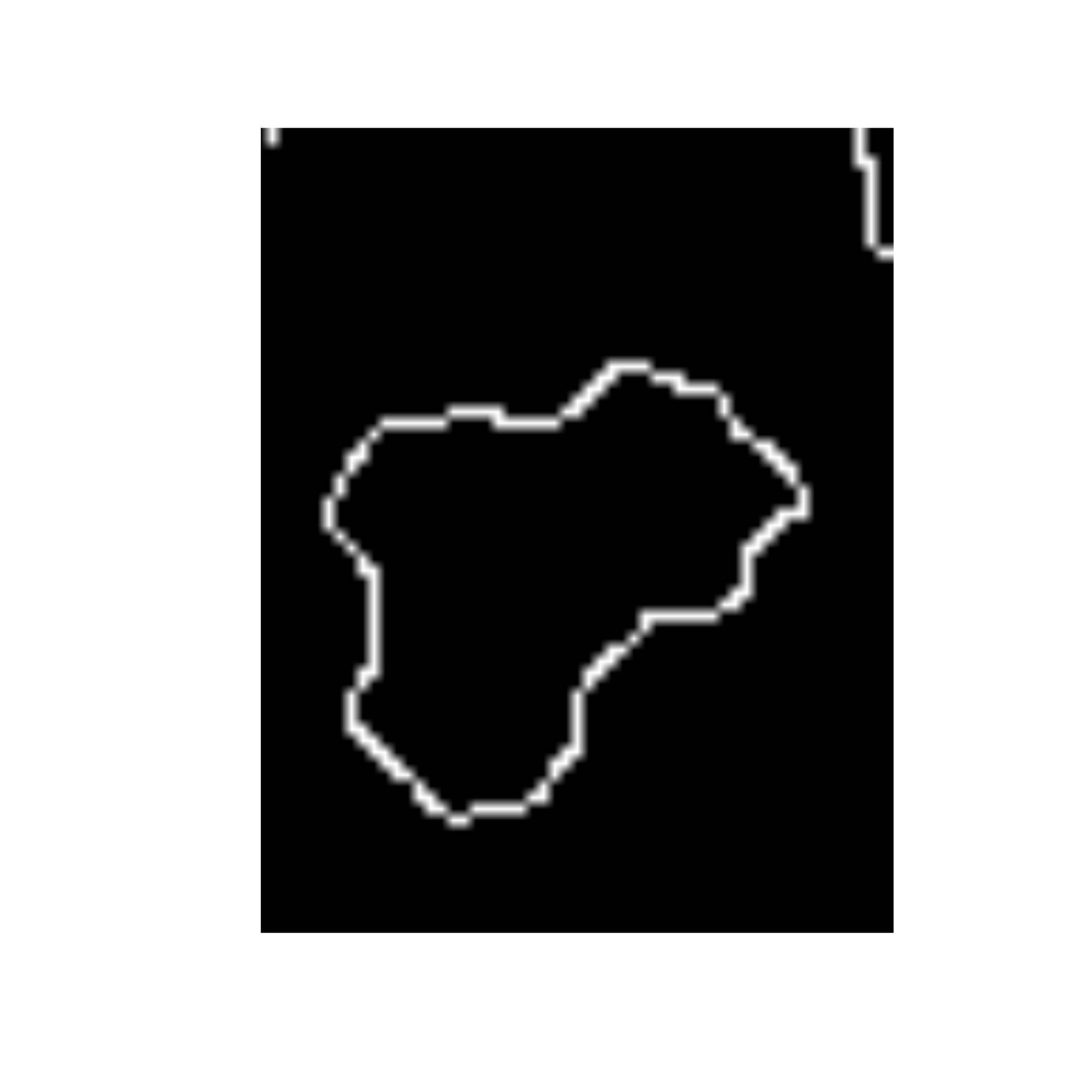}\vspace{-3mm}\\
\includegraphics[scale = 0.1]{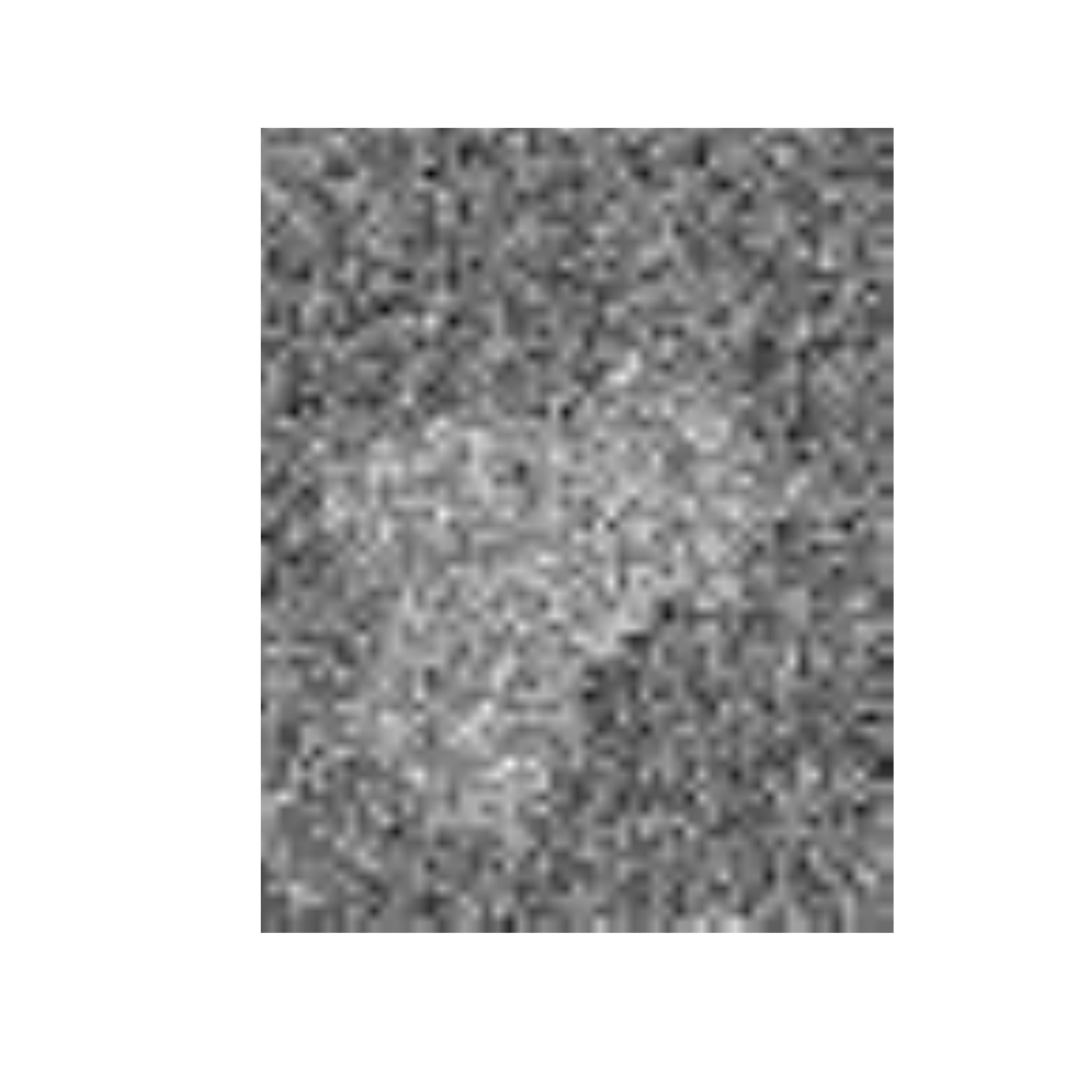}&
\includegraphics[scale = 0.1]{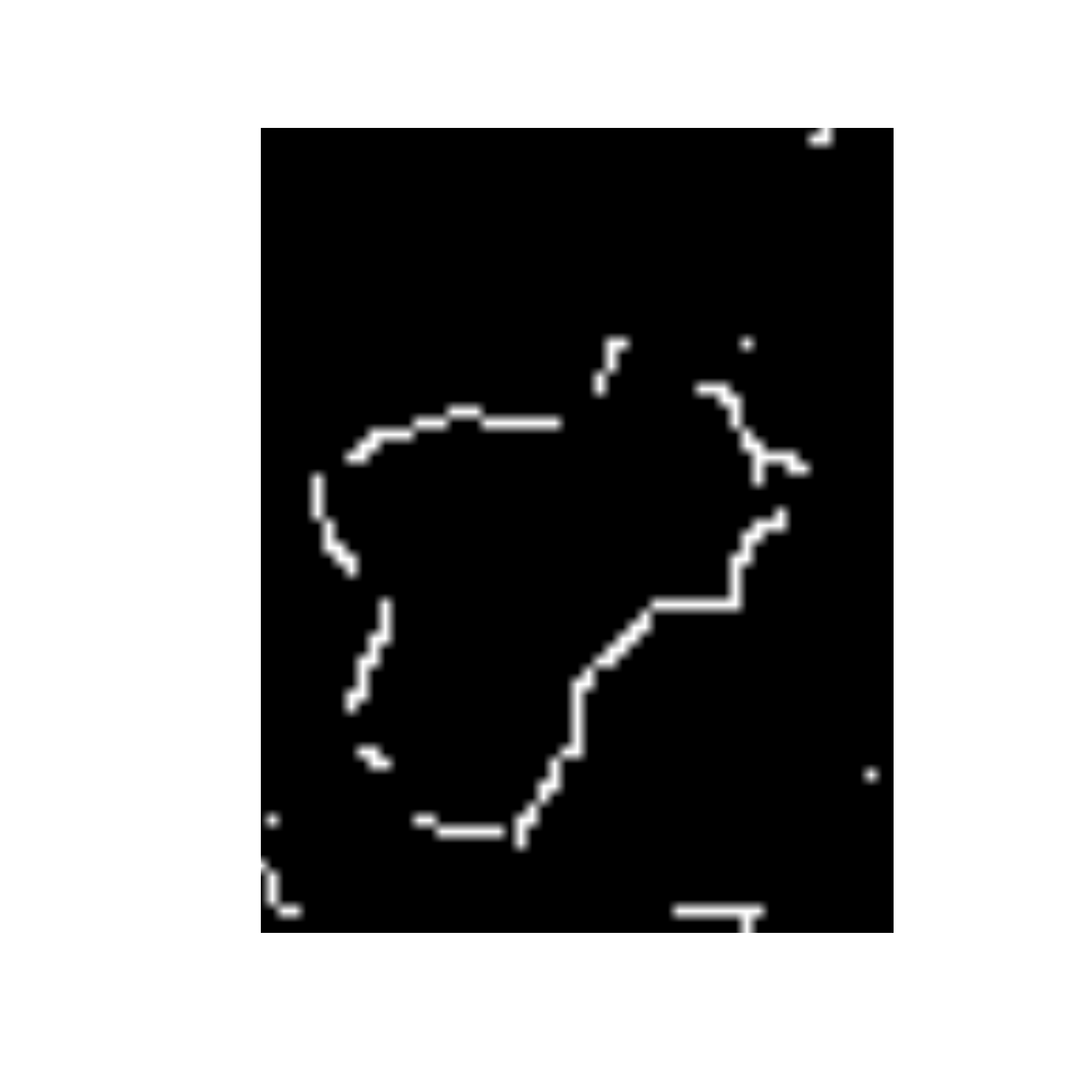}&
\includegraphics[scale = 0.1]{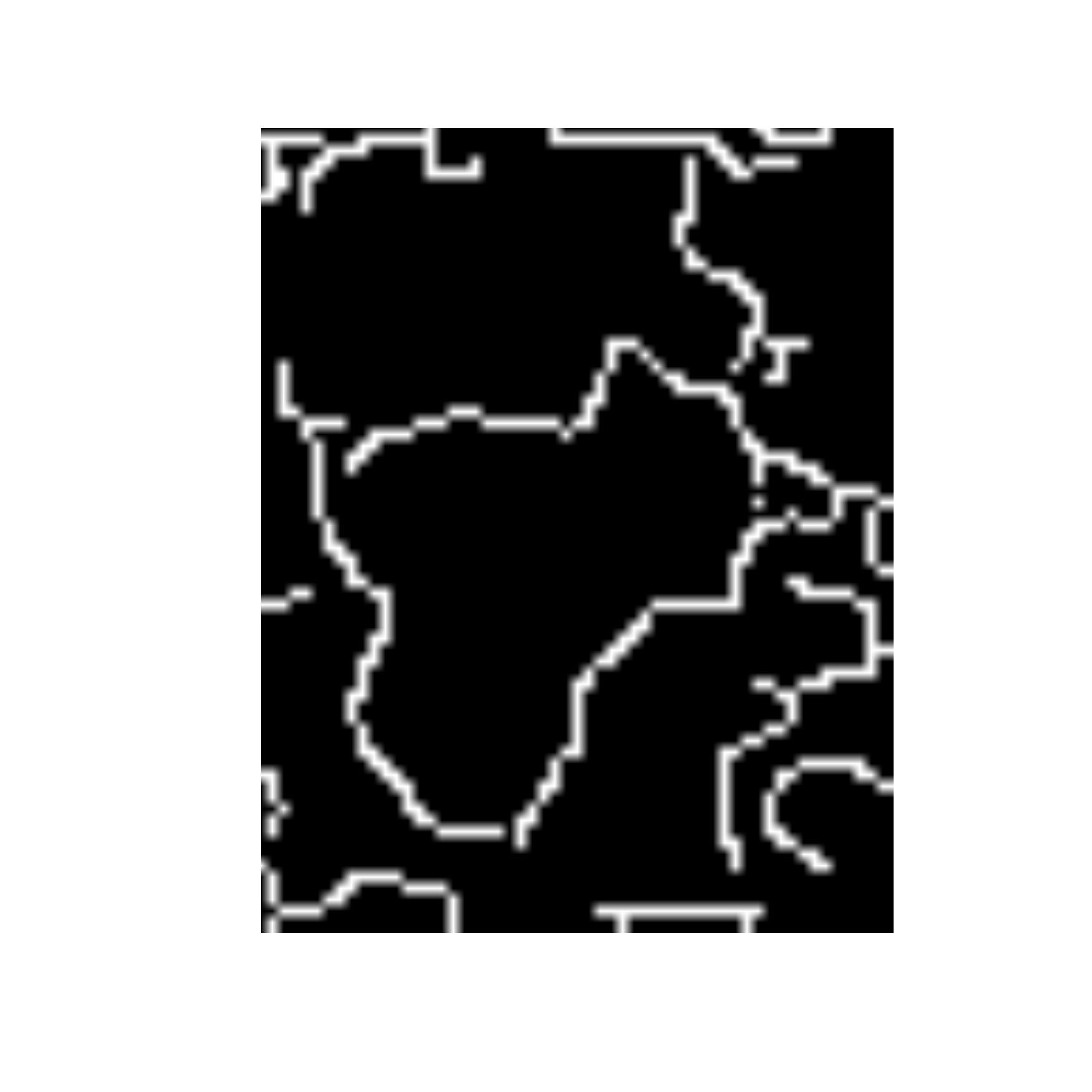}\vspace{-3mm}\\
\includegraphics[scale = 0.1]{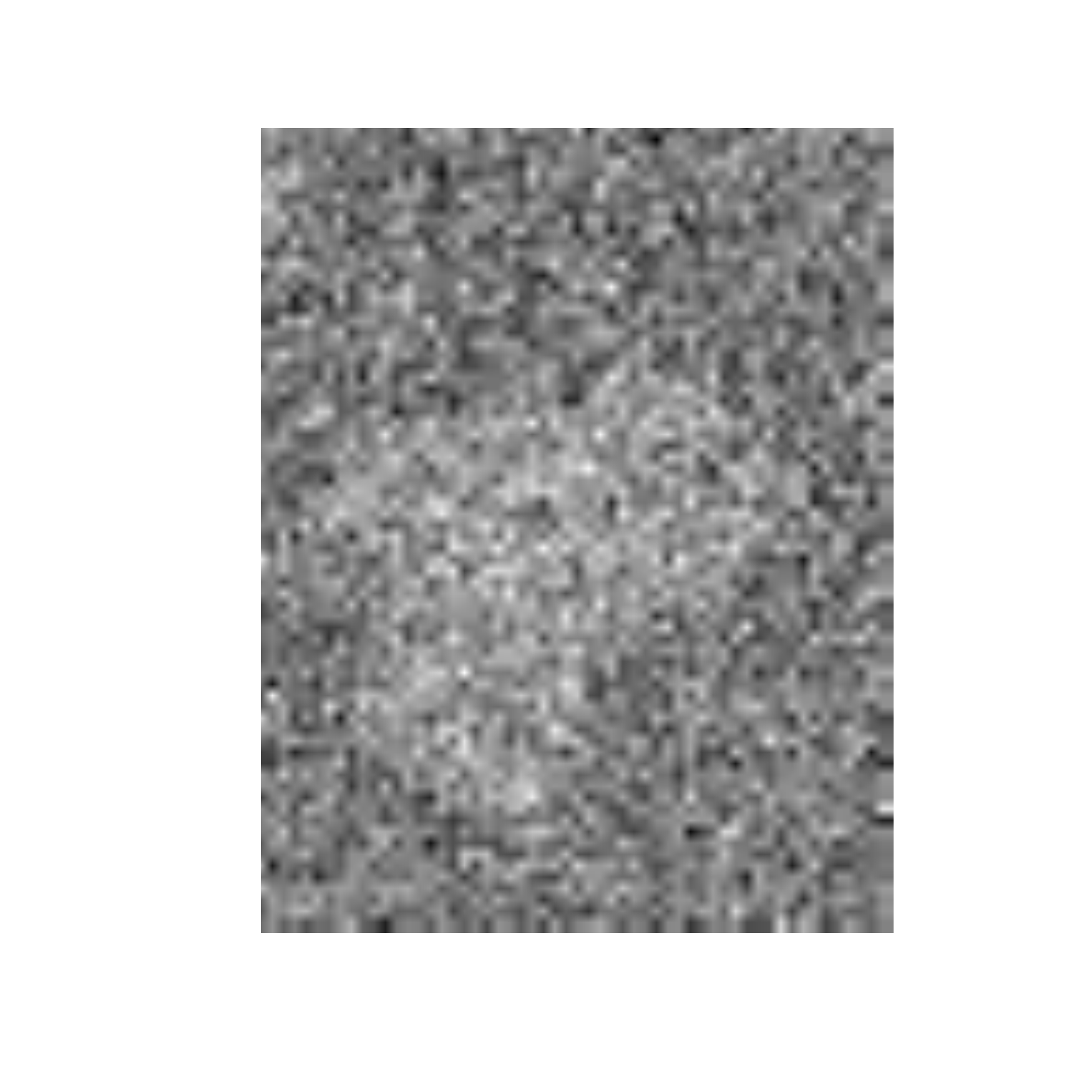}&
\includegraphics[scale = 0.1]{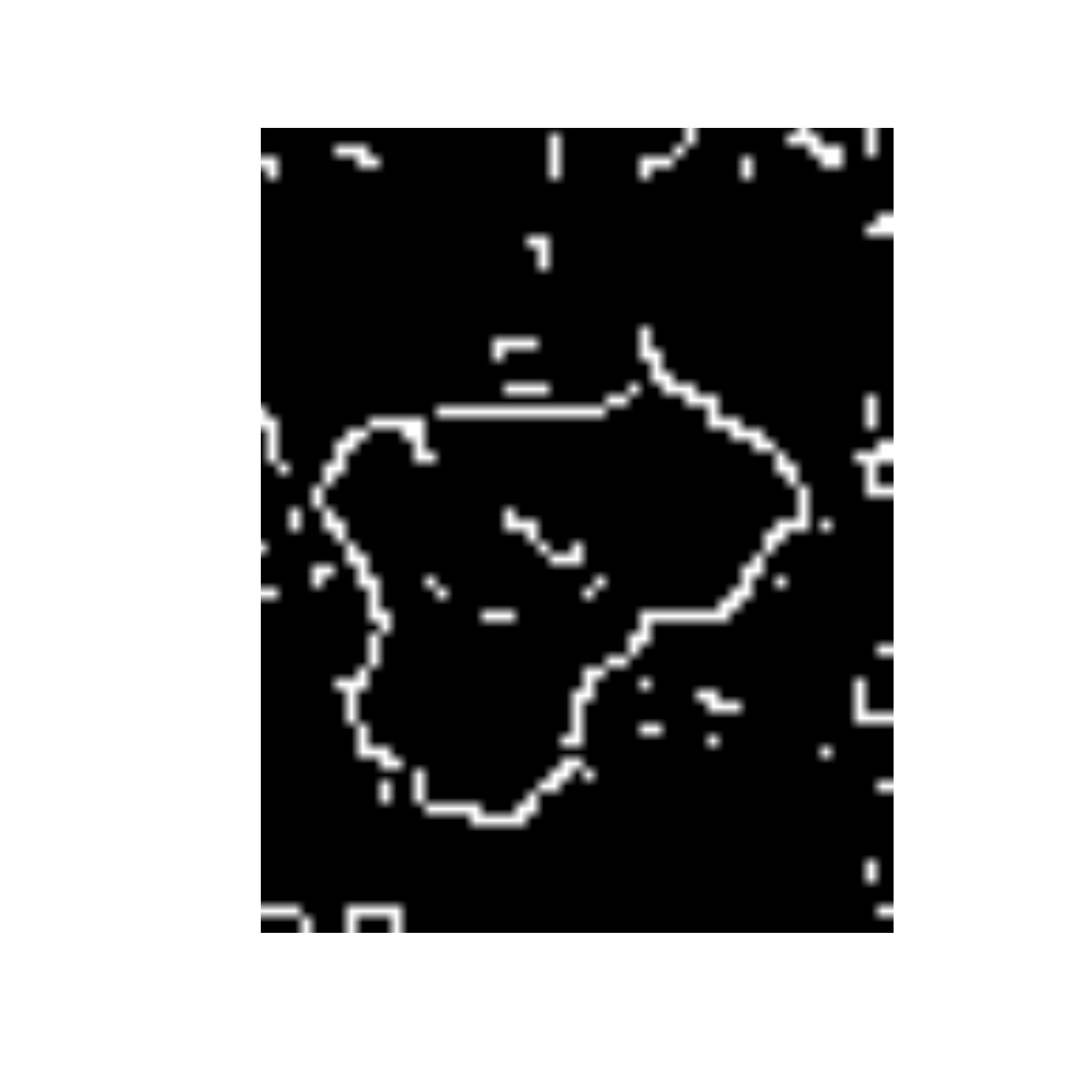}&
\includegraphics[scale = 0.1]{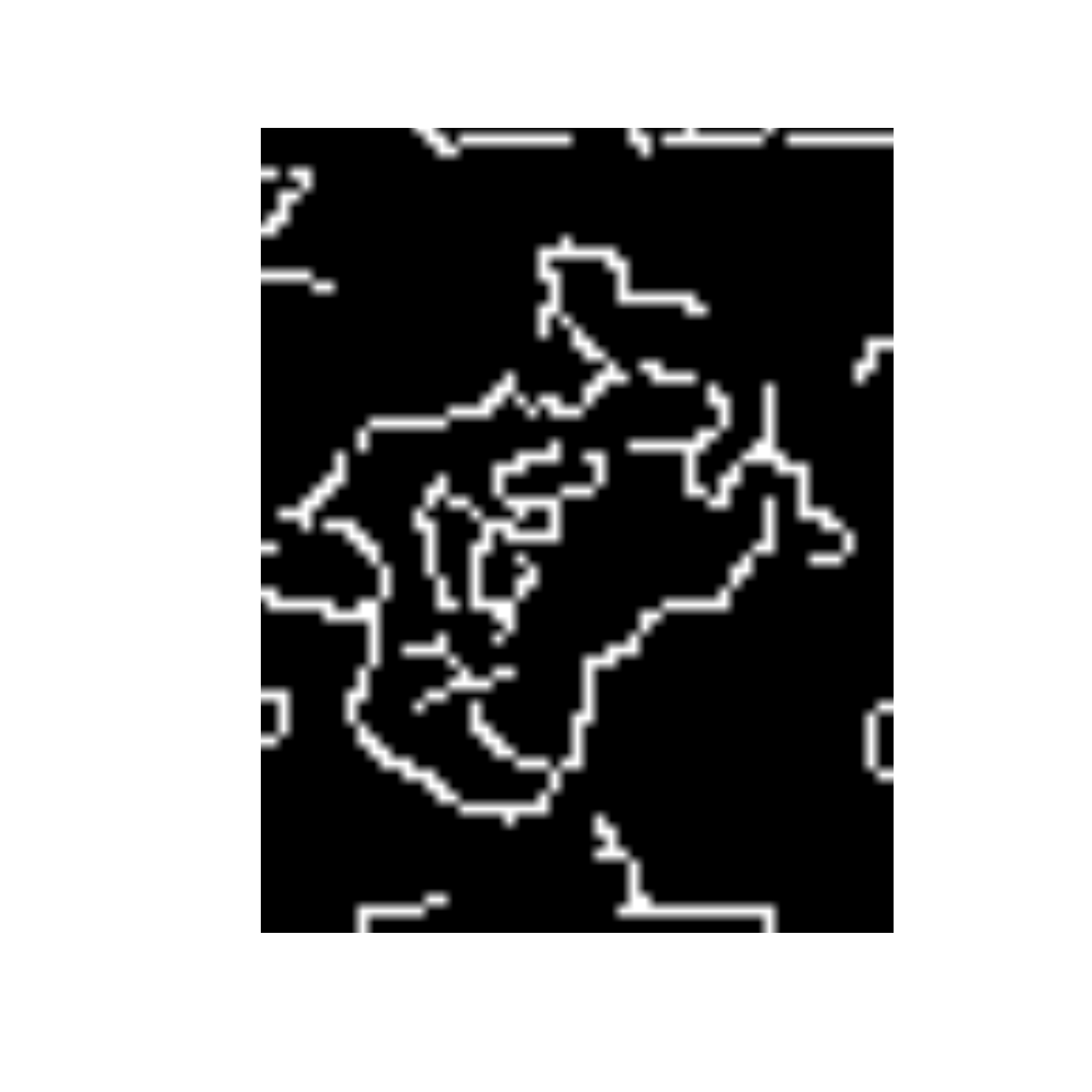}\vspace{-3mm}\\
\includegraphics[scale = 0.1]{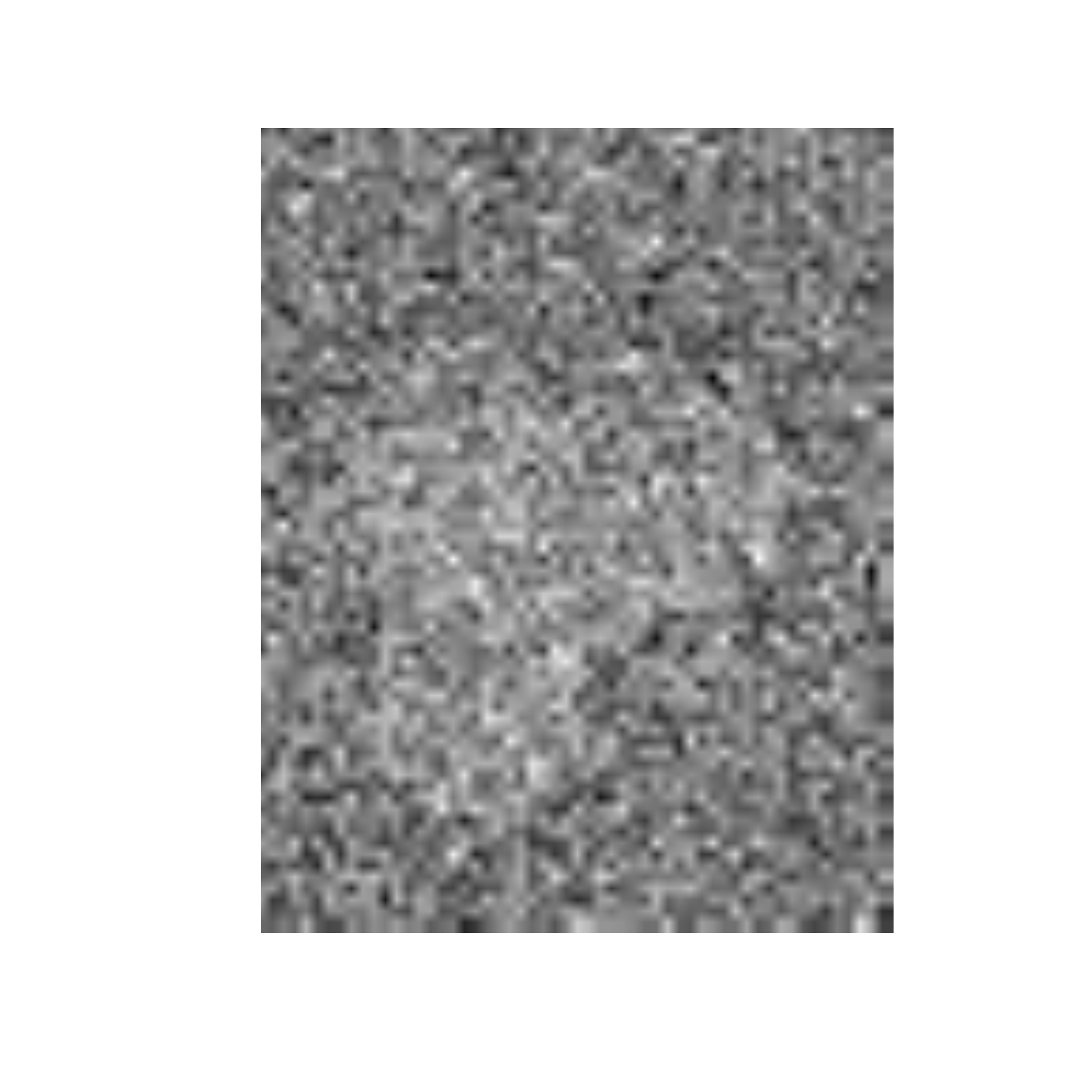}&
\includegraphics[scale = 0.1]{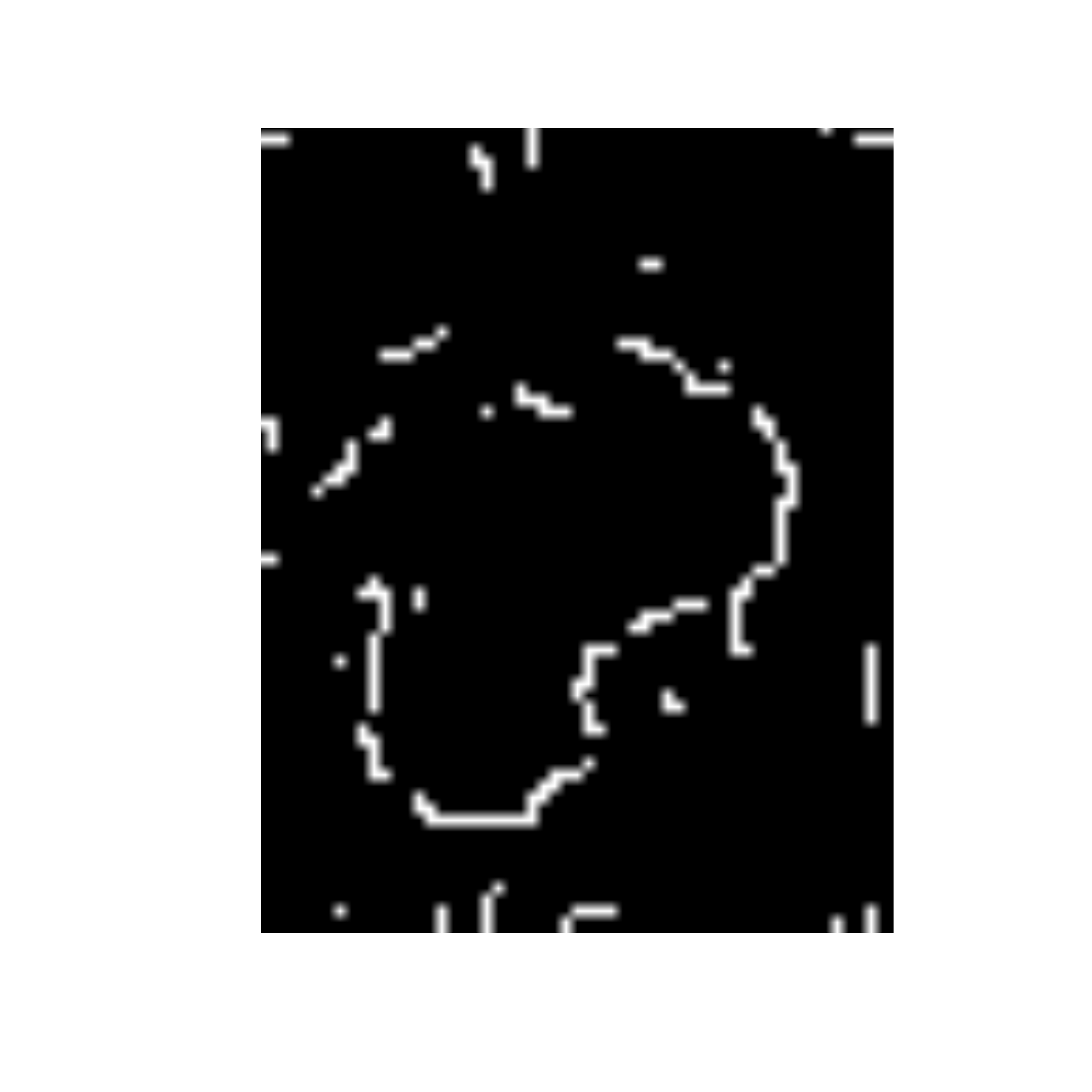}&
\includegraphics[scale = 0.1]{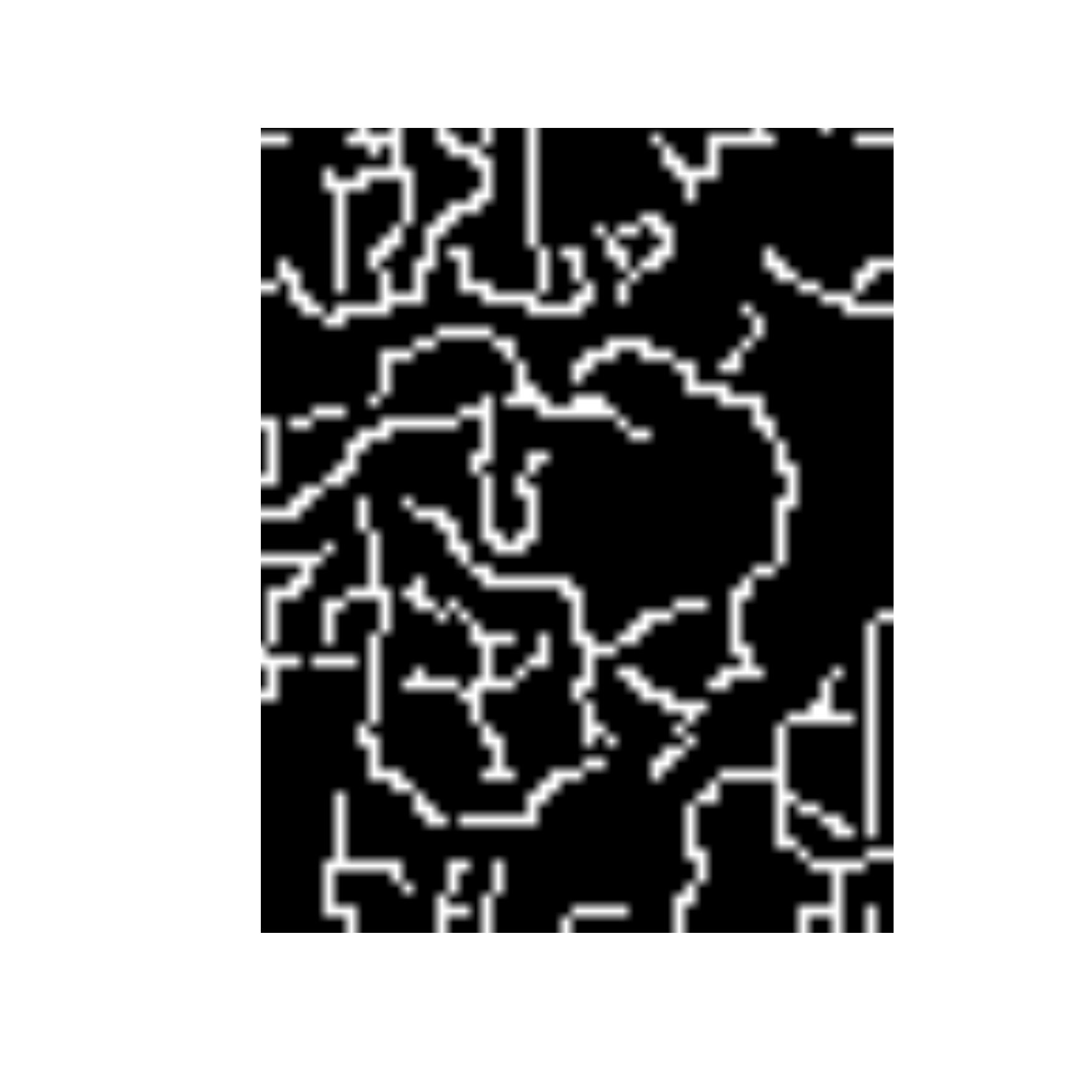}
\end{tabular}
\caption{  The edge detection results from one single simulation run
  with respective noise standard deviations: SD $= 0.2$, $0.5$, $1$, $1.2$,
  and $1.5$ from
  the top to the bottom panels. }\label{fig:edge}
\end{figure}

\begin{figure}
\centering
\begin{tabular}{cc}
$R_s$& $R_t$\\
\includegraphics[scale = 0.1]{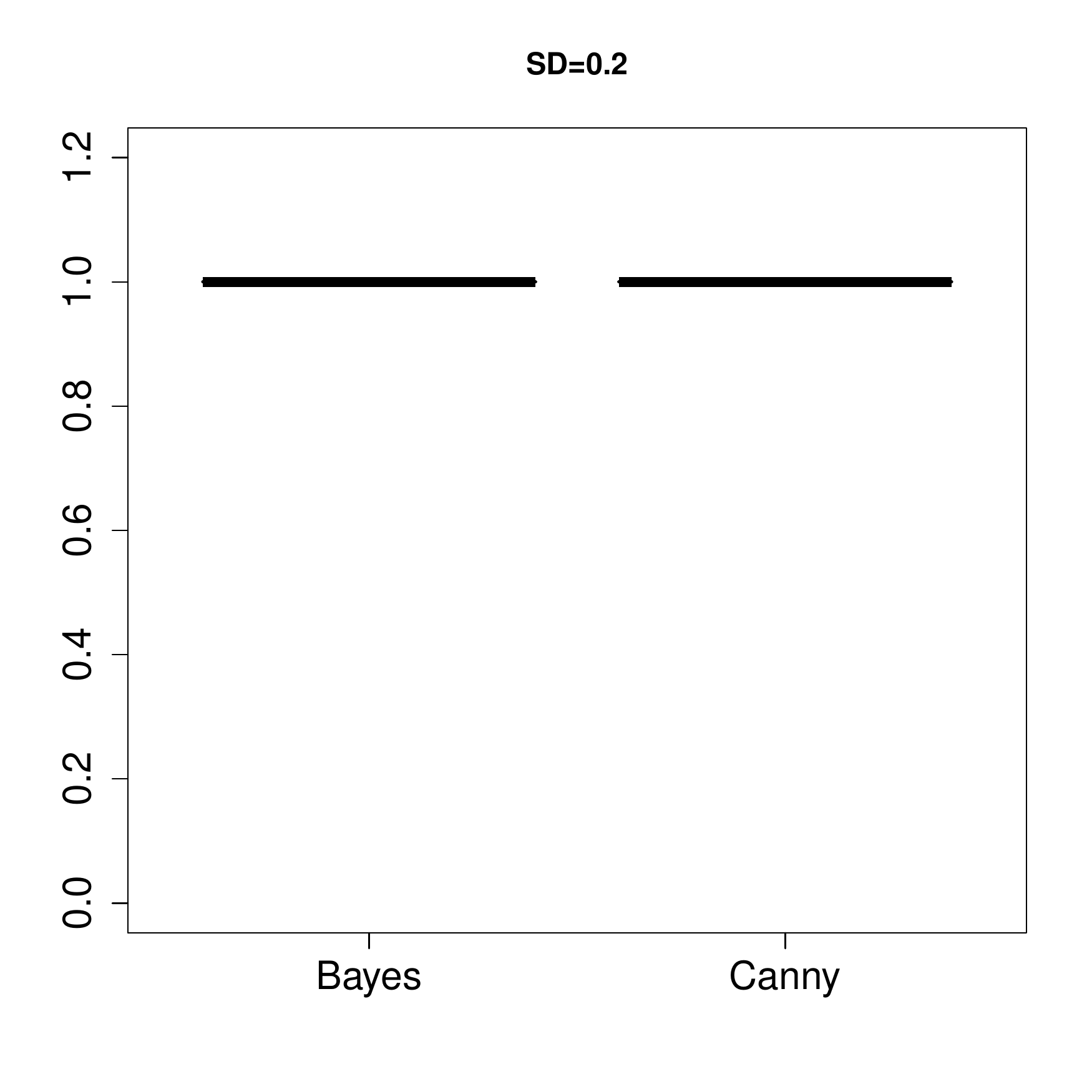}&
\includegraphics[scale = 0.1]{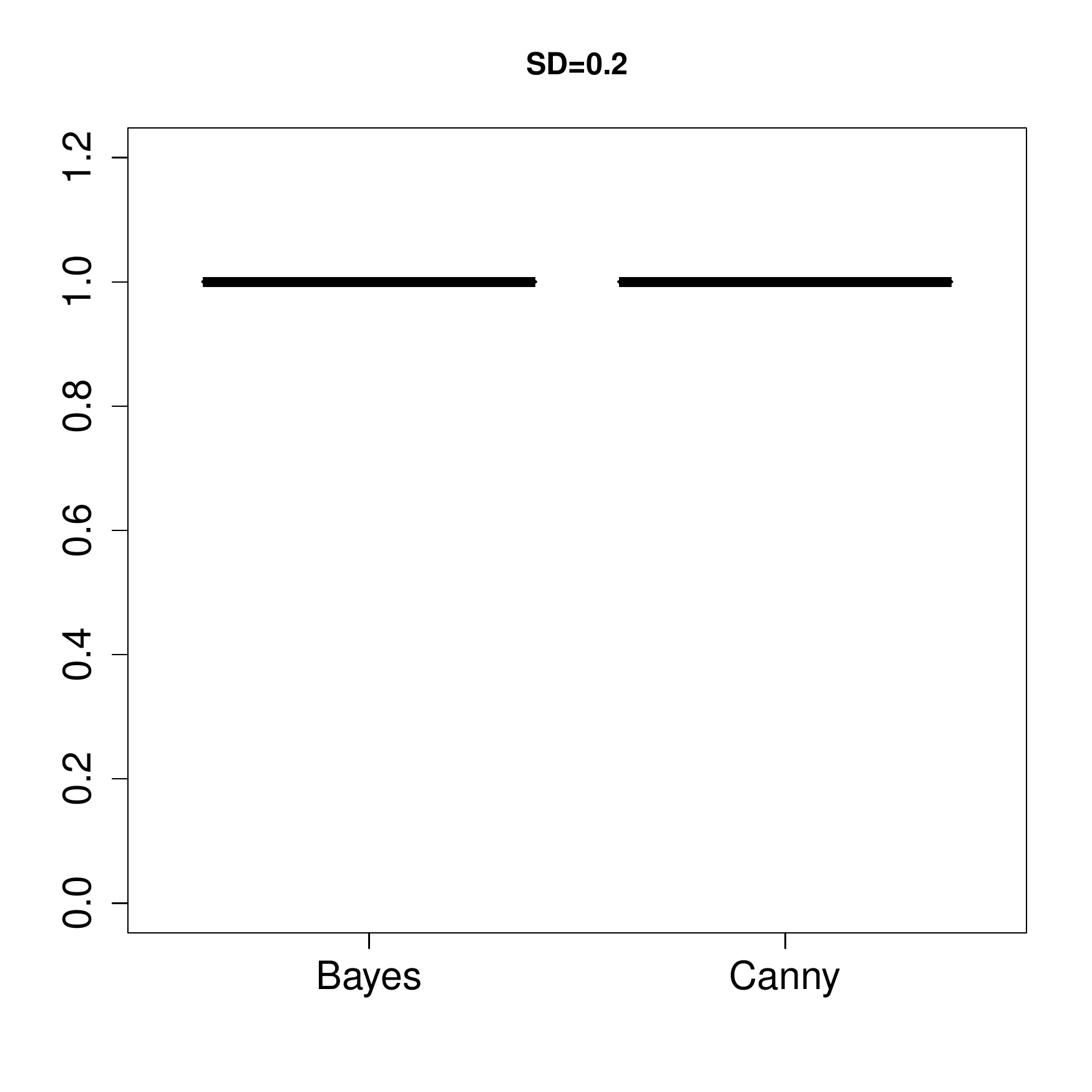}\\
\includegraphics[scale = 0.1]{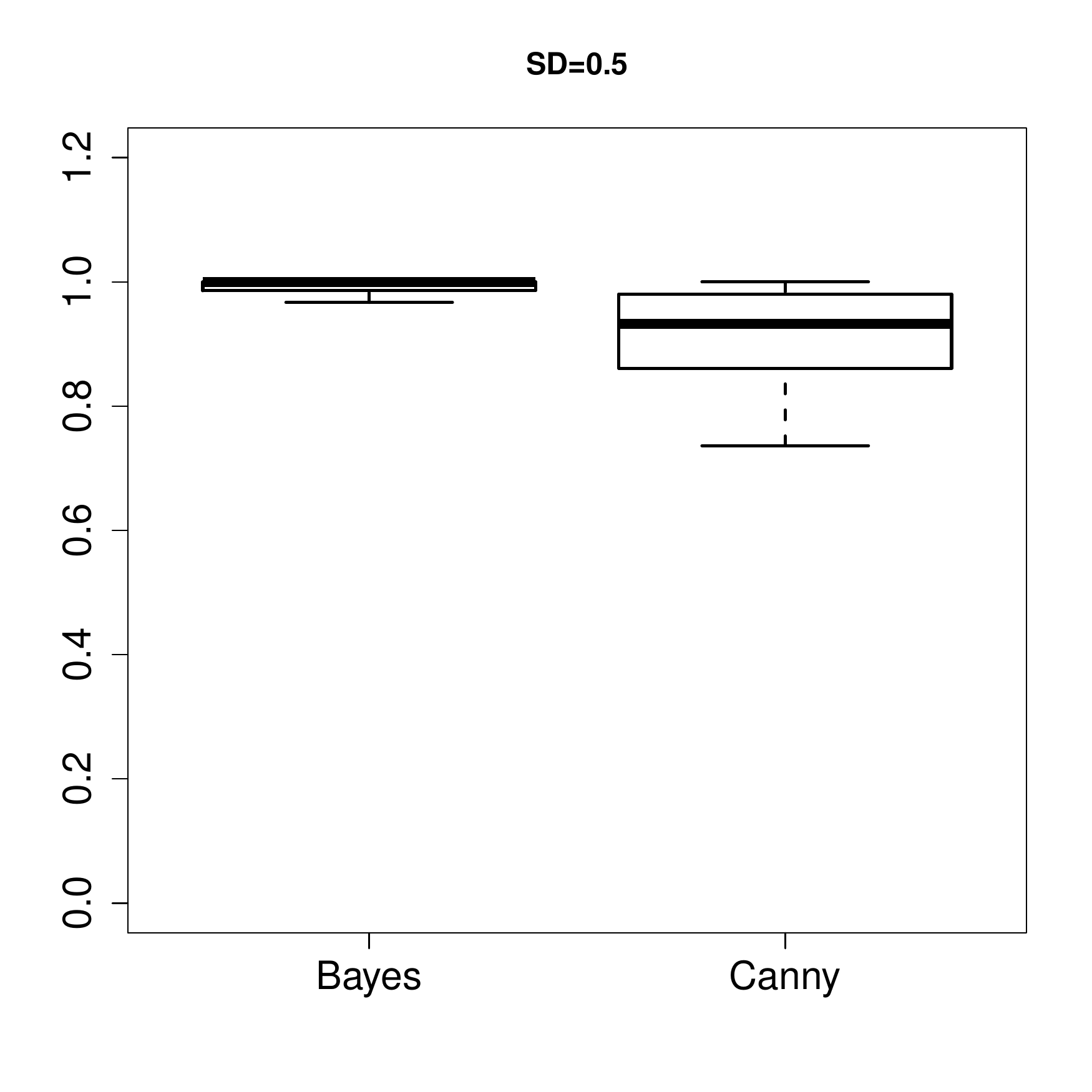}&
\includegraphics[scale = 0.1]{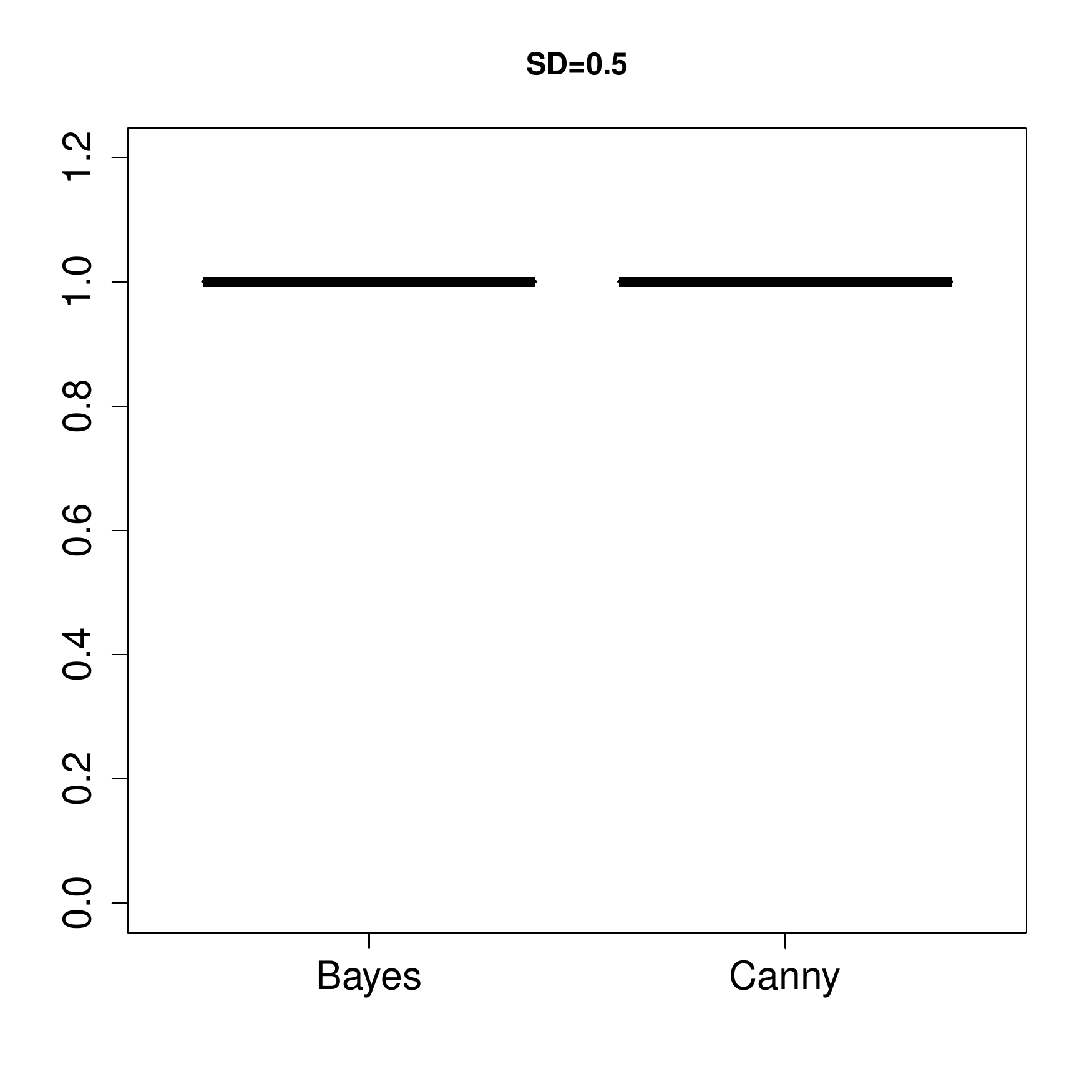}\\
\includegraphics[scale = 0.1]{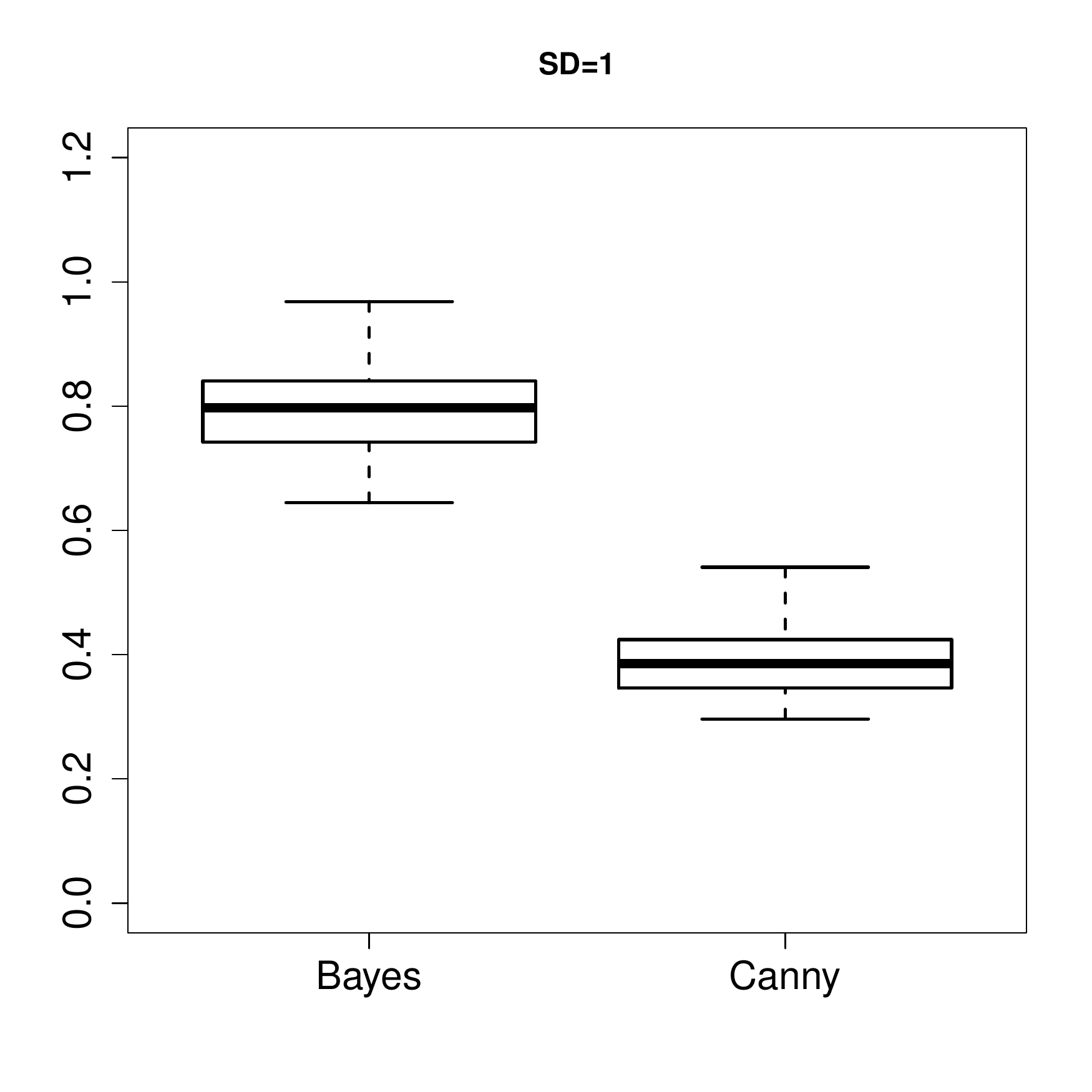}&
\includegraphics[scale = 0.1]{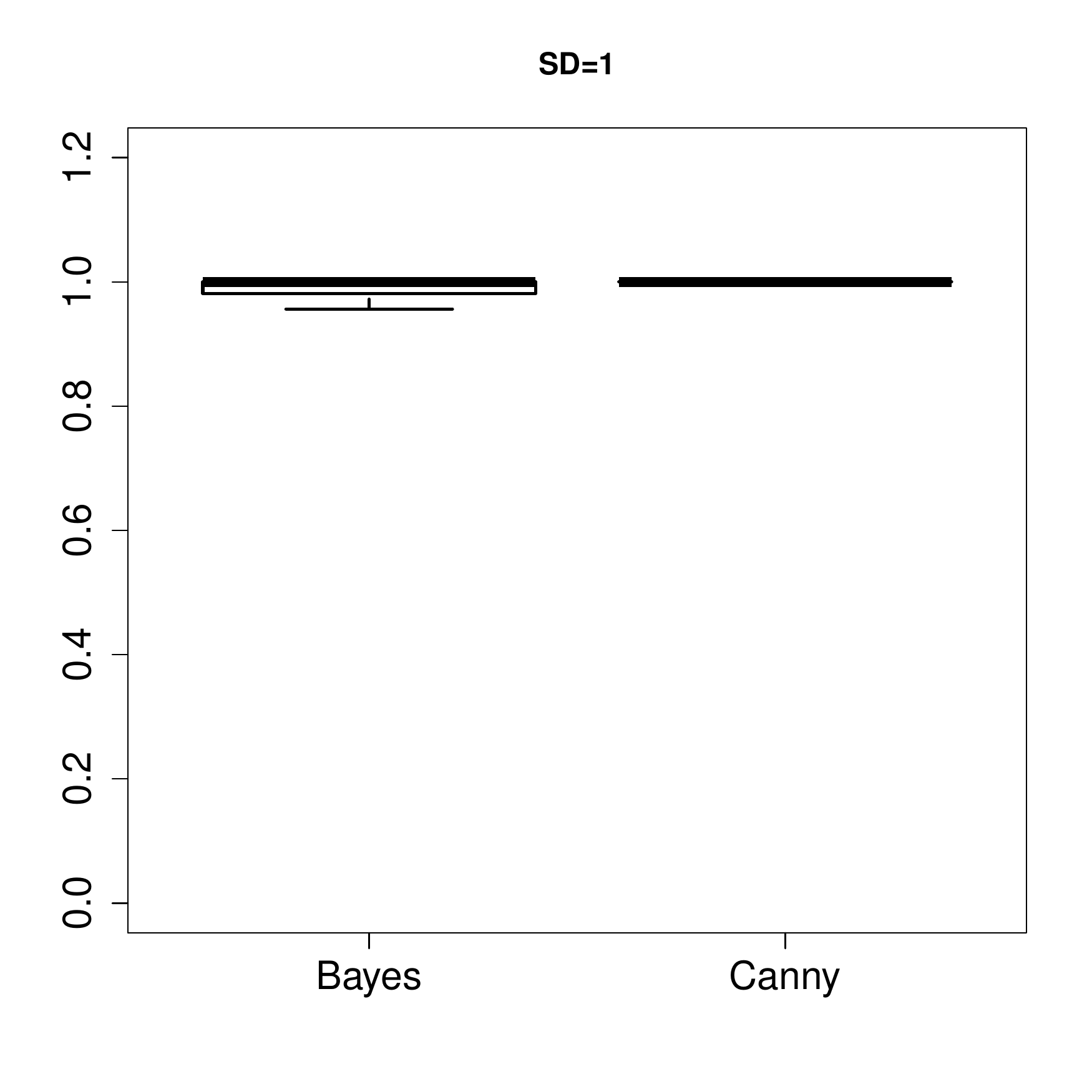}\\
\end{tabular}
\caption{The box-plots for $R_s$ and $R_t$  from 100 simulation runs
  with respective  noise standard deviations: SD $= 0.2$, $0.5$, and $1$ from the top
  to the bottom panels. }\label{fig:box}
\end{figure}

\begin{figure}
\centering
\begin{tabular}{cc}
$R_s$& $R_t$\\
\includegraphics[scale = 0.1]{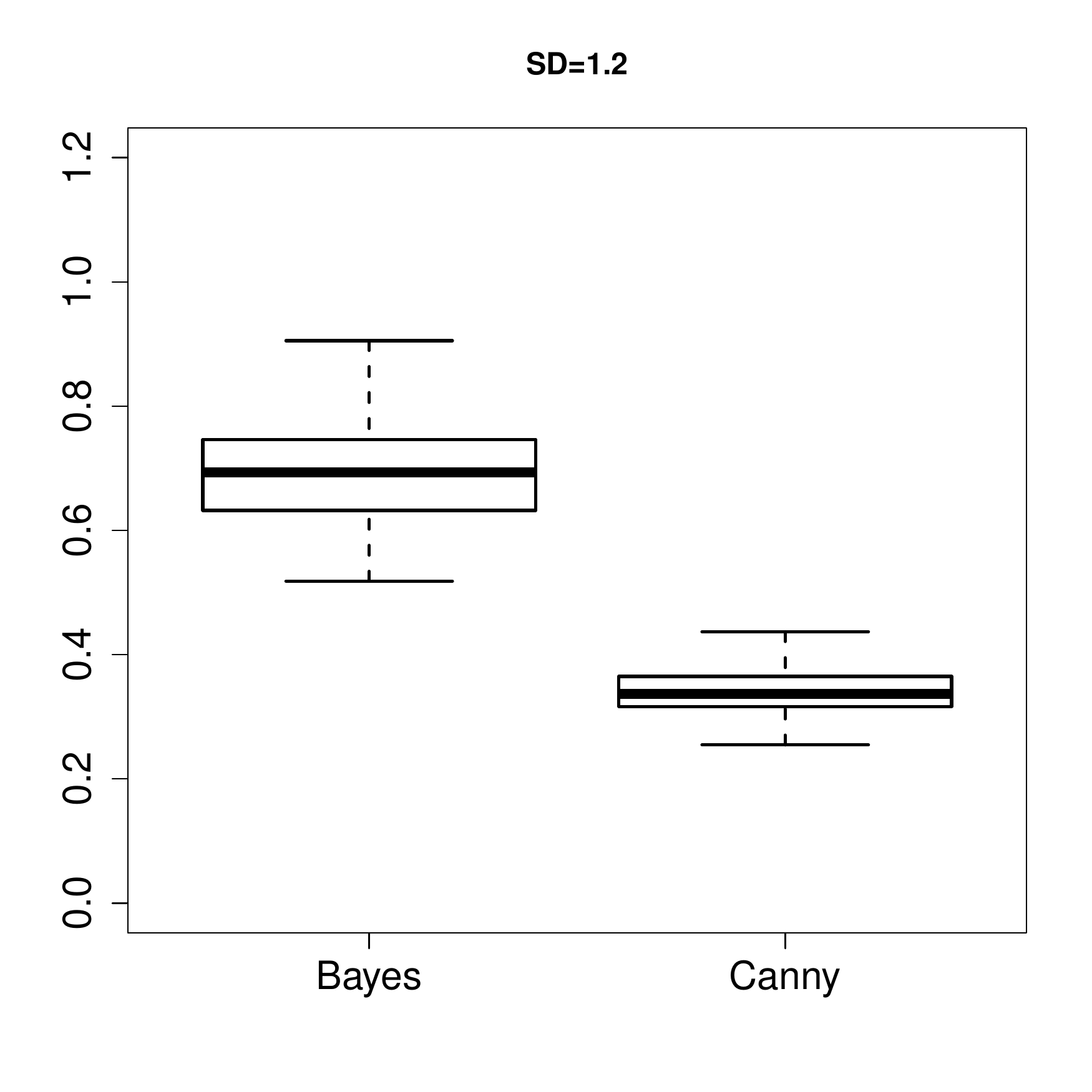}&
\includegraphics[scale = 0.1]{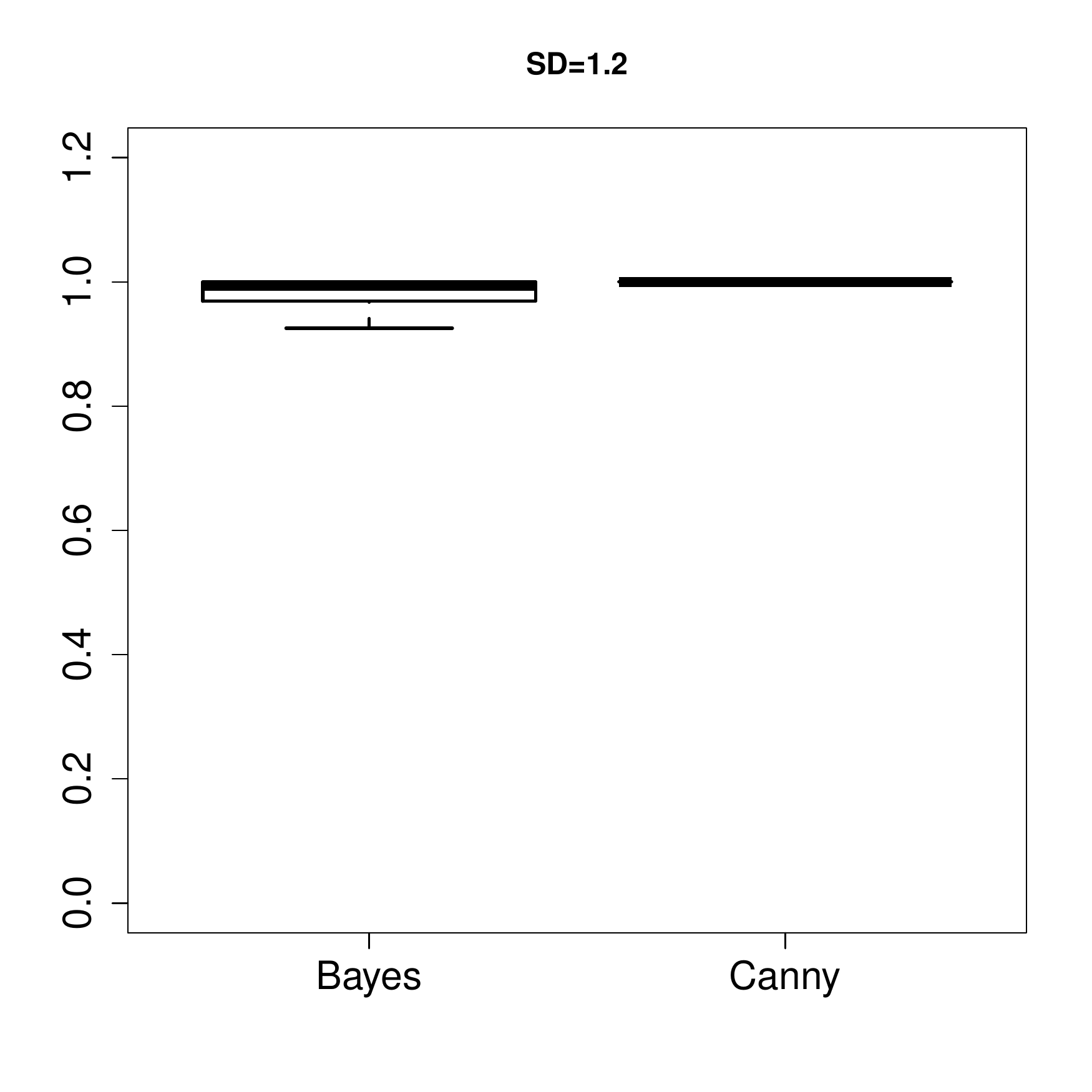}\\
\includegraphics[scale = 0.1]{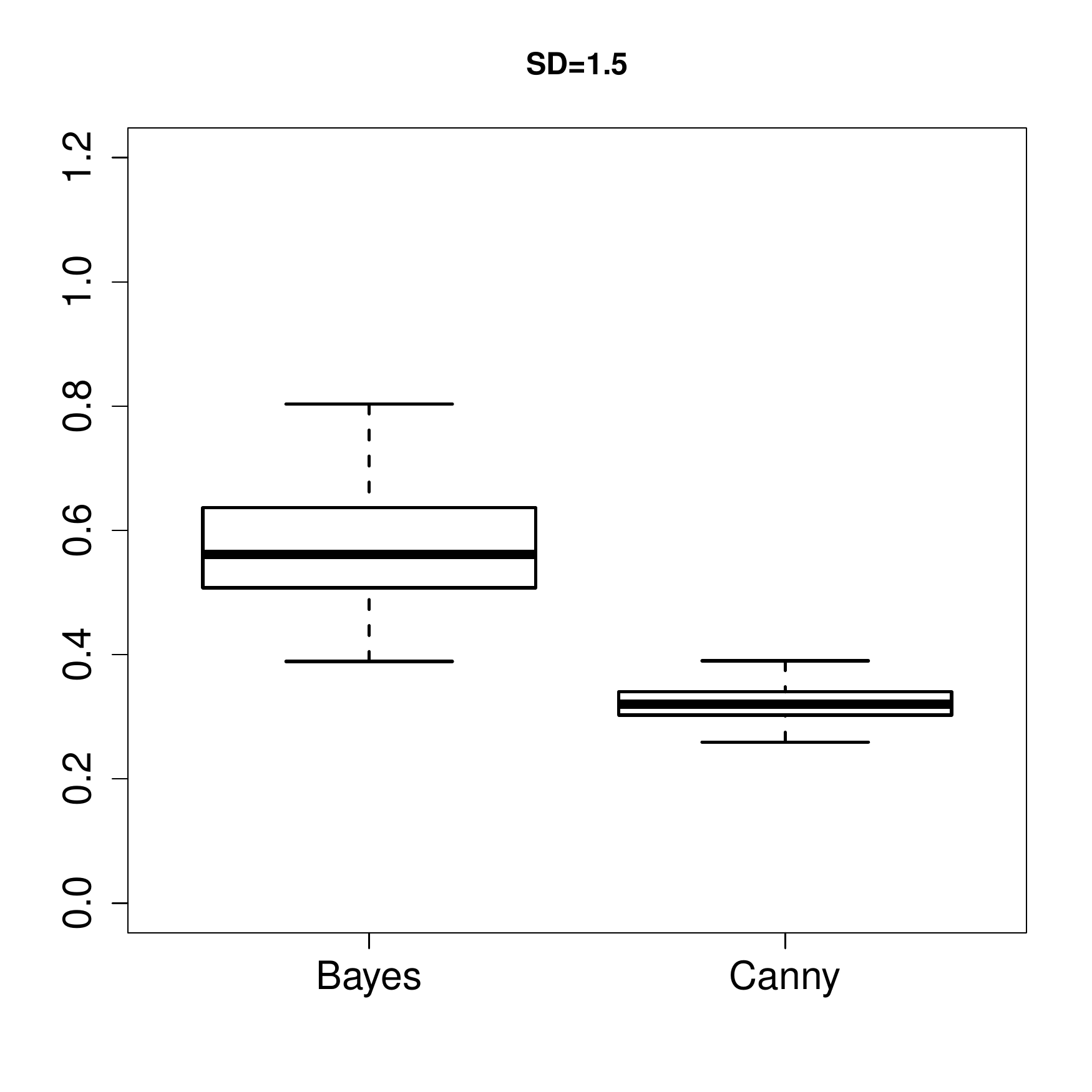}&
\includegraphics[scale = 0.1]{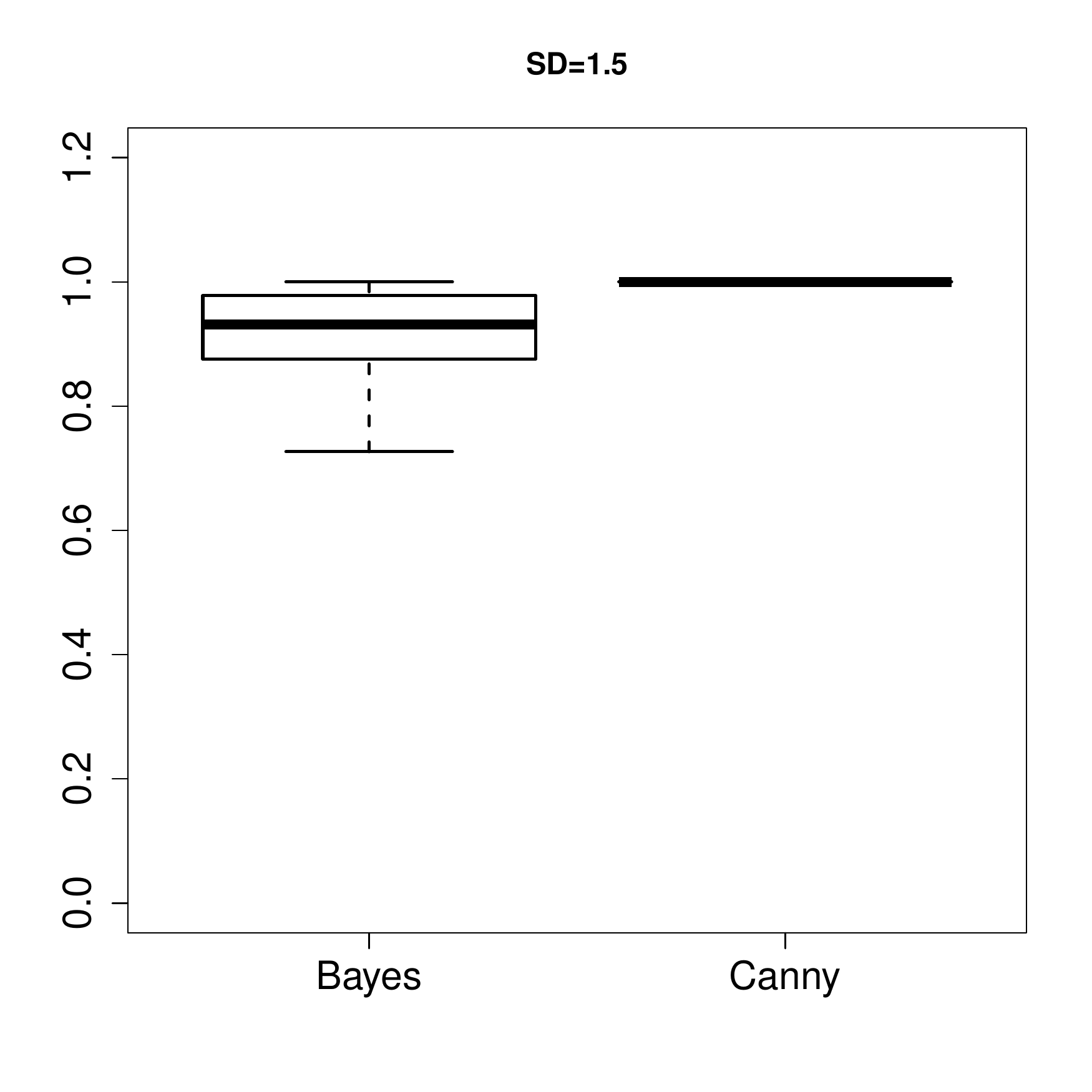}\\
\end{tabular}
\caption{The box-plots for $R_s$ and  $R_t$  from 100 simulation runs
  with respective noise standard deviations: SD $= 1.2$ and 1.5. }\label{fig:box2}
\end{figure}

We perform the simulation studies to evaluate the proposed method in detecting
the edges under the settings with various signal noise ratios. The simulation samples are
generated through adding the white noises with standard deviations 0.2, 0.5, 1, 1.2,
1.5 to a grayscale image.  In all the simulations,  we fix $\beta = -2$ and $\bSig =
\bf I$.  We present the edge detection results along
with the noised images in
Figure \ref{fig:edge} from a single simulation run.  Figure
\ref{fig:edge} shows that the Bayes detector and the Canny
detector have similar performances when the error variations are less
than 0.5. However, with larger error variations, the
Canny detector picks substantially amount of the non-edge pixels. On the
contrary,
Bayesian detector shows stable performance on eliminating the non-edge
pixels.
We further compare the Bayesian and Canny detector over the 100 simulation
studies.
Let $\mathcal S$ be the set containing the selected edge
pixels, and $\mathcal T$ be the one containing the true edge
pixels. The point $(x, y)$ is considered as an edge point if $\|(x, y) -
(x_t, y_t)\| <\kappa$ for at least one $(x_t, y_t)\in \mathcal T$,
where $\kappa$ is the cutoff value to declare the
two points are from different regions and $\|\cdot\|$ is the $L_2$ norm.
We define
\bse R_s = \frac{\sum_ {(x, y) \in \mathcal S} \min_{(x_t, y_t) \in \mathcal T } \|(x, y) - (x_t, y_t)\| \leq
\kappa}{\sum_{x, y} I \{(x, y) \in \mathcal S\}},
\ese
and
\bse R_t = \frac{\sum_ {(x_t, y_t) \in \mathcal T} \min_{(x, y) \in \mathcal S } \|(x, y) - (x_t, y_t)\| \leq
\kappa}{\sum_{x_t, y_t} I \{(x_t, y_t) \in \mathcal T\}}.
\ese
Clearly, $R_s$ represents the proportion of the selected pixels being
the true edge points, while $R_t$ represents the proportion of
the true pixels being selected. The ideal case is $R_s = R_t =
1$. However, when the noise variation increases, there are tradeoffs
between $R_s$ and $R_t$. In Figure \ref{fig:box} and Figure
\ref{fig:box2}, we present the
box-plot based on the 100 simulations and $\kappa = \sqrt{3^2 + 3 ^2}$. It is clear that the Bayesian
detector outperforms the Canny detector in terms of the significant larger
$R_s$ values. On the other side, to achieve a smaller false discovery rate, the
Bayesian detector inevitably  sacrifices the recovery of all the true edge
pixels, leading to smaller $R_t$ values compared to the Canny detector.
Nevertheless, it still yields $R_t > 80$
on average,  which is sufficient to describe the shape of the damage in
practice.

\section{Roof damage detection}
\subsection{Damage detection on the interiors of the shingles}

Nowadays, the
standard three-tab asphalt shingles are the most commonly used roof
materials in the United States, because they are economical and easy to
install. However, such roofing materials are not as durable and long
lasting as some of the others such as metal, slate, or clay tile.
Hence, they are the most vulnerable materials when facing the hail
storms. In addition, the surfaces of the asphalt shingle roofs are
typically rough, which brings more challenges to the edge
detection in the hail damages.
In all the analyses, we fix $\bSig = \bf I$ and
standardize the smoothed image by the sample standard deviations.

\begin{figure}[!h]
\centering
\begin{tabular}{ccc}
Images & Bayesian detector& Canny detector\\
\includegraphics[scale = 0.1]{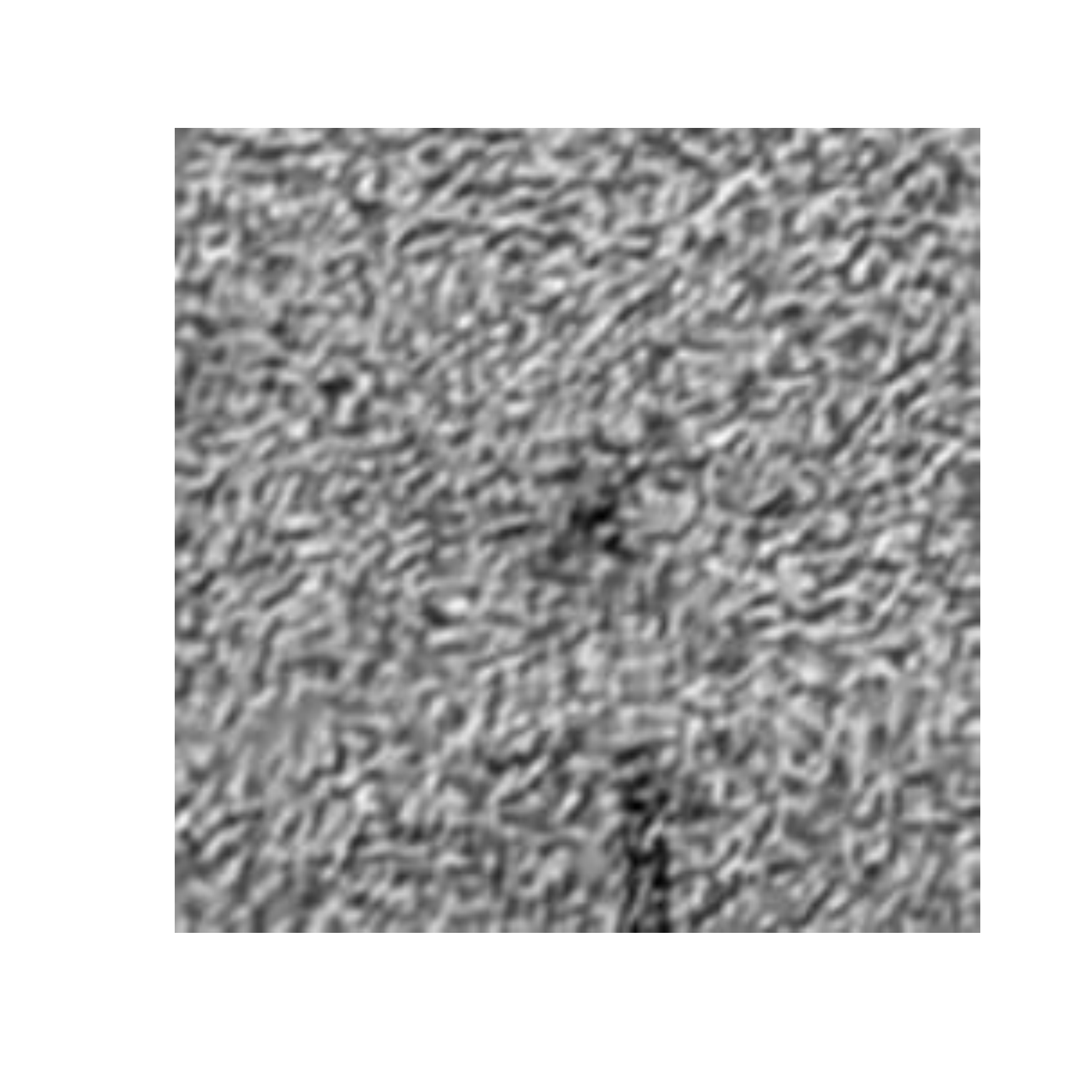}&
\includegraphics[scale = 0.1]{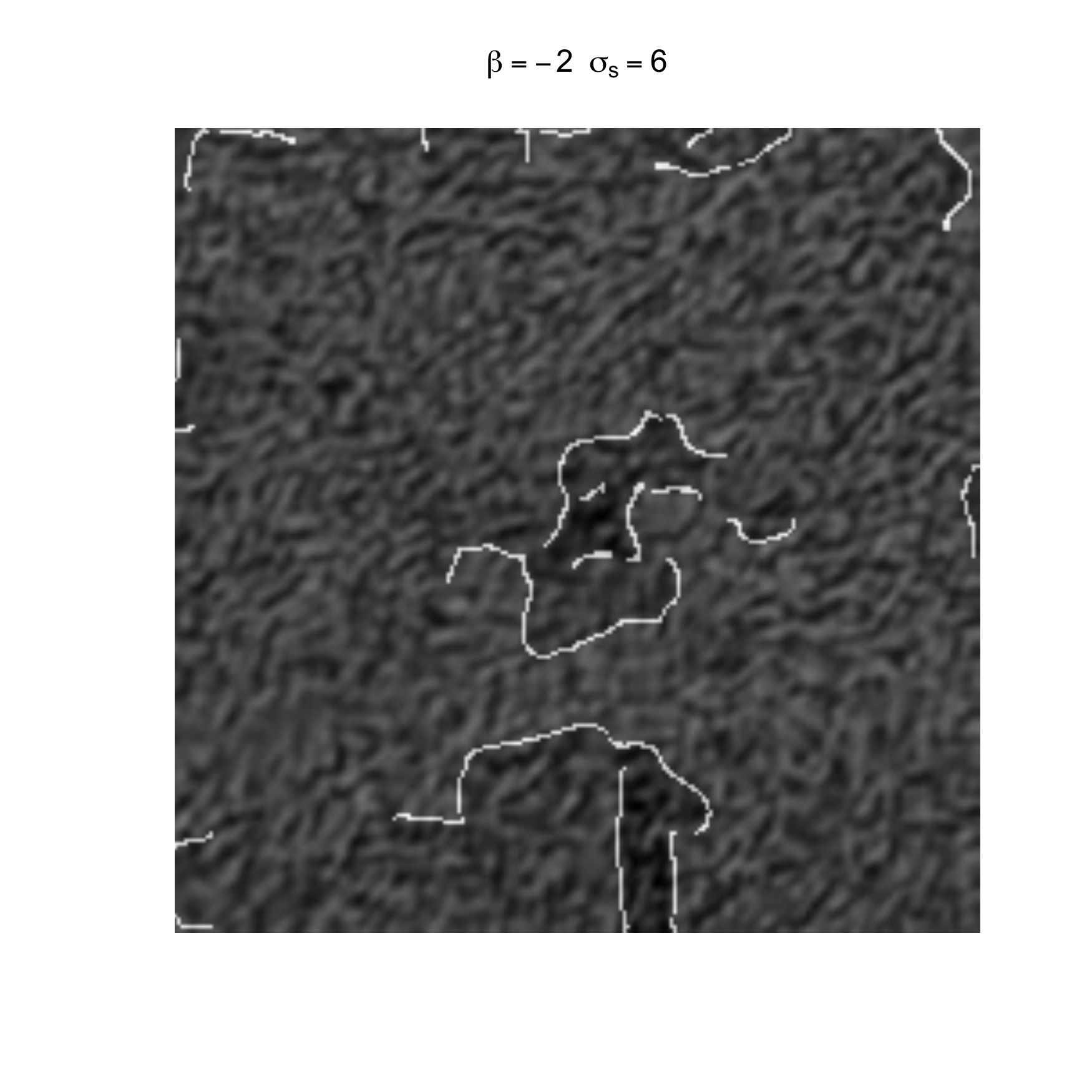}&
\includegraphics[scale = 0.1]{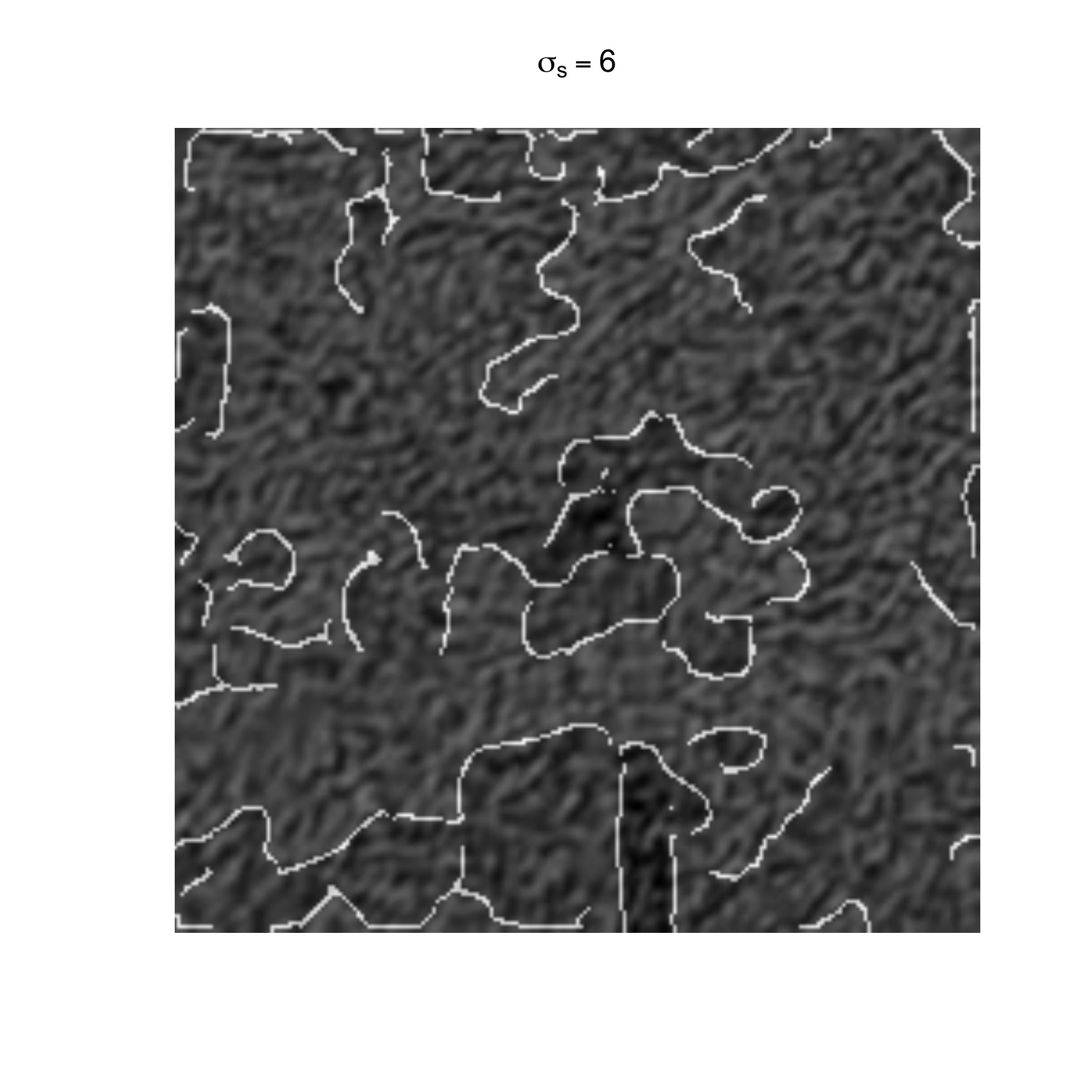}\\
\includegraphics[scale = 0.1]{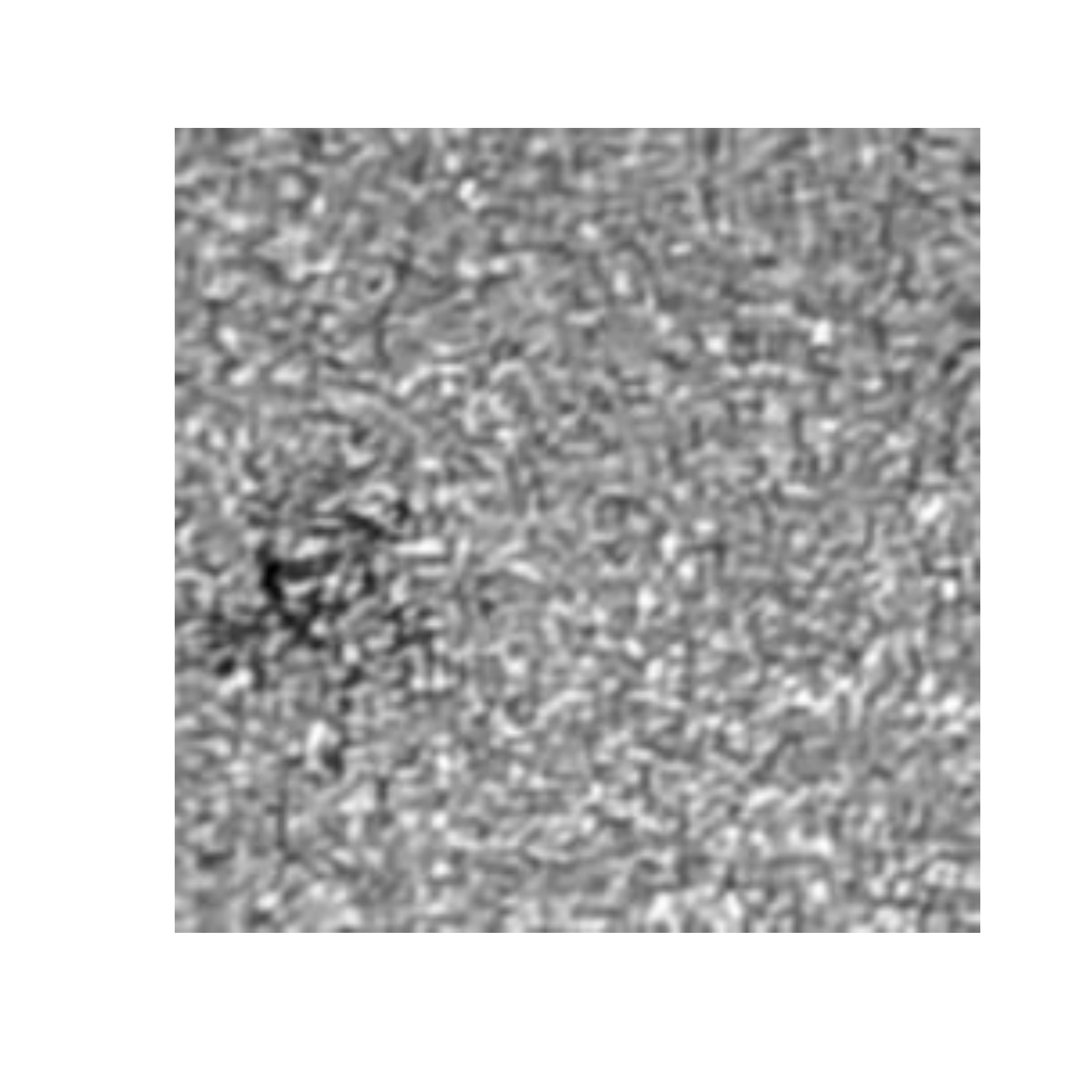}&
\includegraphics[scale = 0.1]{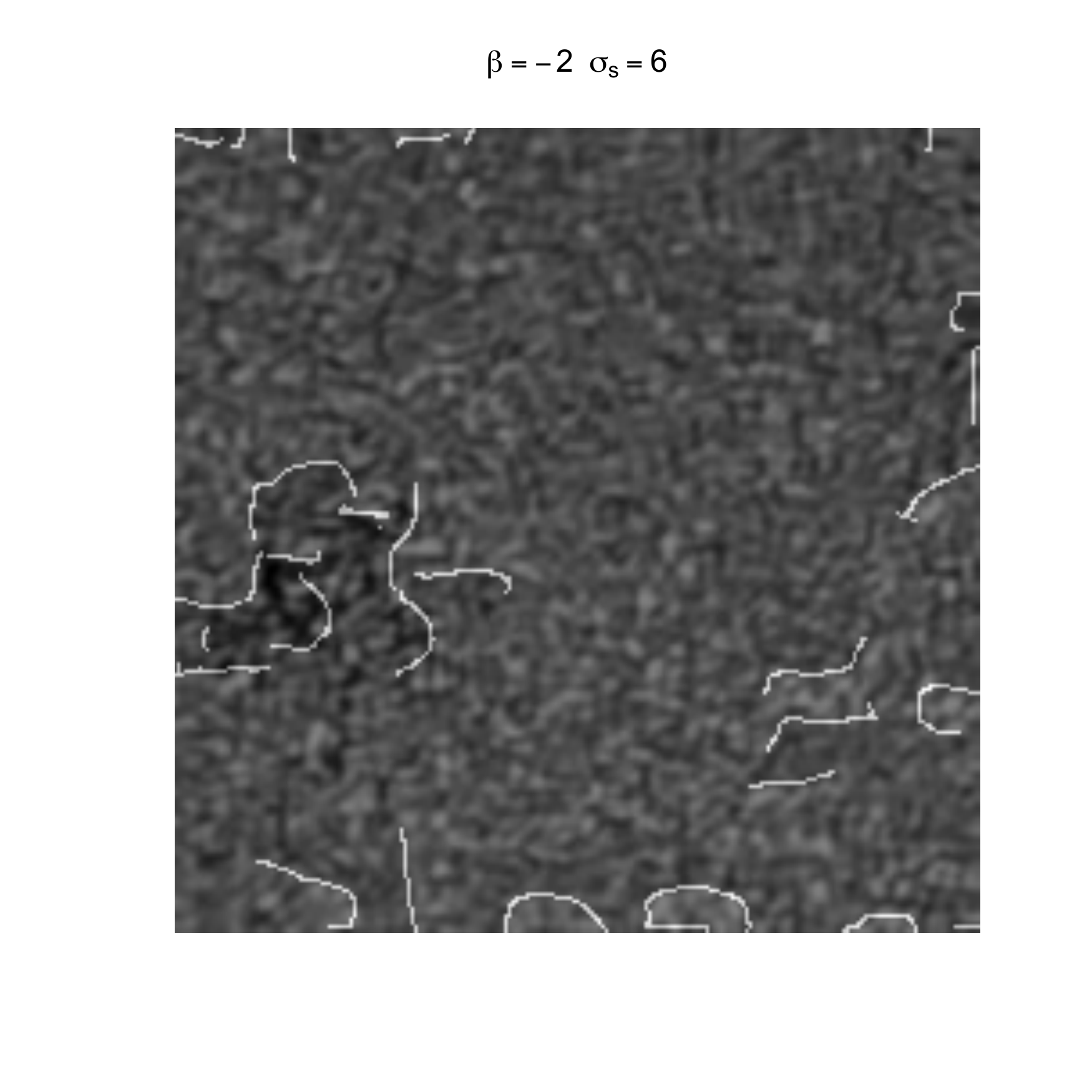}&
\includegraphics[scale = 0.1]{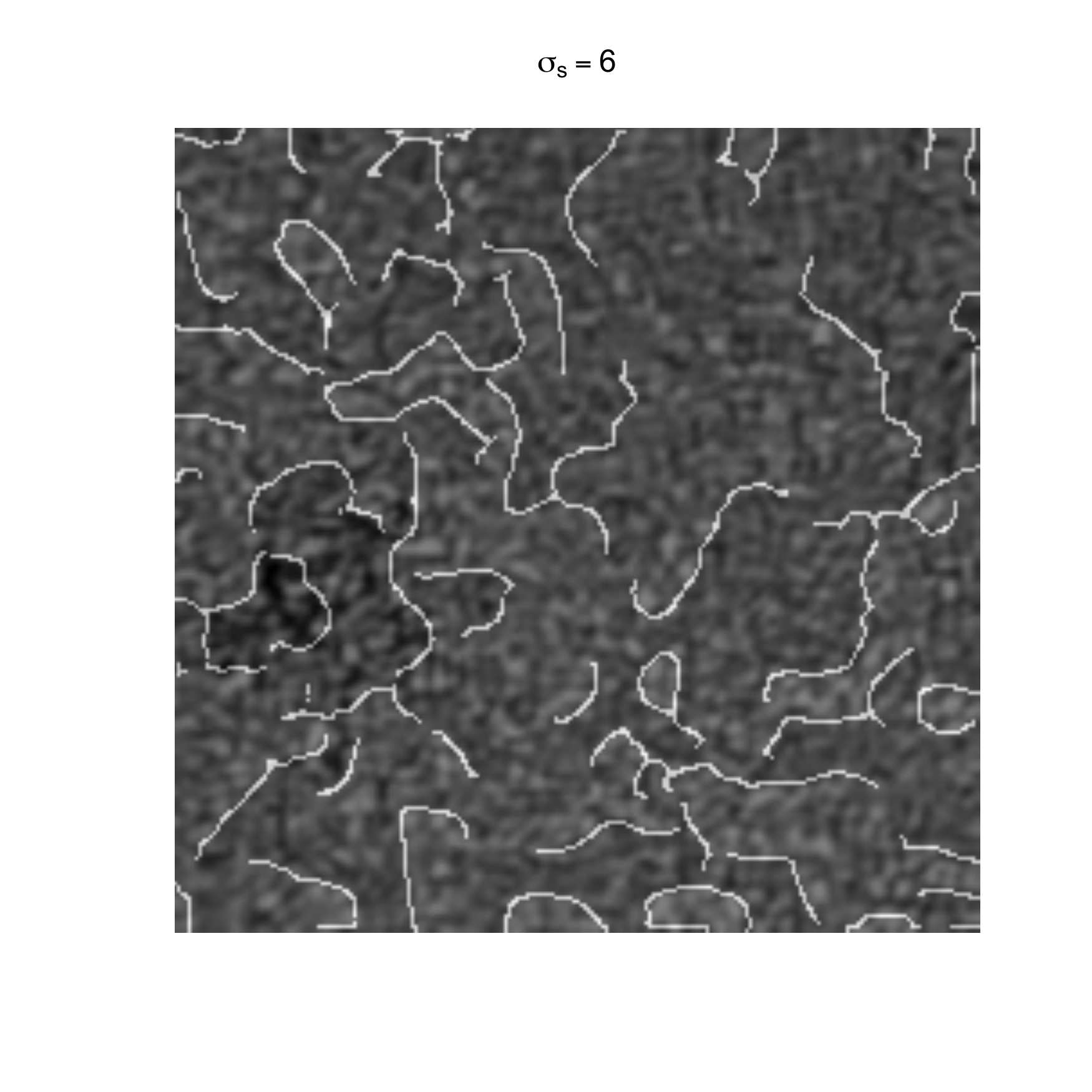}\\
\includegraphics[scale = 0.1]{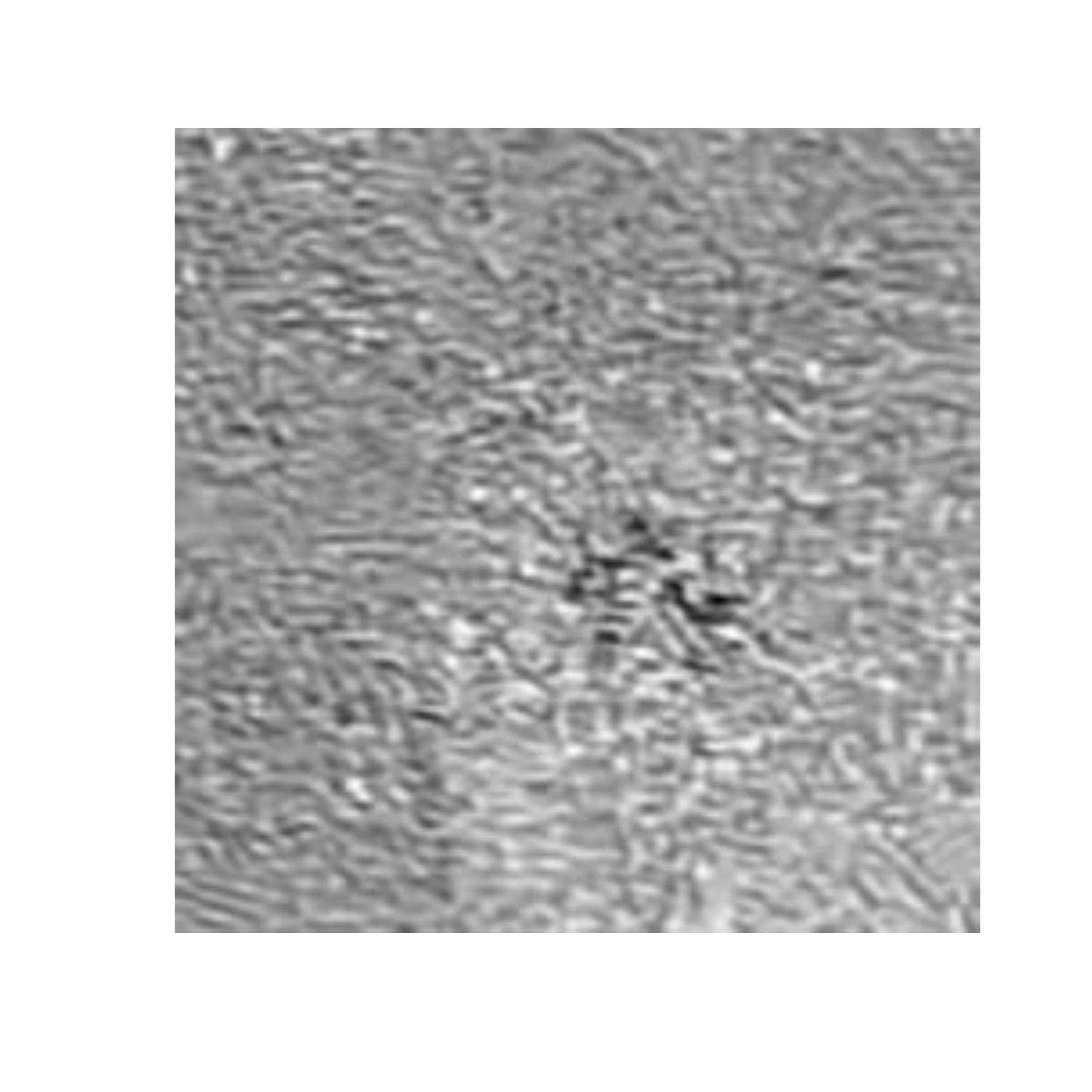}&
\includegraphics[scale = 0.1]{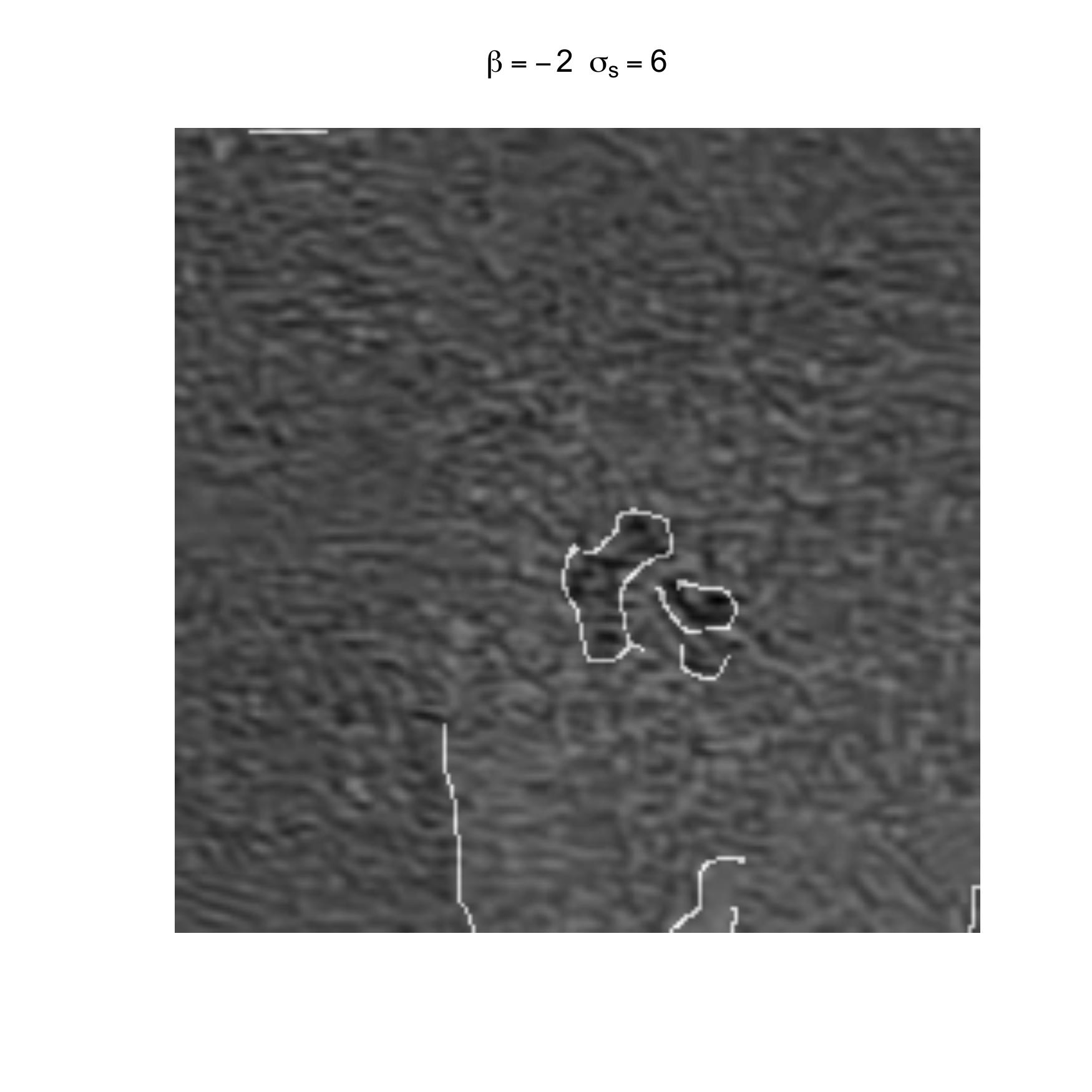}&
\includegraphics[scale = 0.1]{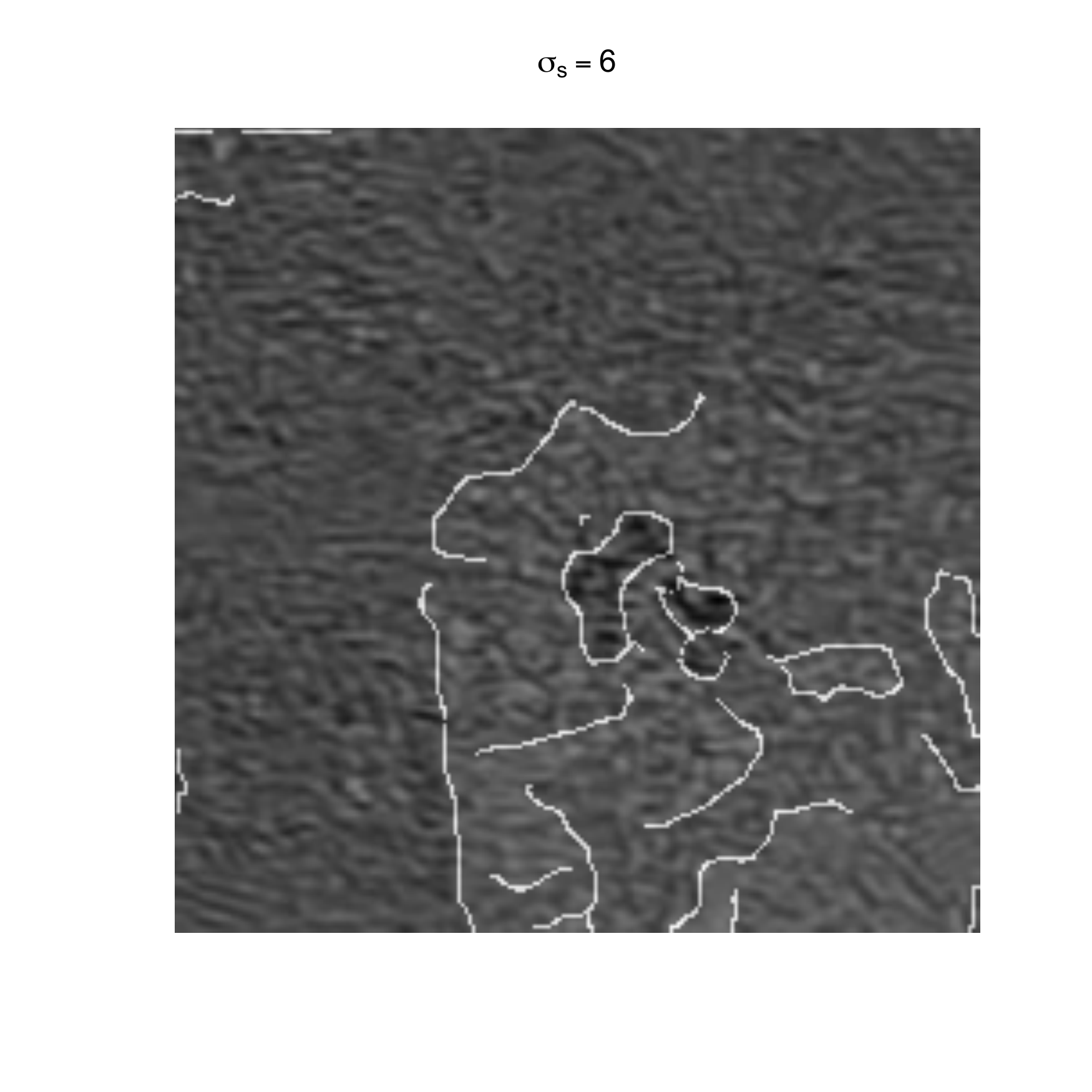}\\
\includegraphics[scale = 0.1]{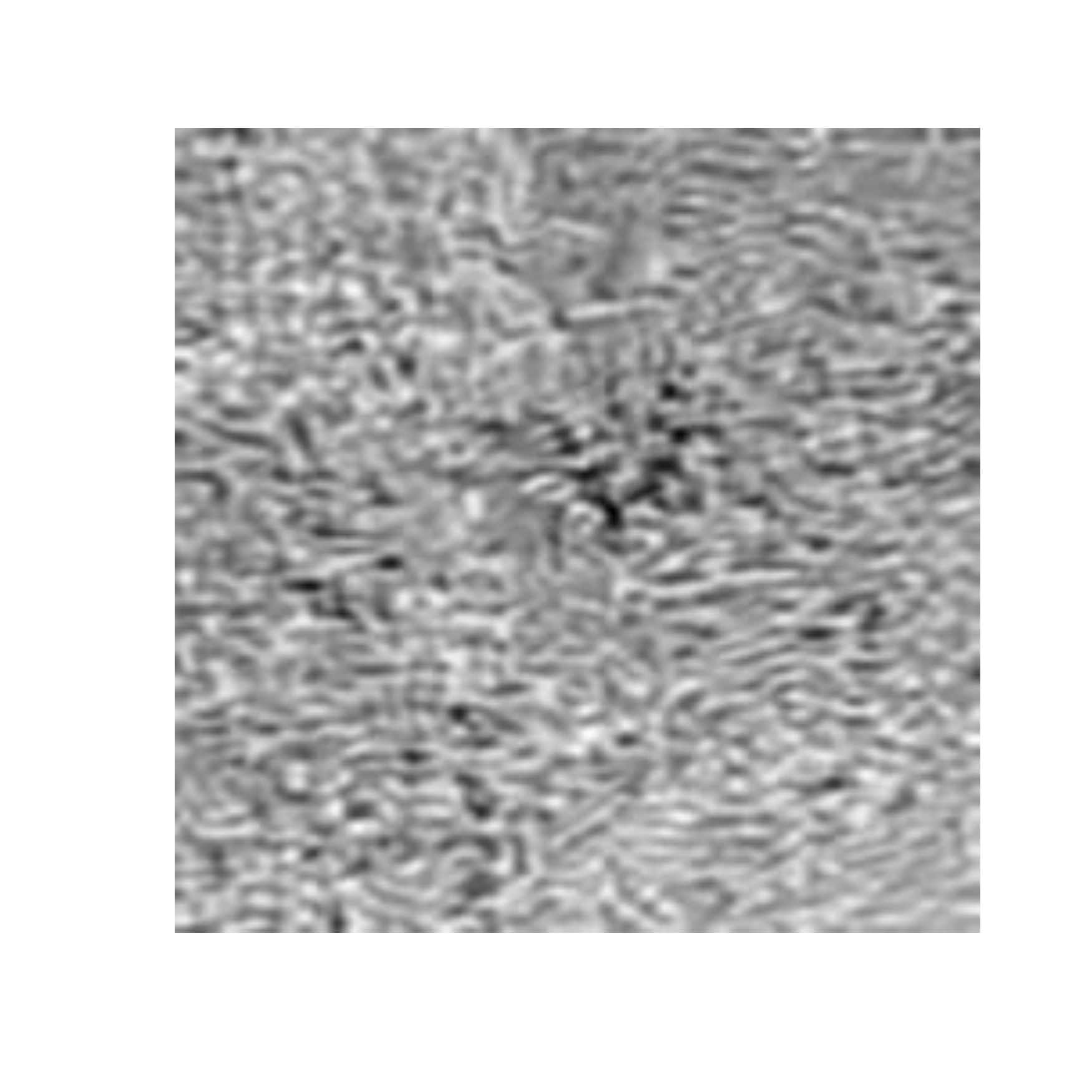}&
\includegraphics[scale = 0.1]{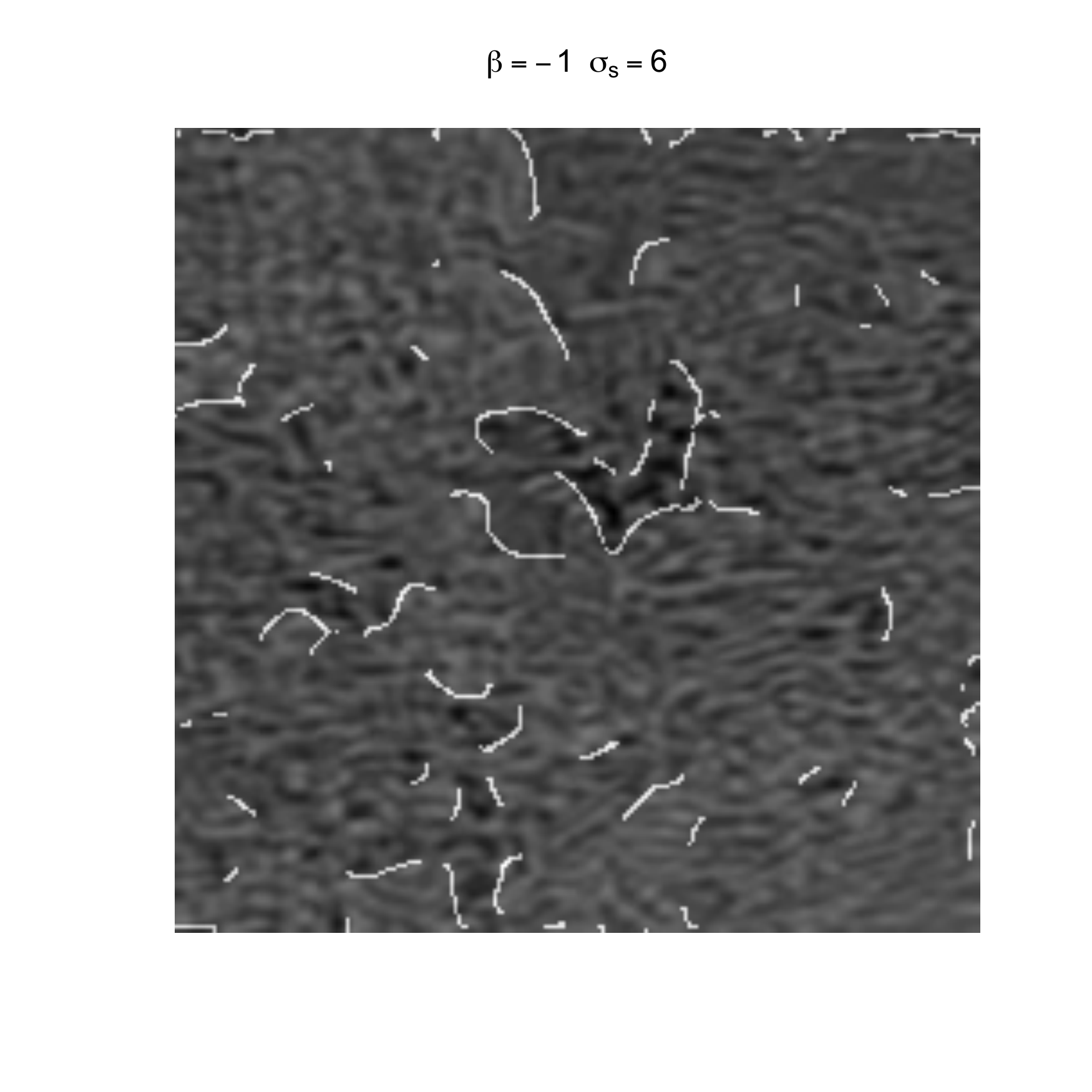}&
\includegraphics[scale = 0.1]{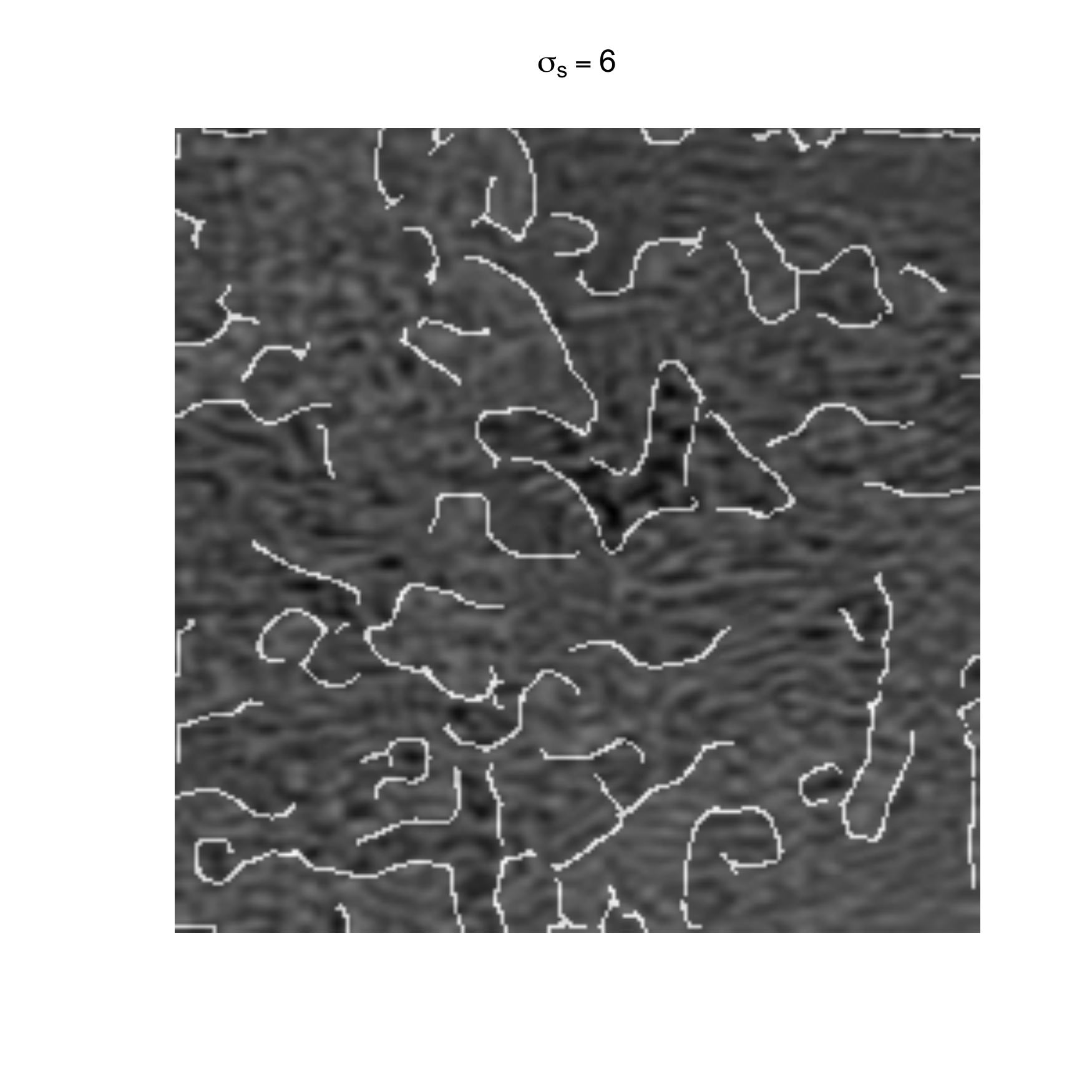}\\
\includegraphics[scale = 0.1]{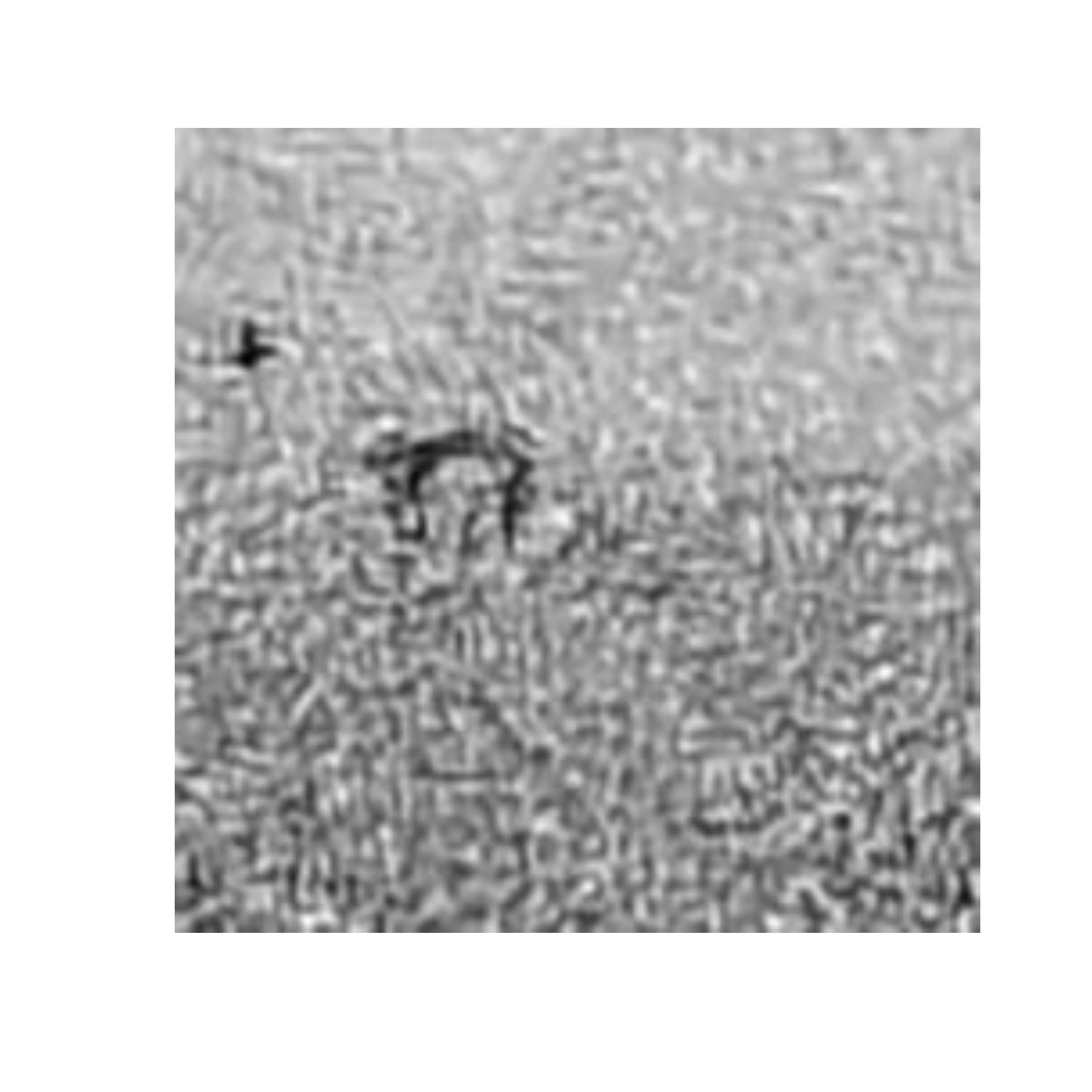}&
\includegraphics[scale = 0.1]{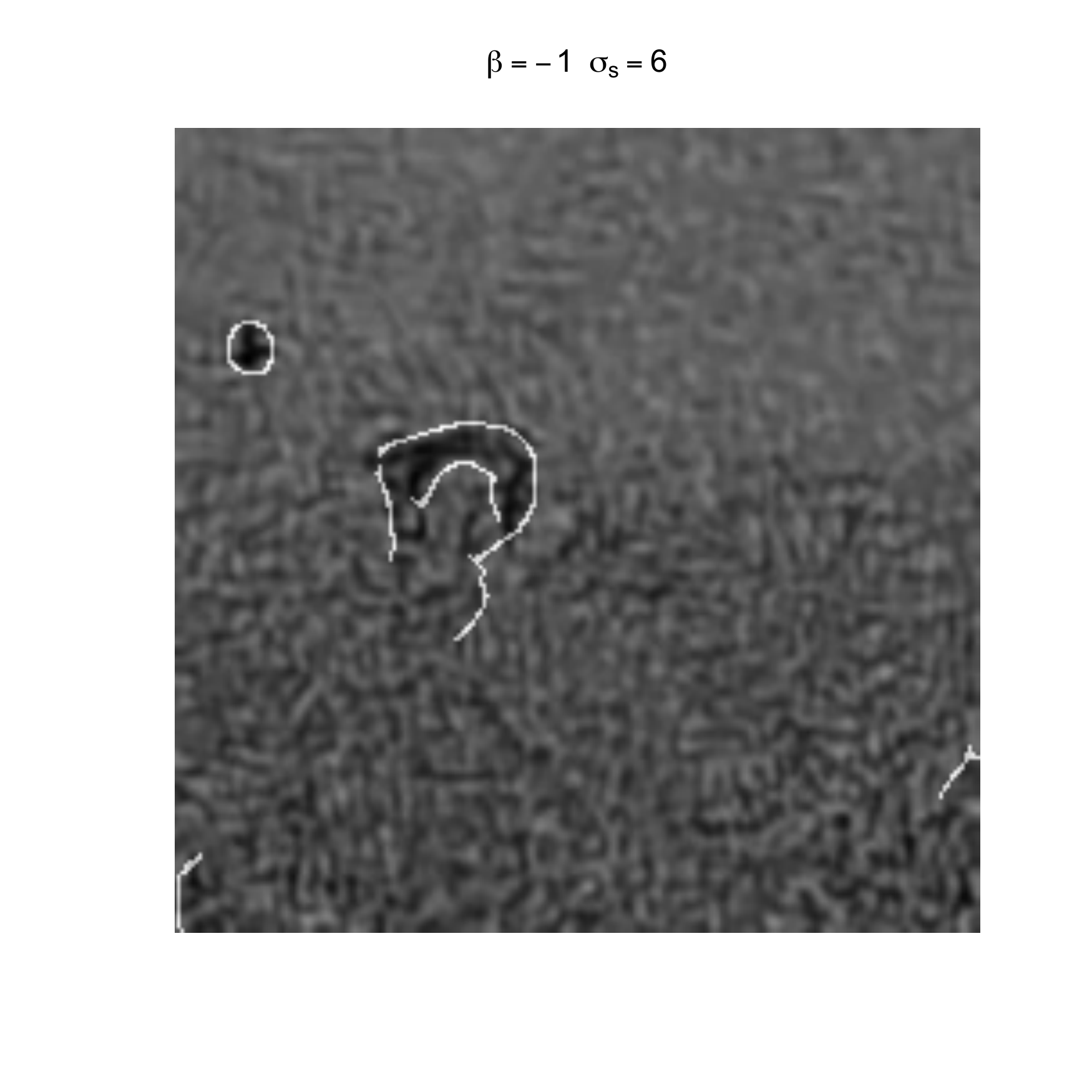}&
\includegraphics[scale = 0.1]{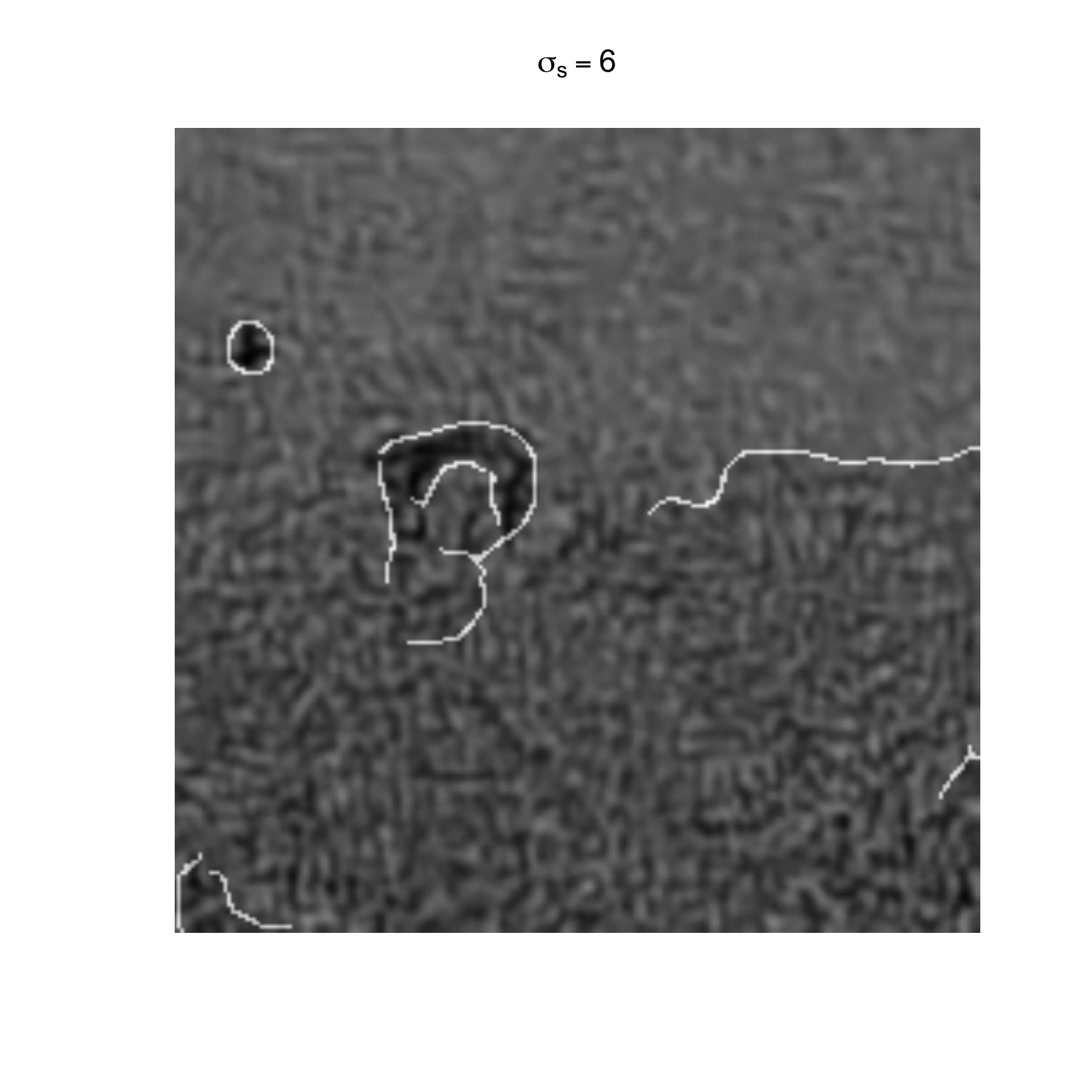}
\end{tabular}
\caption{The edge detection for the asphalt shingles roofs. The
  original images are in the first column, and the edges are overlaid
  with the original images  in the second and  the
  third columns. From the top to the bottom panels, we choose $\beta = -2, -2, -2, -1, -1$
  respectively, and $\sigma_s = 6$ in all settings, 
  where $\sigma_s$ is the Gaussian smoothing standard
  deviation. }\label{fig:real1}
\end{figure}

\begin{figure}[!h]
\centering
\begin{tabular}{ccc}
Images & Bayesian detector& Canny detector\\
\includegraphics[scale = 0.1]{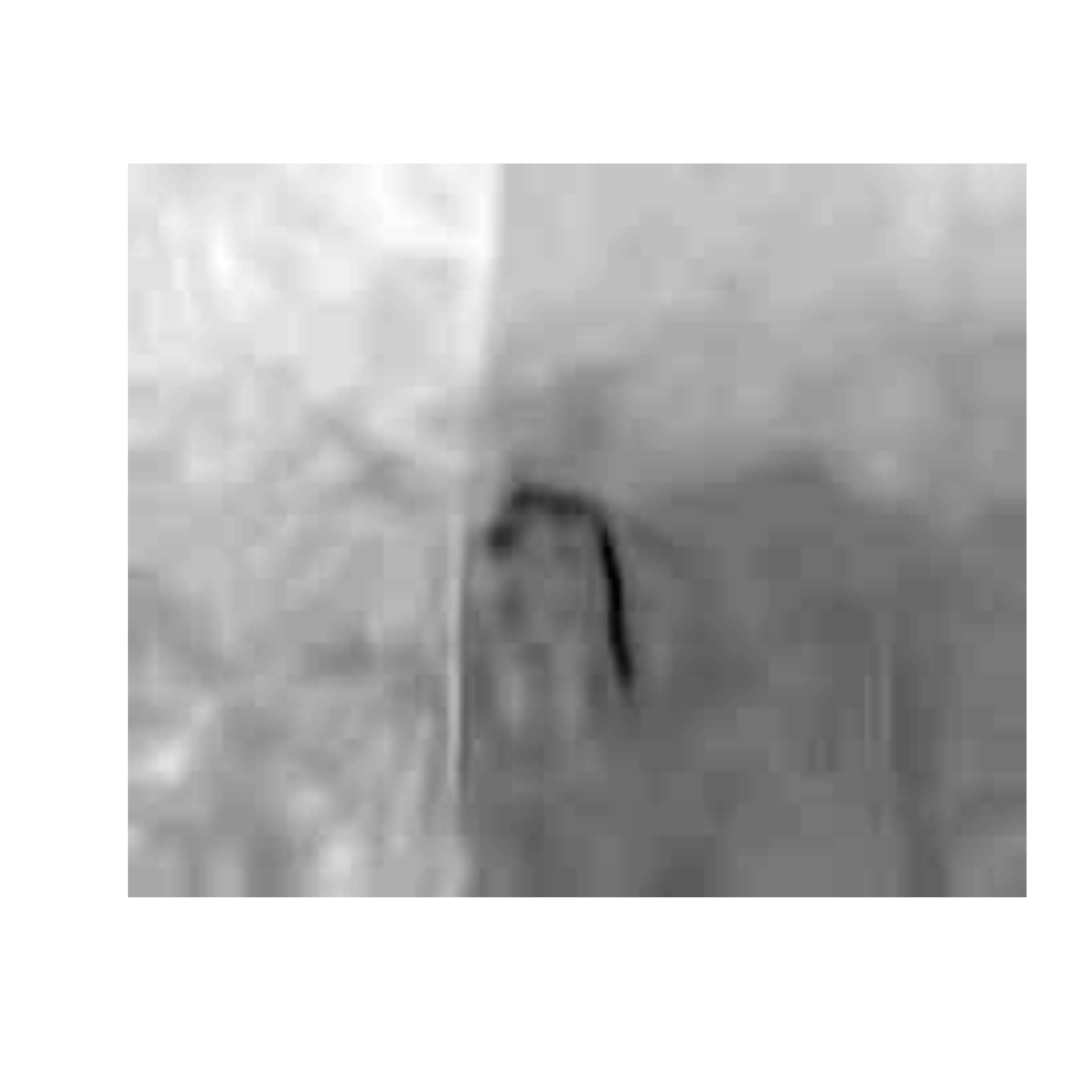}&
\includegraphics[scale = 0.1]{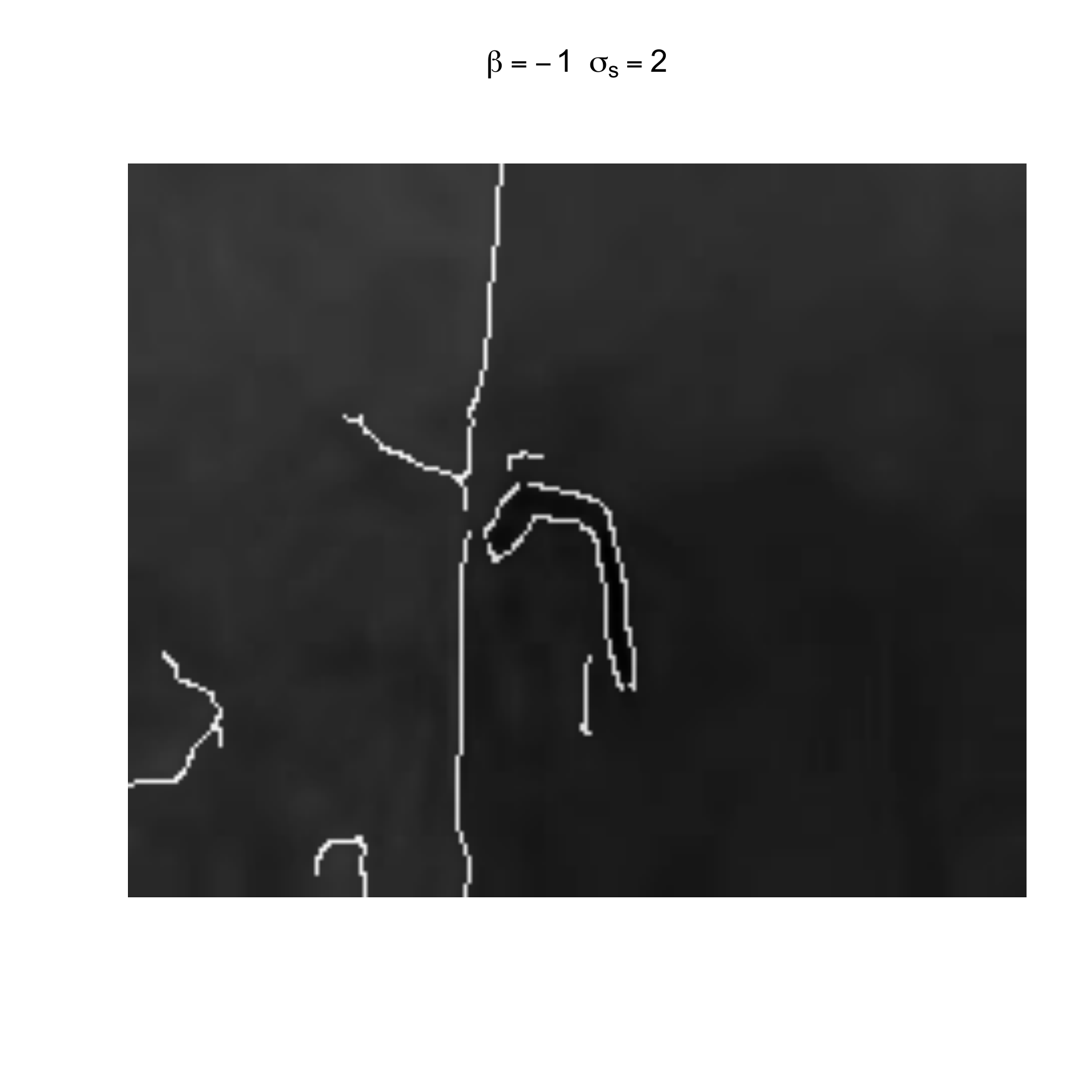}&
\includegraphics[scale = 0.1]{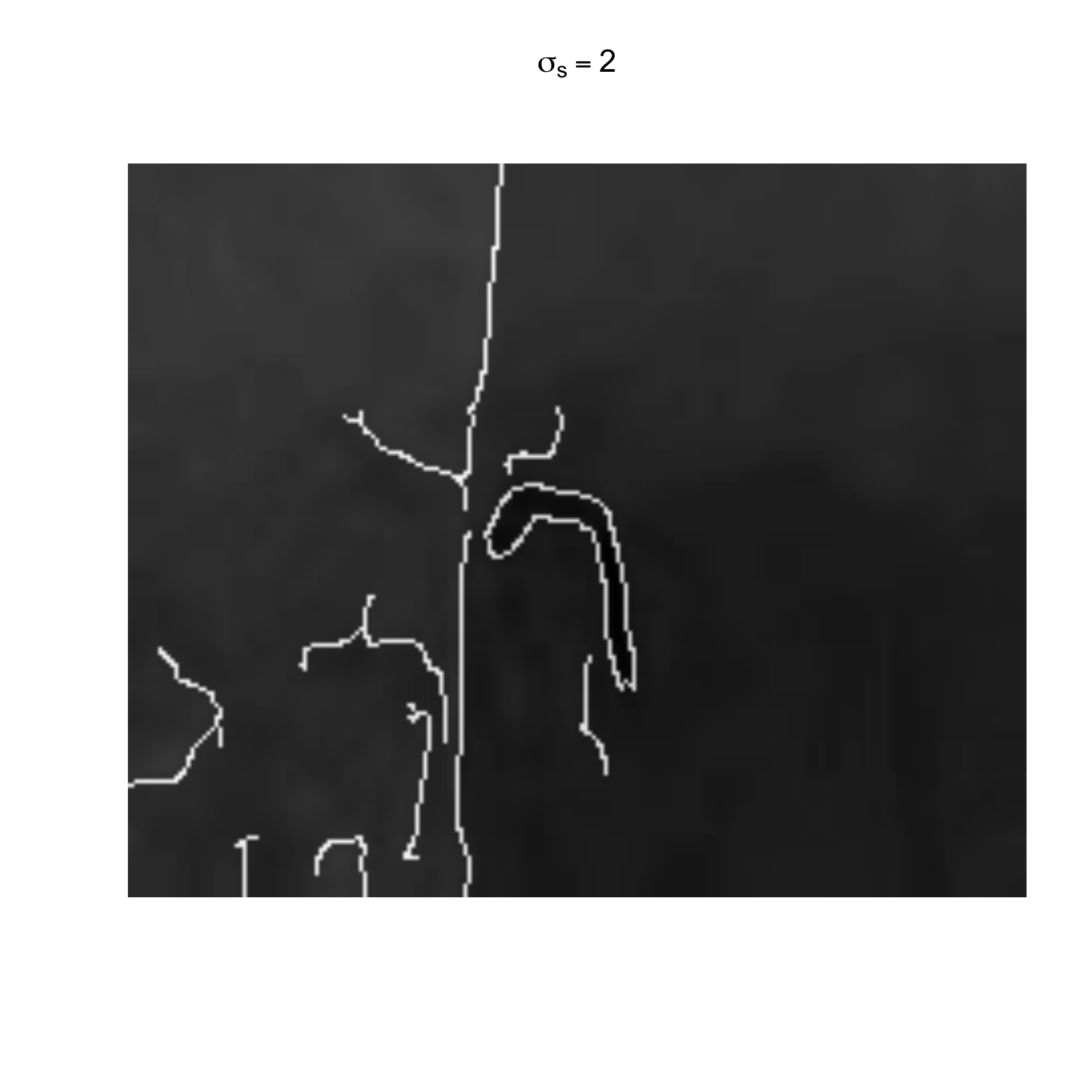}\\
\includegraphics[scale = 0.1]{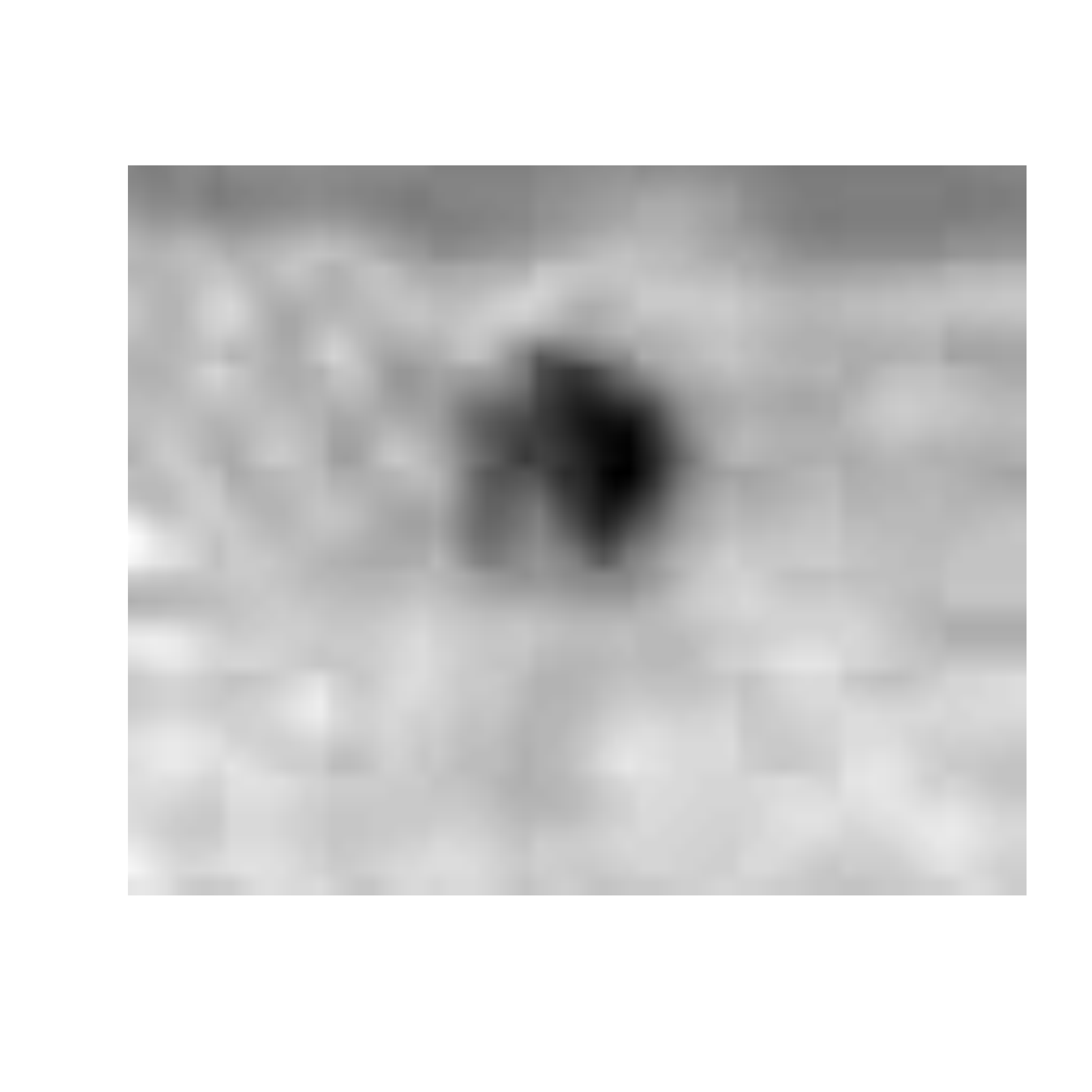}&
\includegraphics[scale = 0.1]{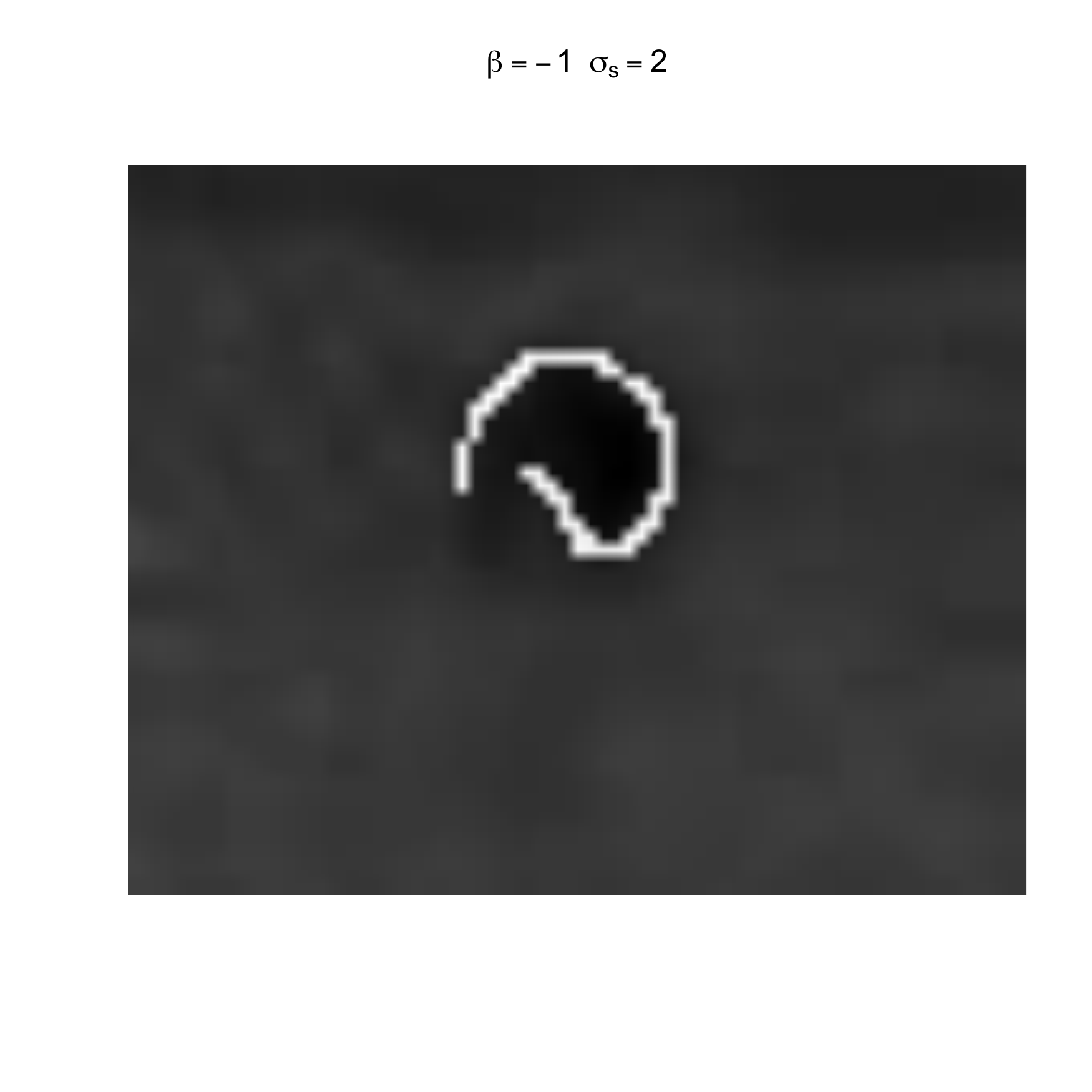}&
\includegraphics[scale = 0.1]{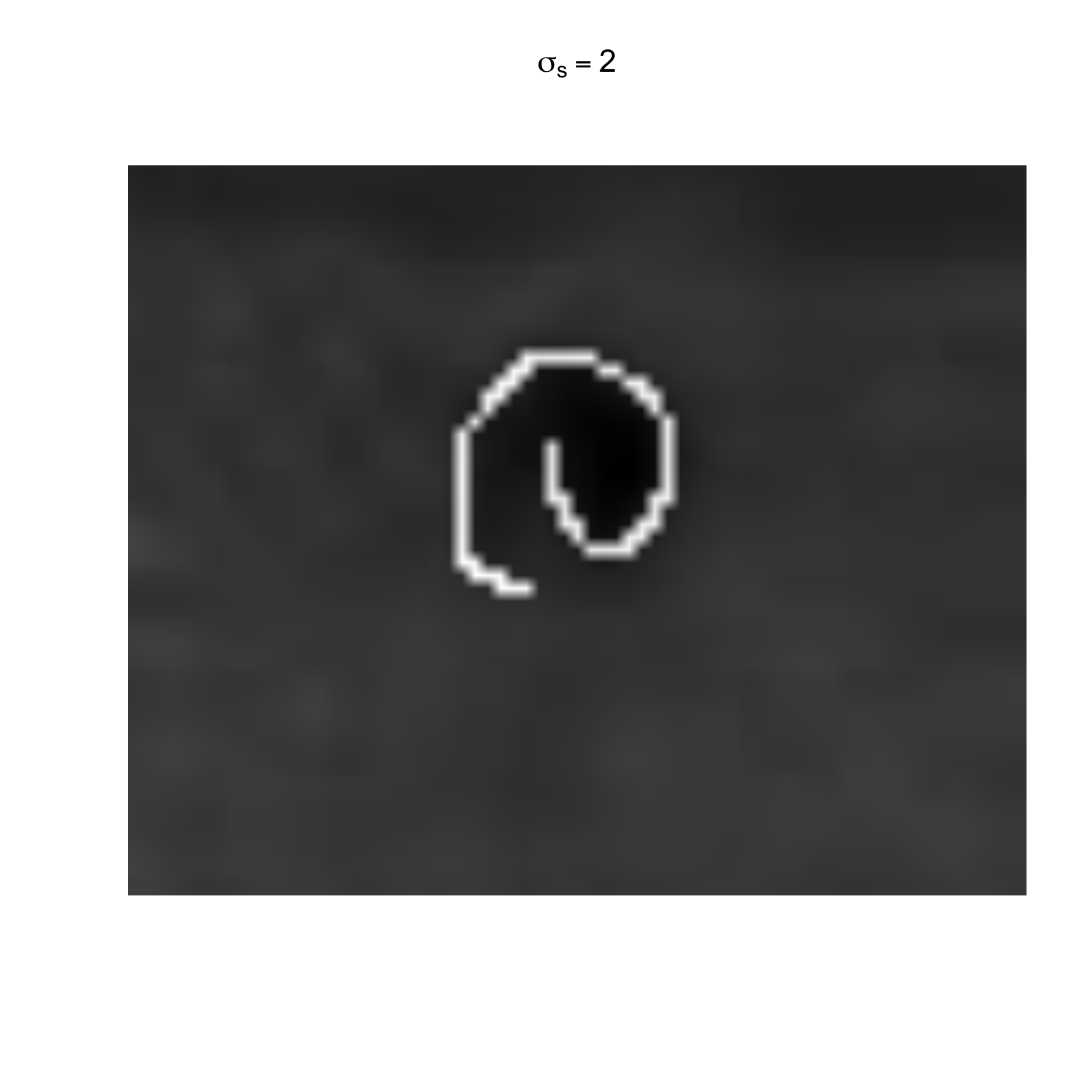}\\
\includegraphics[scale = 0.1]{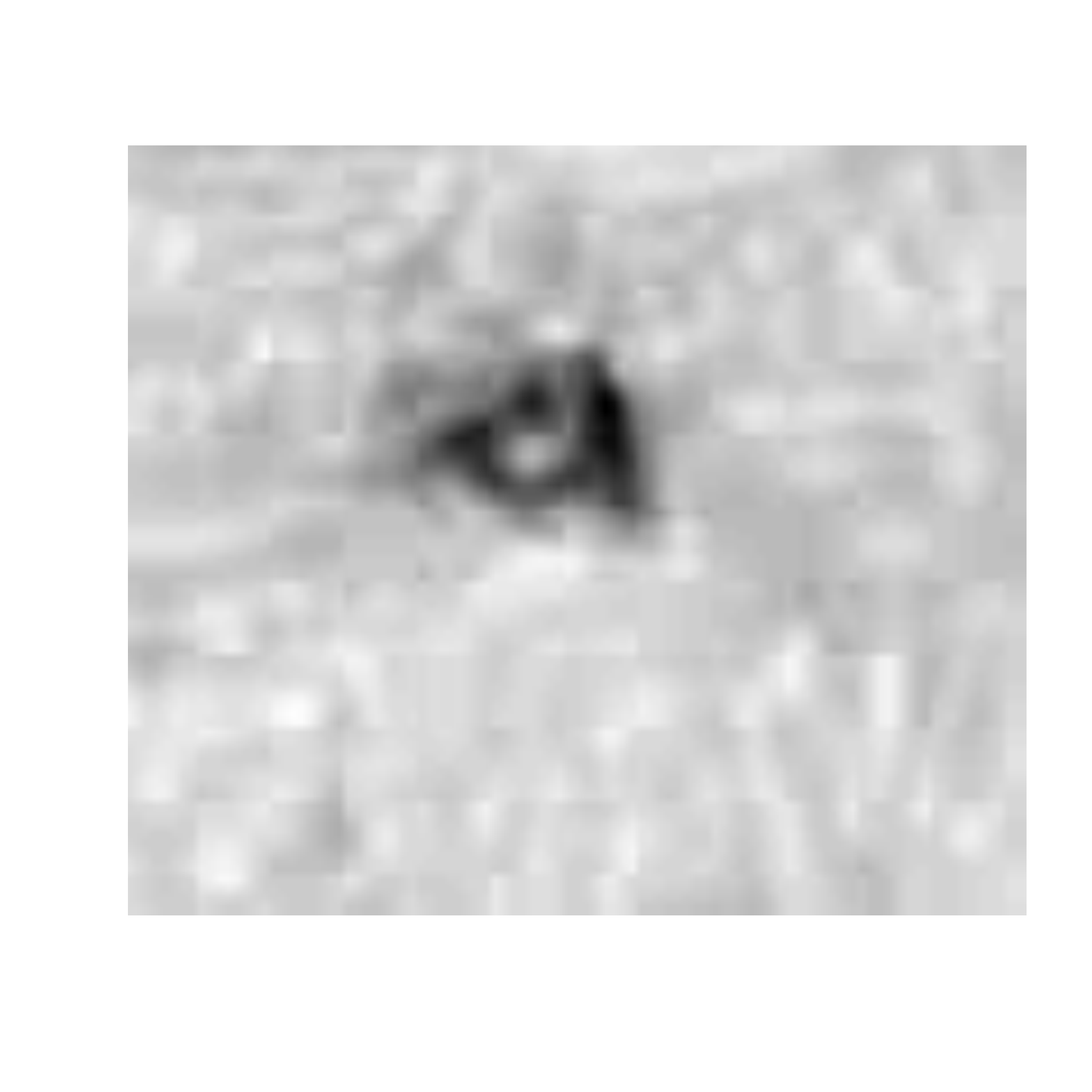}&
\includegraphics[scale = 0.1]{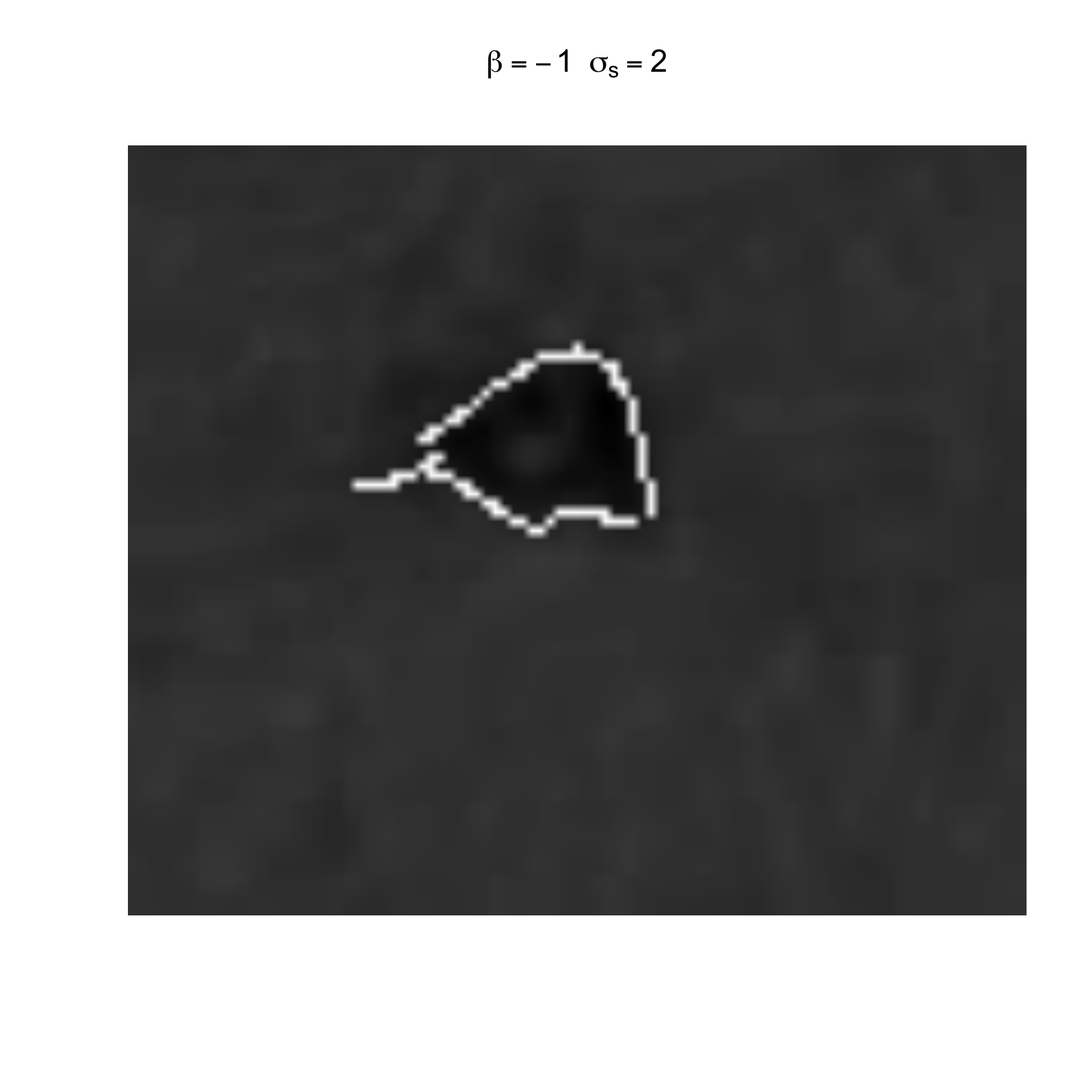}&
\includegraphics[scale = 0.1]{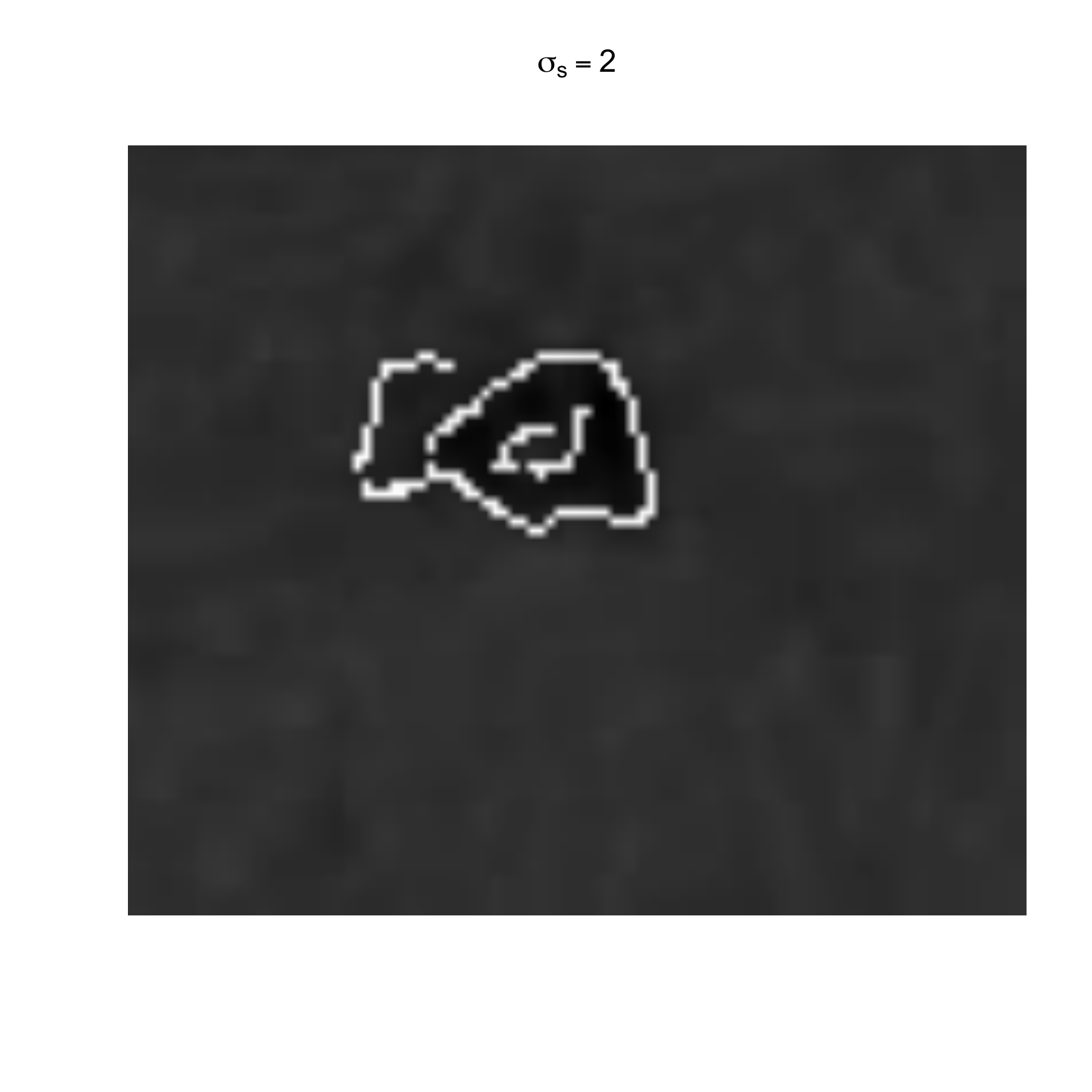}
\end{tabular}
\caption{The edge detection for the slate roofs. The
  original images are in the first column, and the edges are overlaid
  with the original images in the second and the
  third columns. We choose $\beta = -1$
and $\sigma_s = 2$ in all settings, where $\sigma_s$ is the Gaussian smoothing standard
  deviation.}\label{fig:real2}
\end{figure}

To avoid the possible intensity distortion around the shingle joints,
we first apply the proposed Bayesian detector to the edge
detection on the interiors of asphalt
shingles, and the comparison with the Canny detector is shown in
Figure \ref{fig:real1}.
It can be seen that the Bayesian detector is
able to reduce the number of the falsely discovered edge pixels.  We also
apply the Bayesian detector to the slate roofs, as shown in Figure
\ref{fig:real2}. Since the slate roofs are generally smoother than
the asphalt shingle roofs, both the Canny detector and
the Bayesian detector produce satisfactory results. With a closer
look at the resulting edges, however, we see that the Bayesian
detector still selects a smaller number of non-edge points.

In conclusion,
for the asphalt shingle roofs, the Bayesian detector performs significantly better in
controlling the false selection of non-edge points. In addition, the more
noise are the images involved, the  larger $|\beta|$'s are required to reduce the false discovery
rates. Furthermore, for the smoother materials, such as the
slate roofs, both the Bayesian detector and the Canny detector perform
satisfactorily on the edge detection.

\subsection{An integrated pipeline}
Having demonstrated the superior performance of the Bayesian detector, we
apply it to the more complicated settings,  where the images contain  complete shingles with/without damages and the
shingle joints.  We process the roof images through a complete
pipeline starting from the edge detection, edge closing (remove small
holes on edges),  connected components labeling,
to the convex hull fitting and damage extraction.  The edge detection
is the very first step in the process, which affects the outputs from the
subsequent
procedures.

\begin{figure}[!h]
\centering

\begin{tabular}{cc}
\multicolumn{2}{c}{Image}\\
\multicolumn{2}{c}{\includegraphics[scale = 0.1]{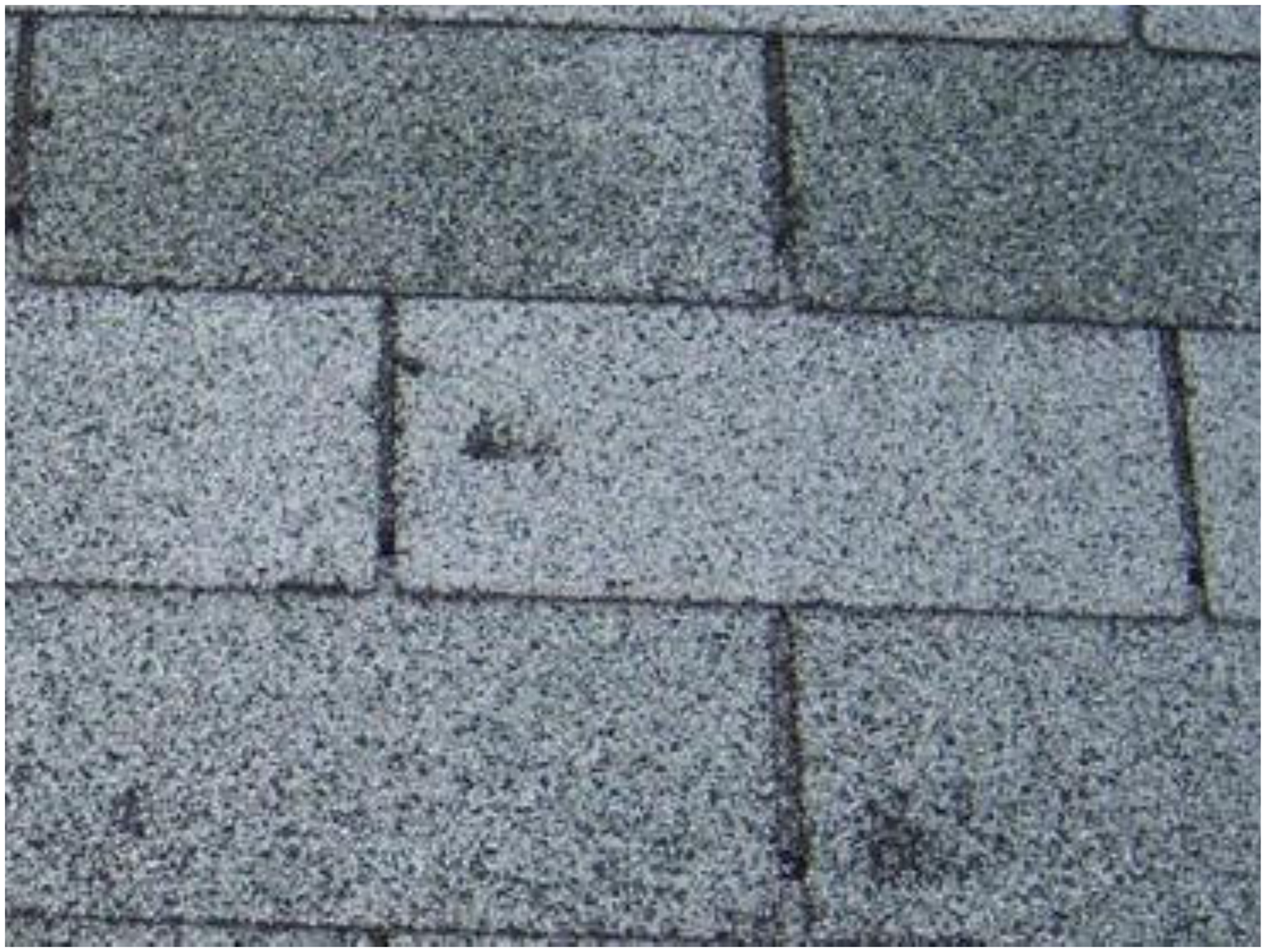}}\\
Bayesian detector& Canny detector\\
\includegraphics[scale = 0.1]{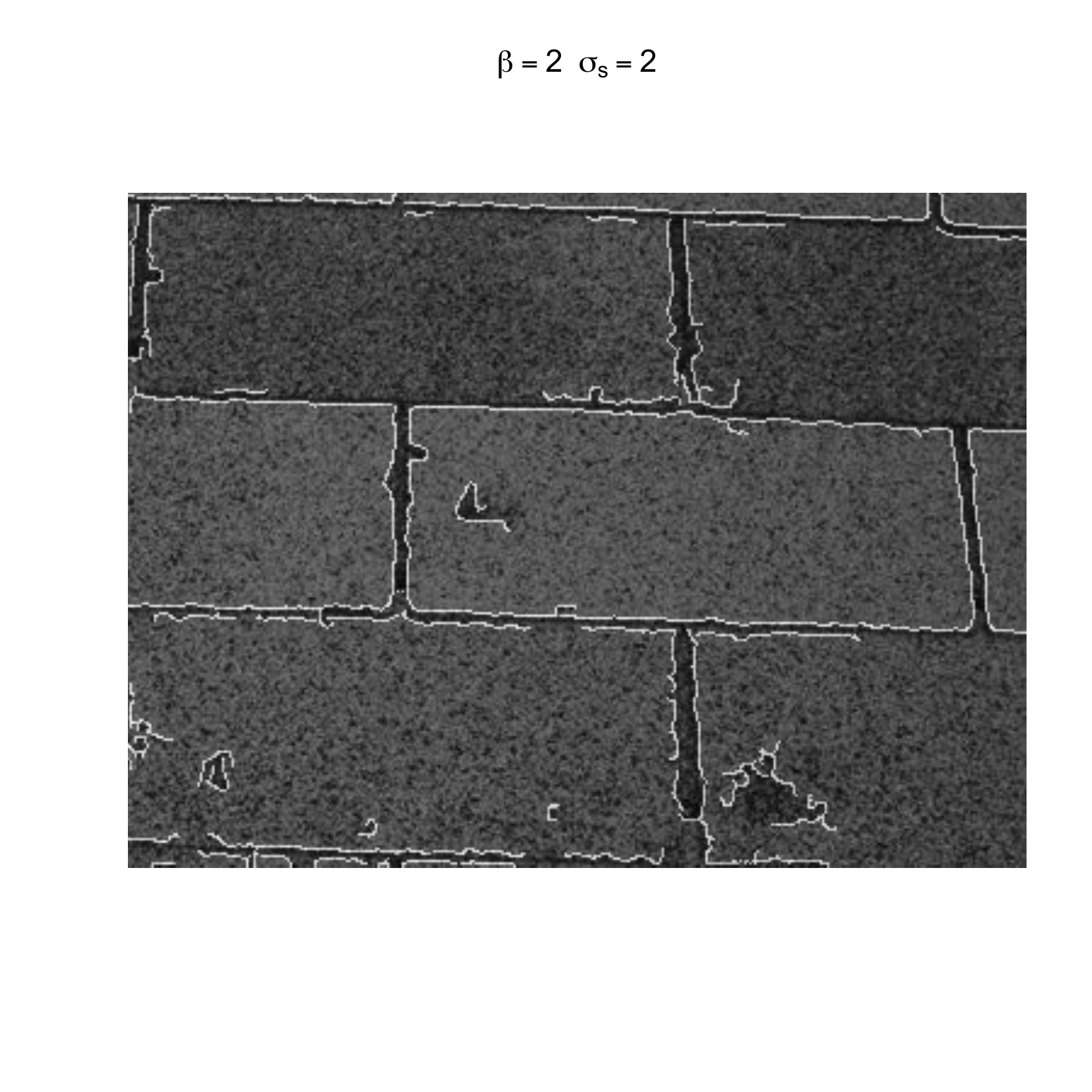}&
\includegraphics[scale = 0.1]{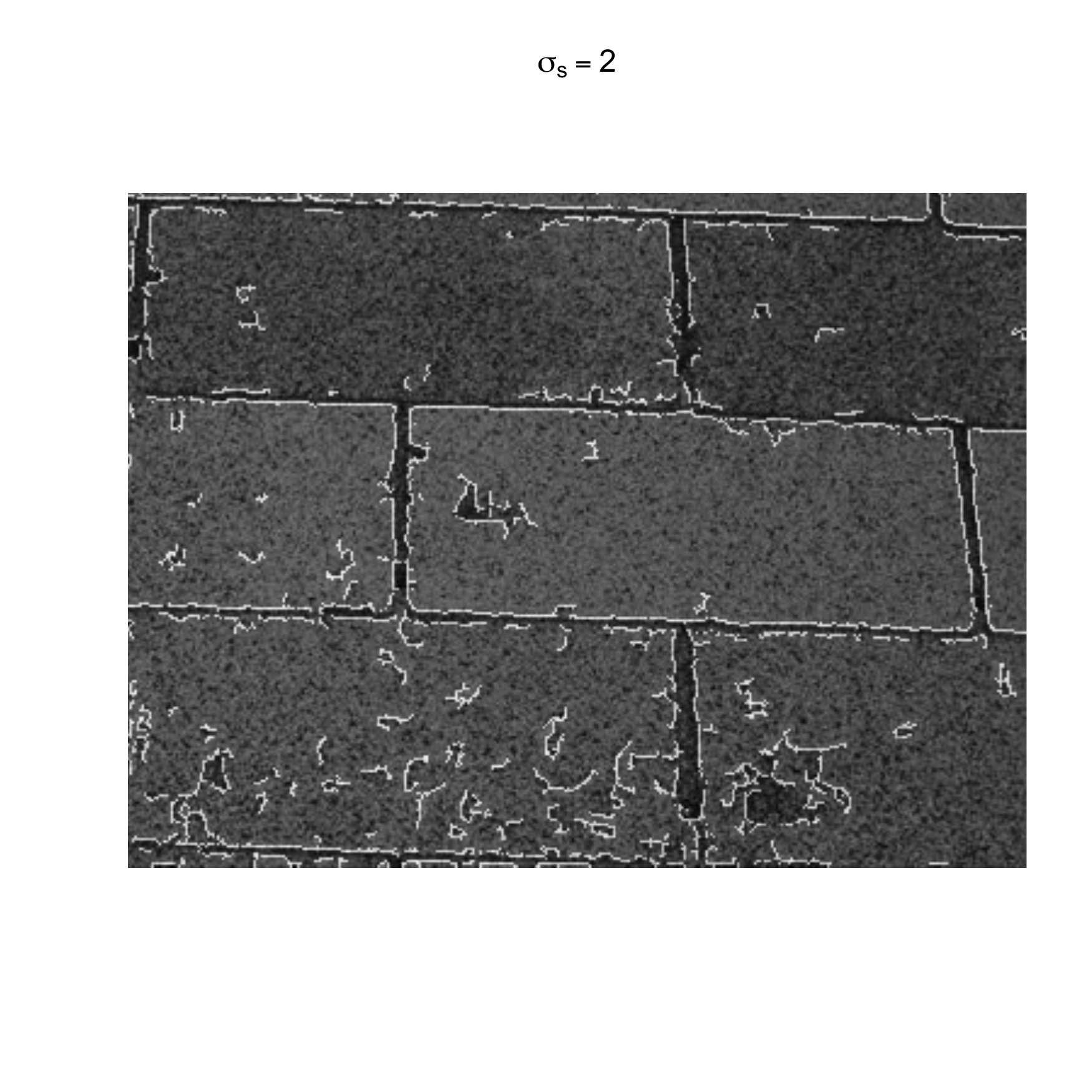}\\
\includegraphics[scale = 0.1]{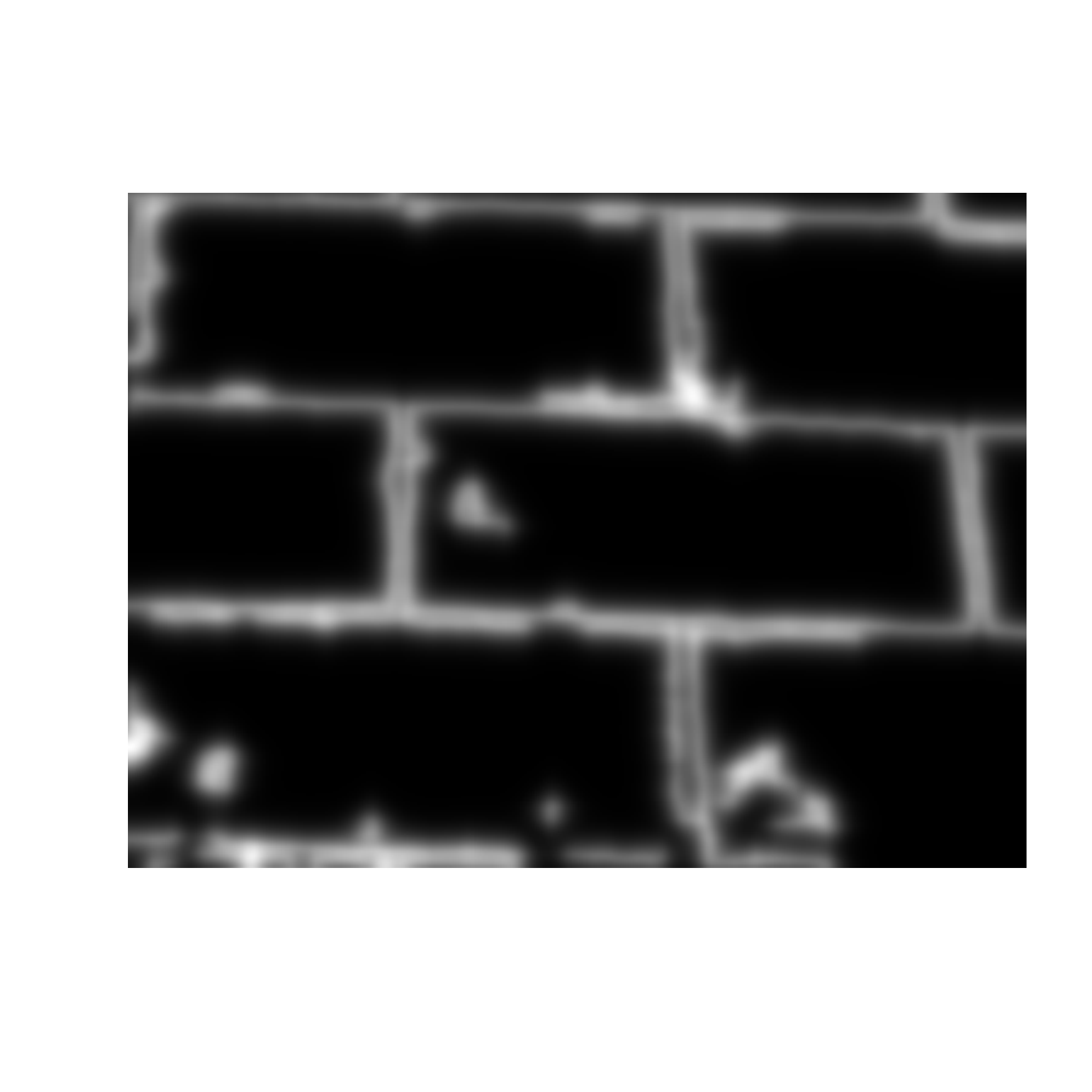}&
\includegraphics[scale = 0.1]{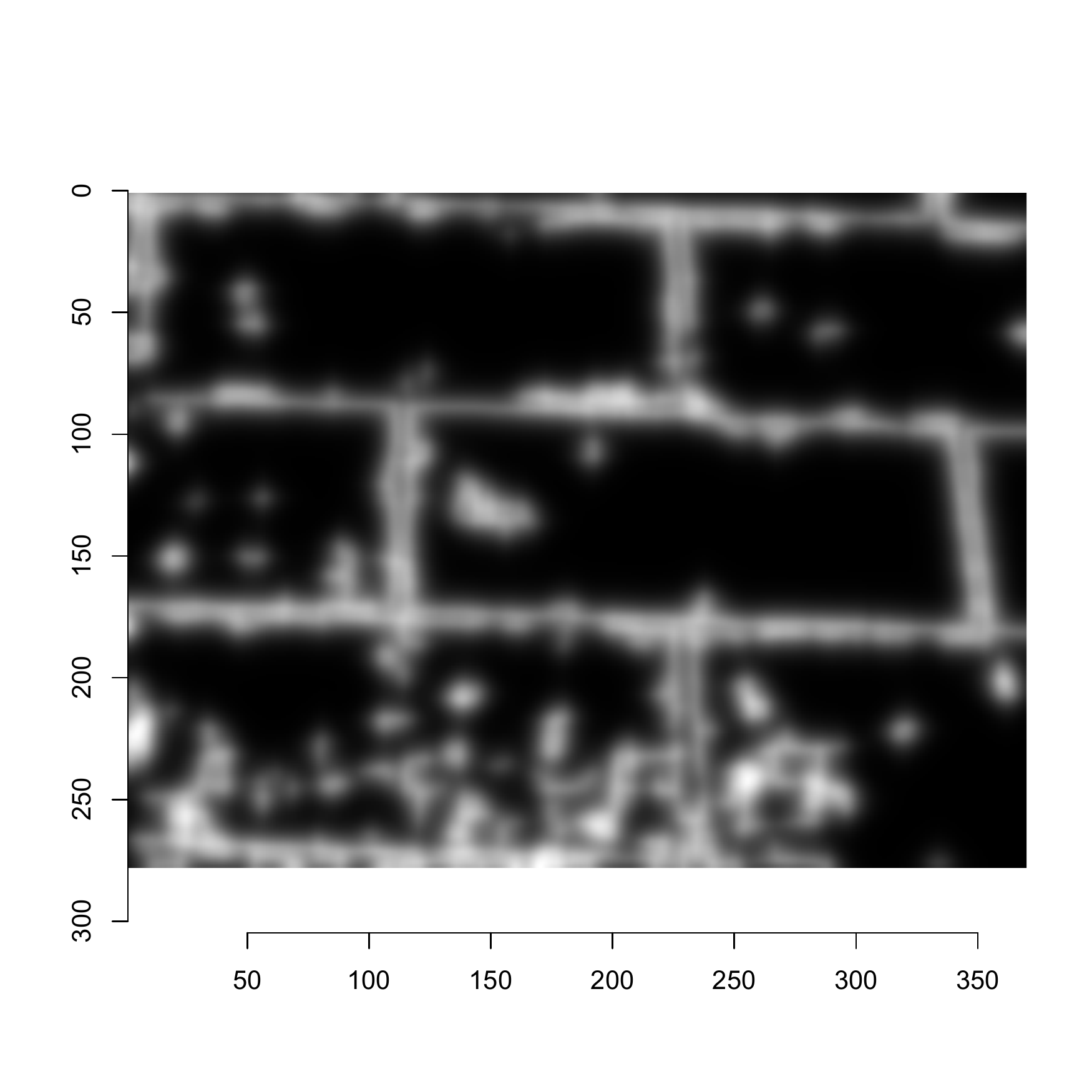}
\end{tabular}
\caption{The edge detection for the slate roofs. The
  original image is shown in the first row, the edges are overlaid
  with the original image in the second rows, and the results from the edge
  closing are in the third row. We choose $\beta = 2$
and $\sigma_s = 2$ in all settings.}\label{fig:pipline}
\end{figure}

In Figure \ref{fig:pipline}, we show the edge
detection and edge closing  results from both the Bayesian and Canny
detectors on a sampled image. The second row in Figure \ref{fig:pipline}
shows that the Bayesian detector suppresses most
non-edge points and provides precise edge regions. This
leads to better edge closing results shown in third row, which in turn
provides the clear separations between the
shingles.

\begin{figure}[!h]
\centering

\begin{tabular}{ccc}
Image & Bayesian detector& Canny detector \\
{\includegraphics[scale = 0.1]{figures/bedger.pdf}} & \includegraphics[scale = 0.1]{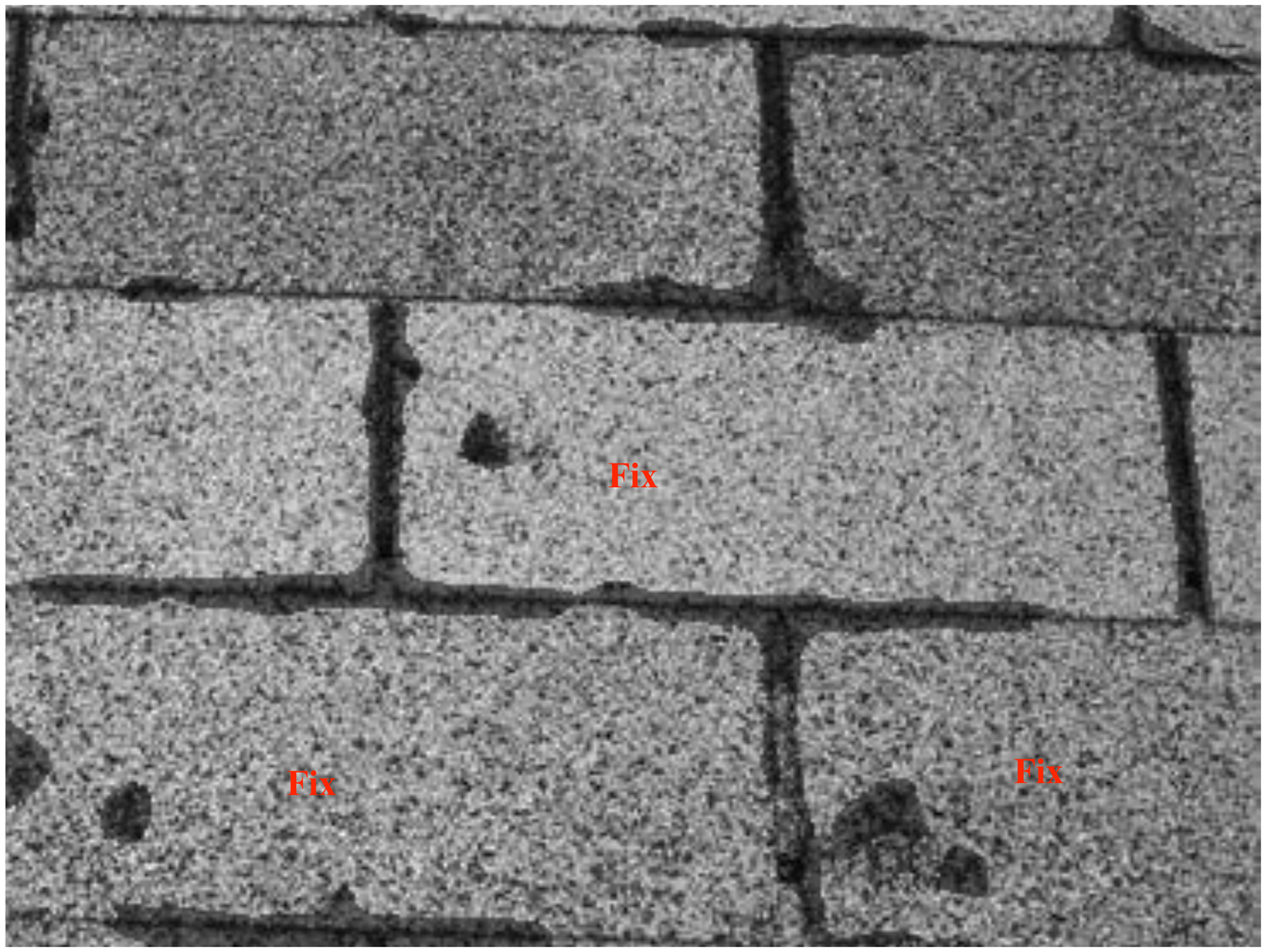}&
\includegraphics[scale = 0.1]{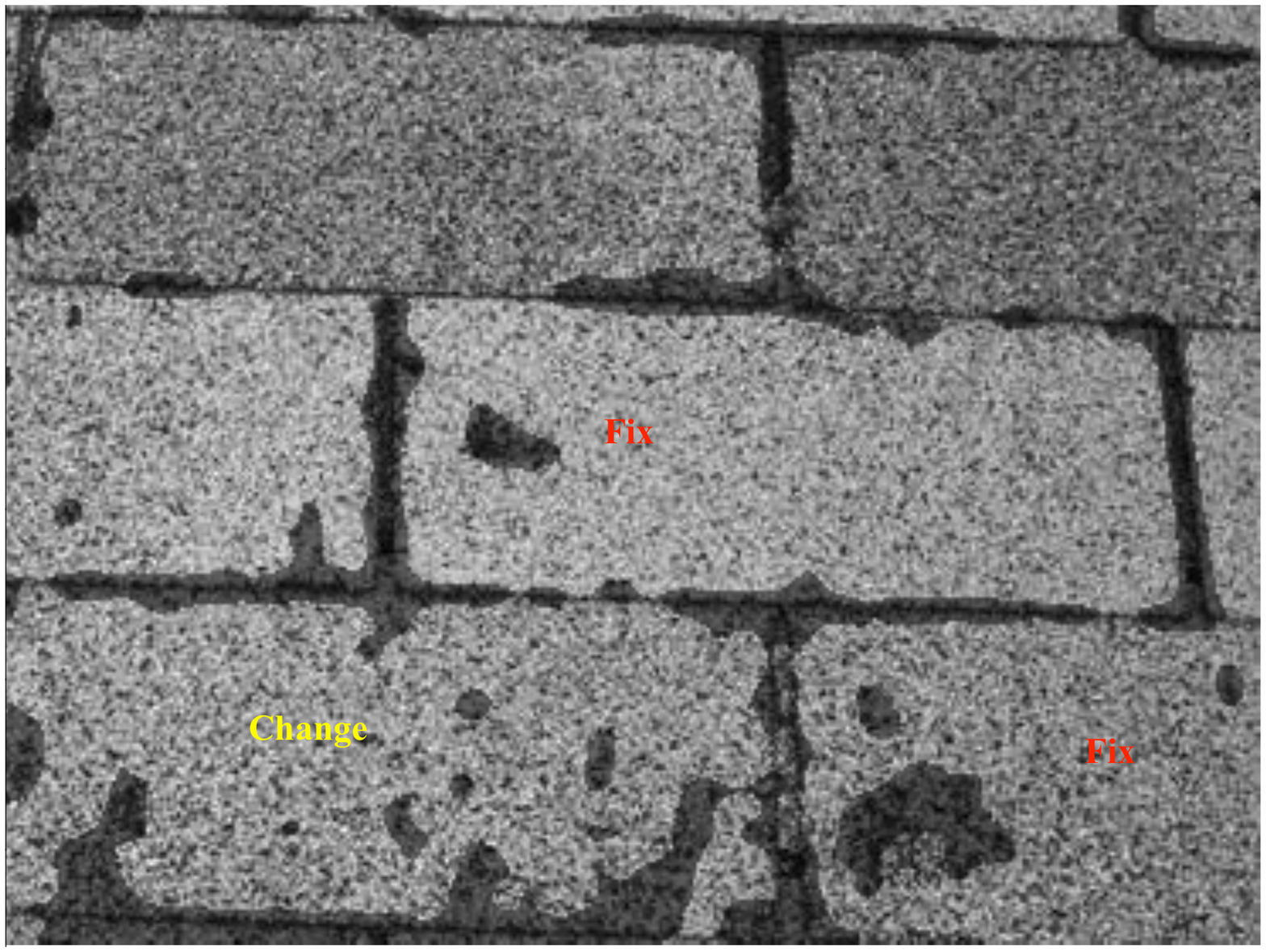}\\
{\includegraphics[scale = 0.1]{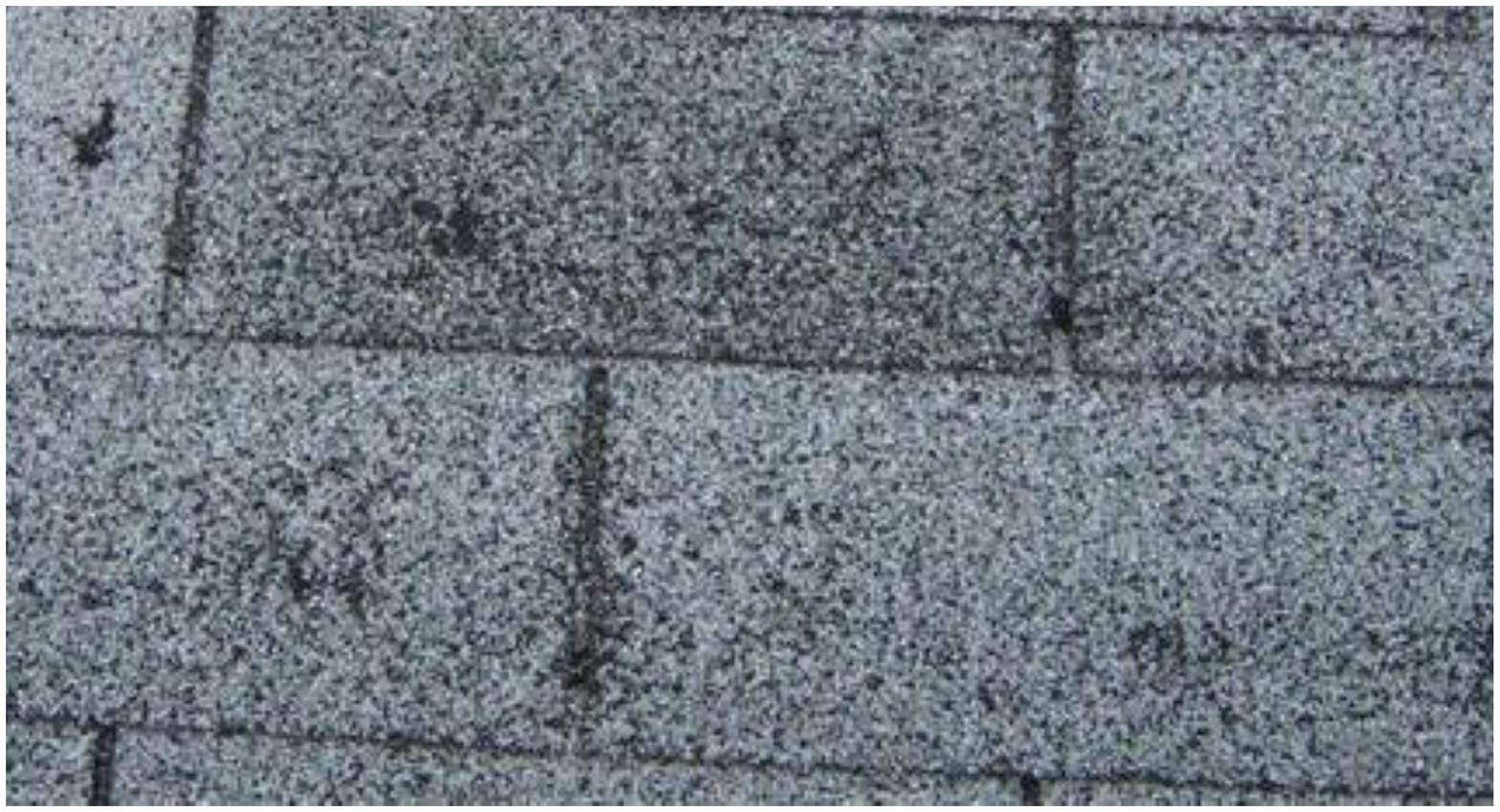}} &\includegraphics[scale = 0.1]{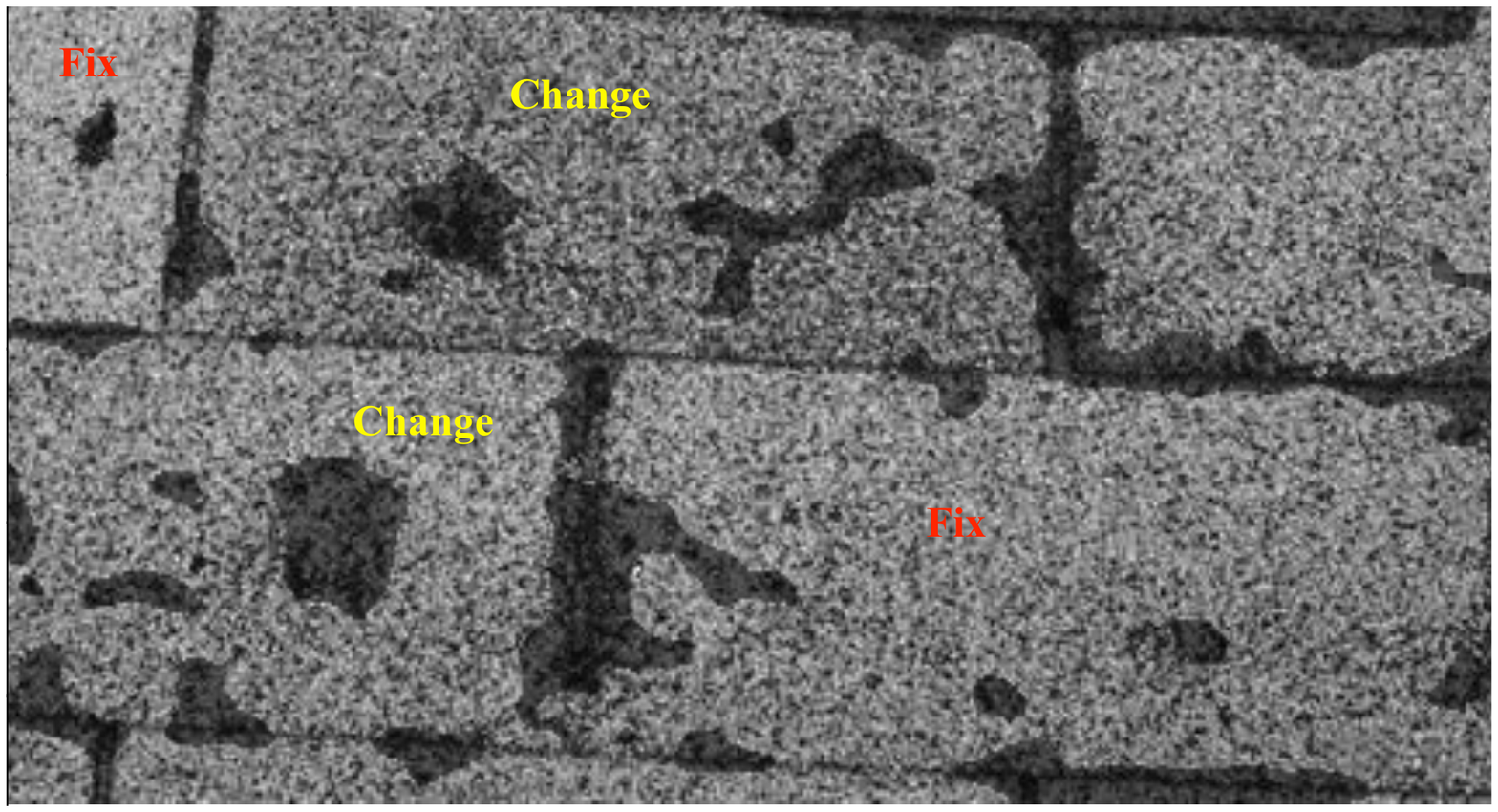}&
\includegraphics[scale = 0.1]{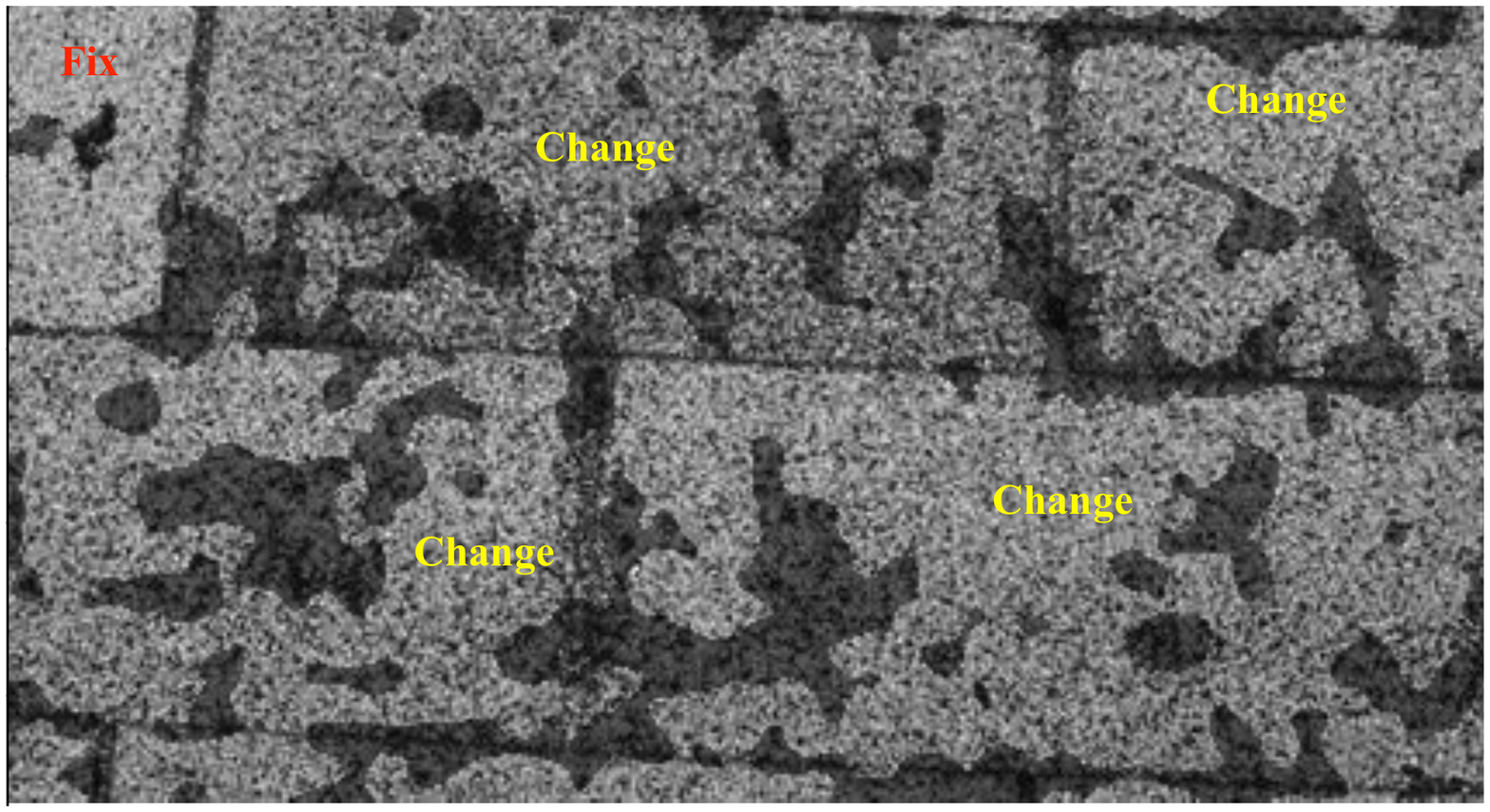}
\end{tabular}
\caption{
%The detected damages.
The detected roof damages are marked in
  black in the second and third columns, and the suggested actions are
  marked in the images. }\label{fig:insur}
\end{figure}

A better separation yields more precise labeling of connected components,
and finally leads to a more accurate damage detection.
This effect can be observed from Figure \ref{fig:insur}, where we
show the extracted damage areas overlaid with the original images from
two images.  Following these results,  the insurance companies decide to fix or
change the roof shingles based on the severeness of the damages on
their interiors. The judgement of the severeness varies across the insurance
companies. In consultation with our collaborating roofing company,
we provide some stylish examples based on our damage
detection results.
As suggested, the shingle must be replaced if more than 1/4 of the
shingle are damaged. In the upper panel of Figure \ref{fig:insur},
both the Bayesian and Canny detectors suggest to fix the shingle on the
middle and bottom-middle shingles, while the Bayesian detector
suggests to fix,  but the Canny detector suggests to
change the bottom-left shingle. Furthermore, in the bottom panel, the
Bayesian detector suggests to fix the upper-left and bottom-right shingles, keep the
right shingle, and change the others, while the Canny detector
suggests to change all the shingles besides the upper-left one. Compared with
our observations on the real images, the conclusions from the
Bayesian detector are much closer to the human decisions.

\section{Discussion}\label{sec:con}
Edge detection is a classical problem in the computer vision area,
which is dominated by the convolution based methods. We view the
problem from the statistical perspective and creatively introduce the Bayesian tools to
tackle the challenging problems in the field.
The proposed Bayesian detector can successfully
identify the edge points, while inducing smaller probabilities of the false
discovery compared to the standard Canny detector. In addition,
we introduce a general framework in constructing the non-local priors
which generalizes the moment
and inverse moment priors in \cite{johnson2010}. The
reflected priors achieve the same asymptotic properties as the moment
prior when $\beta>0$ and as the inverse moment prior when $\beta<0$.

The goal of the paper is to develop a robust method for extracting
damages from the drone images. As a by-product, we introduce a class of new prior and discuss
their applications in the image edge detection problem. Therefore, we
do not discuss the optimal selection of the tuning parameters,
including $\beta$ and $\bSig$ in (\ref{eq:mgnonlocal}) and the number
of the edge pixels.  In
fact, the  parameter tuning process is a difficult task in the
current situation, because there is no enough labeled training
samples, and hence we are not able to verify the selection.
However, in our simulation and real data analysis, the
performance of the Bayesian detector is not sensitive to the changes
in the tuning parameters.  Although it is critical in the statistical
learning procedure, we leave the selection of the tuning parameters to
future works when sufficient samples are available for training.

\bibliographystyle{IEEEbib}
\bibliography{nonlocal.bib}
\begin{comment}
\begin{IEEEbiography}[{\includegraphics[width=1in,height=1.25in,clip,keepaspectratio]{feijiang.jpg}}]{Fei Jiang}
received the Ph.D. degree in statistics from Rice University in
2013. Before join the University of Hong Kong, she was a postdoctoral
research fellow in Harvard University. She joined the Department of
Statistics \& Actuarial Science, University of Hong
Kong as an assistant professor in statistics  in 2016. 
\end{IEEEbiography}

\begin{IEEEbiography}[{\includegraphics[width=1in,height=1.5in,clip,keepaspectratio]{gyin.jpg}}]{Guosheng Yin}
received the Ph.D. in Biostatistics from the University of North
Carolina at Chapel Hill in 2003. Before join the University of Hong
Kong, he was an associated professor in the Department of
Biostatistics, University of Texas M.D. Anderson Cancer Center. He
joined the Department of Statistics \& Actuarial Science, University of
Hong Kong in the year 2014. He was world's top 1\% of scientists by
Thomson Reuters (2015). He has been the fellow of American Statistics
Association since 2013, and elected member of International
Statistics Institute since 2012. 
\end{IEEEbiography}
\vfill
\end{comment}

\end{document}